%% file: eccv2022submission.tex
\documentclass[runningheads]{llncs}
\usepackage{graphicx}

\usepackage{tikz}
\usepackage{comment}
\usepackage{amsmath,amssymb} %
\usepackage{color}
\usepackage{copyrightbox}
\usepackage{xcolor}
\usepackage{color, colortbl}
\usepackage{soul}

\input{tools}

\algrenewcommand\alglinenumber[1]{\tiny #1:}
\algrenewcommand\algorithmicindent{1.em}

\begin{document}
\pagestyle{headings}
\mainmatter
\def\ECCVSubNumber{7093}  %

\title{Improving Generalization in Federated Learning by Seeking Flat Minima} %

\titlerunning{Improving Generalization in Federated Learning by Seeking Flat Minima}
\newcommand*\samethanks[1][\value{footnote}]{\footnotemark[#1]}

\author{
Debora Caldarola\thanks{Equal contribution}$^1$%
\and Barbara Caputo$^{1,2}$%
\and Marco Ciccone\samethanks$^1$%
}
\authorrunning{D. Caldarola et al.}
\institute{
$^1$Politecnico di Torino, $^2$CINI\\
\email{name.surname@polito.it}\\
}

\maketitle

\begin{abstract}
Models trained in federated settings often suffer from degraded performances and fail at generalizing, especially when facing heterogeneous scenarios. In this work, we investigate such behavior through the lens of geometry of the loss and Hessian eigenspectrum, linking the model's lack of generalization capacity to the sharpness of the solution.
Motivated by prior studies connecting the sharpness of the loss surface and the generalization gap, we show that i) training clients locally with Sharpness-Aware Minimization (SAM) or its adaptive version (ASAM) and ii) averaging stochastic weights (SWA) on the server-side can substantially improve generalization in Federated Learning and help bridging the gap with centralized models. 
By seeking parameters in neighborhoods having uniform low loss, the model converges towards flatter minima and its generalization significantly improves in both homogeneous and heterogeneous scenarios. Empirical results demonstrate the effectiveness of those optimizers across a variety of benchmark vision datasets (e.g. \textsc{Cifar10/100}, Landmarks-User-160k, \textsc{Idda}) and tasks (large scale classification, semantic segmentation, domain generalization).\blfootnote{Official code: \url{https://github.com/debcaldarola/fedsam}} 
\end{abstract}

\input{sections/1.introduction}
\input{sections/2.related}
\input{sections/3.heterogeneous_fl}
\input{sections/4.method}
\input{sections/5.experiments}
\input{sections/6.conclusions}

\bibliographystyle{splncs04}
\bibliography{bibliography}

\clearpage
\appendix
\input{supplementary}

\end{document}

%% file: tools.tex
\usepackage{xspace}
\usepackage{xstring}
\usepackage{xcolor-material}
\usepackage{color, colortbl}
\usepackage{soul}
\usepackage{nicefrac}
\usepackage{subfig}
\usepackage{multirow}
\usepackage{multicol}
\usepackage{booktabs}
\usepackage{algorithm,algpseudocode}
\usepackage{graphics}
\usepackage{adjustbox}
\usepackage{amssymb}%
\usepackage{pifont}%
\usepackage{arydshln}
\usepackage{caption}

\usepackage{amsfonts}
\usepackage{amsmath}
\usepackage{amssymb}
\usepackage{array}
\usepackage{bm}
\usepackage{booktabs}
\usepackage{comment}
\usepackage{graphicx}
\usepackage{hhline}
\usepackage[pagebackref=true,breaklinks=true,letterpaper=true,colorlinks,bookmarks=false]{hyperref}
\usepackage{ifthen}
\usepackage{microtype}
\usepackage{nicefrac}
\usepackage{pifont}
\usepackage{soul}
\usepackage[normalem]{ulem}
\usepackage{url}
\usepackage{enumerate}

\makeatletter
\let\origparagraph\paragraph
\renewcommand\paragraph{\@ifstar{\starparagraph}{\nostarparagraph}}
\newcommand\nostarparagraph[1]
{\origparagraph{#1}\paragraphpostlude}
\newcommand\starparagraph[1]
{\paragraphprelude\origparagraph*{#1}\paragraphpostlude}
\newcommand\paragraphprelude{%
  \vspace{-14pt}%
}
\newcommand\paragraphpostlude{%
}

\UseRawInputEncoding
\hypersetup{
    colorlinks,
    linkcolor={red},
    citecolor={green!80!black},
    urlcolor={blue}
}

\renewcommand{\subsubsection}[1]{\paragraph{\textbf{#1}}}

\newcommand\blfootnote[1]{%
  \begingroup
  \renewcommand\thefootnote{}\footnote{#1}%
  \addtocounter{footnote}{-1}%
  \endgroup
}

\newcommand{\fedavgm}{\texttt{FedAvgM}\xspace}
\newcommand{\fedavg}{\texttt{FedAvg}\xspace}
\newcommand{\fedsam}{\texttt{FedSAM}\xspace}  %
\newcommand{\fedasam}{\texttt{FedASAM}\xspace}  %
\newcommand{\swa}{\texttt{SWA}\xspace}  %

\newcommand{\fedavgswa}{\texttt{FedAvg + SWA}\xspace}
\newcommand{\fedsamswa}{\texttt{FedSAM + SWA}\xspace}  %
\newcommand{\fedasamswa}{\texttt{FedASAM + SWA}\xspace}
\newcommand{\fedavgmswa}{\texttt{FedAvgM + SWA}\xspace}

\newcommand{\fedavgmasam}{\texttt{FedAvgM + ASAM}\xspace}

\newcommand{\sam}{\texttt{SAM}\xspace}  %
\newcommand{\asam}{\texttt{ASAM}\xspace}  %
\newcommand{\samswa}{\texttt{SAM + SWA}\xspace}  %
\newcommand{\asamswa}{\texttt{ASAM + SWA}\xspace}
\newcommand{\mixup}{\texttt{Mixup}\xspace}
\newcommand{\cutout}{\texttt{Cutout}\xspace}

\newcommand{\ie}{i.e.\ }
\newcommand{\eg}{e.g.\ }

\newcommand{\fedprox}{\texttt{FedProx}\xspace}  %

\newcommand{\fedproxasam}{\texttt{FedProx + ASAM}\xspace}

\newcommand{\scaffold}{\texttt{SCAFFOLD}\xspace}  %

\newcommand{\scaffoldasam}{\texttt{SCAFFOLD + ASAM}\xspace}

\newcommand{\feddyn}{\texttt{FedDyn}\xspace}  %

\newcommand{\feddynsam}{\texttt{FedDyn + SAM}\xspace}
\newcommand{\feddynasam}{\texttt{FedDyn + ASAM}\xspace}

\newcommand{\adabest}{\texttt{AdaBest}\xspace}  %

\newcommand{\adabestasam}{\texttt{AdaBest + ASAM}\xspace}

\newcommand{\quotes}[1]{``#1''} 

\makeatletter
\DeclareRobustCommand\onedot{\futurelet\@let@token\@onedot}
\def\@onedot{\ifx\@let@token.\else.\null\fi\xspace}

\def\eg{\emph{e.g}\onedot} 
\def\ie{\emph{i.e}\onedot}

\makeatother

\makeatletter
\newcommand*{\inlineequation}[2][]{%
  \begingroup
    \refstepcounter{equation}%
    \ifx\\#1\\%
    \else
      \label{#1}%
    \fi
    \relpenalty=10000 %
    \binoppenalty=10000 %
    \ensuremath{%
      #2%
    }%
    ~\@eqnnum
  \endgroup
}
\makeatother

%% file: sections/1.introduction.tex
\section{Introduction}
Federated Learning (FL)~\cite{mcmahan2017communication} is a machine learning framework enabling the training of a prediction model across distributed clients while maintaining their privacy, never disclosing local data. In recent years it has had a notable resonance in the world of computer vision, with applications ranging from large-scale classification \cite{hsu2020federated} to medical imaging \cite{Guo_2021_CVPR} to domain generalization \cite{liu2021feddg} and many others \cite{li2021model,Zhang_2021_ICCV,Gong_2021_ICCV,yao2022federated}. The learning paradigm is based on communication rounds where a sub-sample of clients trains the global model independently on their local datasets, and the produced updates are later aggregated on the server-side. The heterogeneous distribution of clients' data, which is usually non-i.i.d. and unbalanced, poses a major challenge in realistic federated scenarios, leading to degraded convergence performances~\cite{zhao2018federated,hsu2019measuring,li2020federated}. Locally, the model has only access to a small portion of the data failing to generalize to the rest of the underlying distribution. That contrasts with the standard centralized training, where the learner can uniformly sample from the whole distribution. %
While many promising works in the literature focus on regularizing the local objective to align the global and local solutions, thus reducing the client drift~\cite{li2020federated,karimireddy2020scaffold,acar2021federated}, less attention has been given to the explicit optimization of the loss function for finding better minima. Several works studied the connection between the sharpness of the loss surface and model's generalization \cite{hochreiter1997flat,keskar2016large,visualloss,kleinberg2018alternative,smith2017bayesian,jastrzkebski2018relation,chen2021outperform}, and proposed effective solutions based on the minimization of the derived generalization bound \cite{yue2020salr,foret2020sharpness,kwon2021asam} or on averaging the network's parameters along the trajectory of SGD \cite{izmailov2018averaging}.

\captionsetup[subfloat]{font=scriptsize,labelformat=empty}
\begin{figure}[!t]
    \centering
    \subfloat[][\fedavg $\alpha=0$]{\includegraphics[width=.25\linewidth]{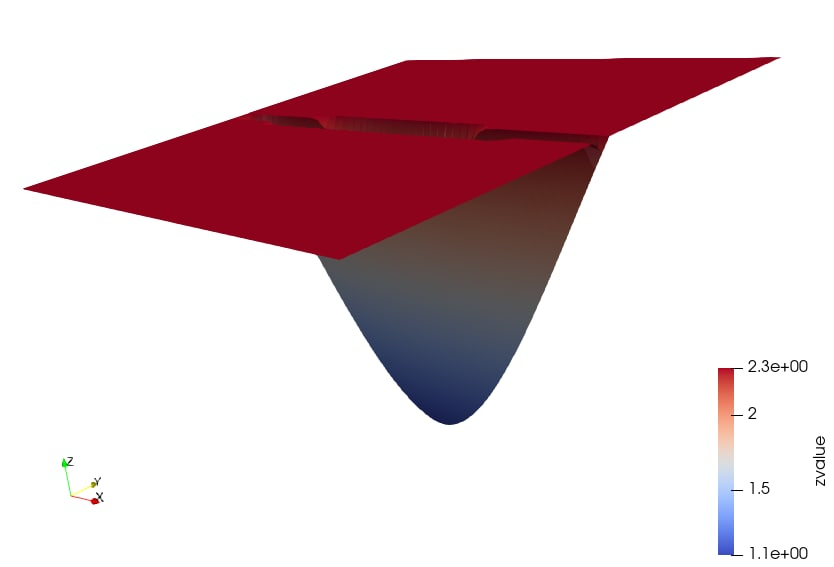}}
    \subfloat[][\fedasam $\alpha=0$]{\includegraphics[width=.25\linewidth]{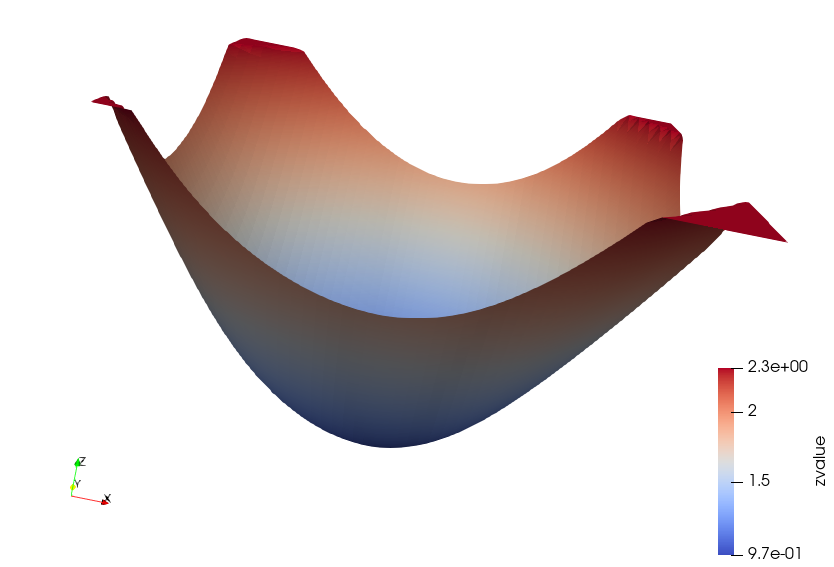}}
    \subfloat[][\fedavg $\alpha=1k$]{\includegraphics[width=.25\linewidth]{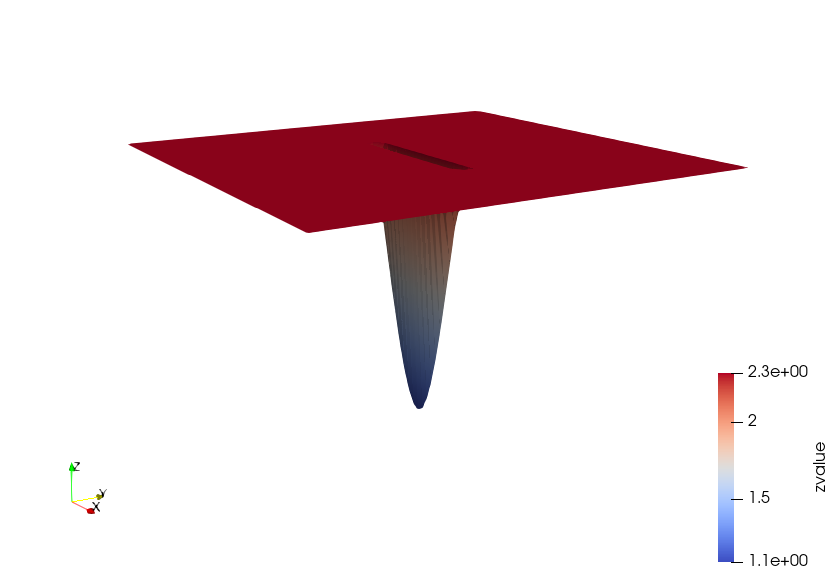}}
    \subfloat[][\fedasam $\alpha=1k$]{\includegraphics[width=.25\linewidth]{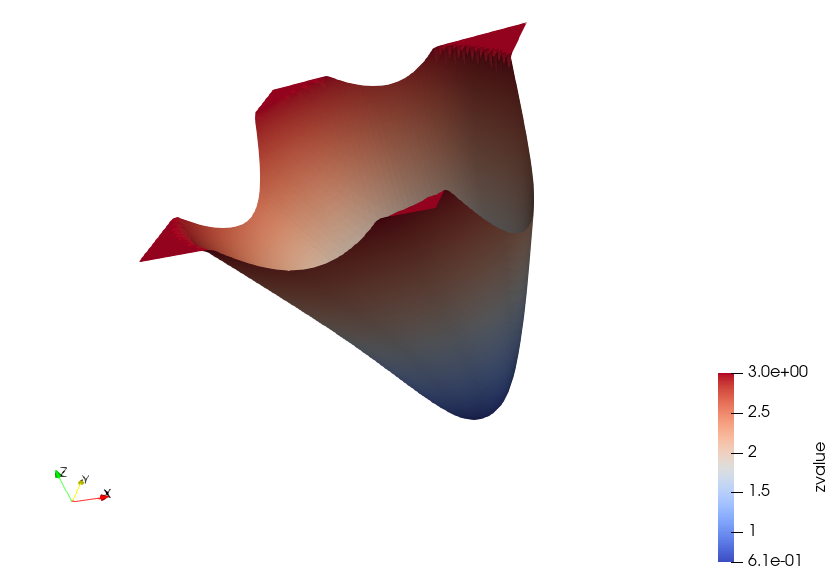}}
    \caption{\footnotesize{Cross-entropy loss landscapes of the global model in heterogeneous ($\alpha=0$) and homogeneous ($\alpha=1k$) federated scenarios on \textsc{Cifar100}. When trained with \fedavg, the global model converges towards sharp minima. The sharpness-aware optimizer \asam significantly smooths the surfaces.}}
    \label{fig:loss_landscape}
\end{figure}

In this work, we first analyze the heterogeneous federated scenario to highlight the causes behind the poor generalization of the federated algorithms. We hypothesize during local training the model overfits the current distribution, and the resulting average of the updates is strayed apart from local minima. Thus, the global model is not able to generalize to the overall underlying distribution and has a much slower convergence rate, \textit{i.e.} it needs a much larger number of rounds to reach the performance of the homogeneous setting. To speed up training and reduce the performance gap in the case of non-i.i.d. data, we look at improving the generalization ability of the model. Motivated by recent findings relating the geometry of the loss and the generalization gap \cite{keskar2016large,dziugaite2017computing,visualloss,jiang2019fantastic} and by the achievements in the field of Vision Transformers \cite{chen2021outperform}, we analyze the loss landscape in the federated scenario and find out that models converge to sharp minima (Fig.\ref{fig:loss_landscape}), hence the poor generalization. 
As a solution, we introduce methods of the current literature that explicitly look for flat minima: i) Sharpness-Aware Minimization (\sam) \cite{foret2020sharpness} and its adaptive version (\asam) \cite{kwon2021asam} on the client-side and ii) Stochastic Weight Averaging (\swa) \cite{izmailov2018averaging} on the server-side. These modifications, albeit simple, surprisingly lead to significant improvements. Their use is already effective if taken individually, but the best performance is obtained when combined. The resultant models exhibit smoother loss surfaces and improved final performance consistently across several vision tasks. %
To summarize, our main contributions are:
\begin{itemize}
    \item We analyze the behavior of models trained in heterogeneous and homogeneous federated scenarios by looking at their convergence points, loss surfaces and Hessian eigenvalues, linking the lack in generalization to sharp minima.
    \item To encourage convergence towards flatter minima, we introduce \sam and \asam in the local client-side training and \swa in the aggregation of the updates on the server-side. The resultant models show smoother loss landscapes and lower Hessian eigenvalues, with improved generalization capacities.
    \item We test our approach on multiple vision tasks, \textit{i.e.} small and large scale classification \cite{hsu2020federated}, domain generalization \cite{blanchard2011generalizing} and semantic segmentation \cite{long2015fully,chen2017deeplab}.
    \item We compare our method with strong data augmentations techniques  and state-of-the-art FL algorithms, further validating its effectiveness.
\end{itemize}

%% file: sections/2.related.tex
\section{Related Works}
We describe here the existing approaches closely related to our work. For a comprehensive analysis of the state of the art in FL, we refer to   \cite{kairouz2019advances,li2020federated_survey,zhang2021survey}.

\subsection{Statistical Heterogeneity in Federated Learning}
Federated Learning is a topic in continuous growth and evolution. Aiming at a real-world scenario, the non-i.i.d. and unbalanced distribution of users' data poses a significant challenge. The \textit{statistical heterogeneity} of local datasets leads to unstable and slow convergence, suboptimal performance and poor generalization of the global model \cite{zhao2018federated,hsu2019measuring,hsu2020federated}. \fedavg \cite{mcmahan2017communication} defines the standard optimization method and is based on multiple local SGD \cite{ruder2016overview} steps per round. The server-side aggregation is a weighted average of the clients' updates. This simple approach is effective in homogeneous scenarios. Still, it fails to achieve comparable performance against non-i.i.d. data due to local models straying from each other and leading the central model away from the global optimum \cite{karimireddy2020scaffold}. To mitigate the effect of the \textit{client drift}, many works enforce regularization in local optimization so that the local model is not led too far apart from the global one~\cite{li2020federated,karimireddy2020scaffold,hsu2020federated,acar2021federated,li2021model}. Indeed, averaging models/gradients collected from clients having access to a limited subset of tasks may translate into oscillations of the global model and suboptimal performance on the global distribution \cite{lin2020ensemble}. Therefore, other lines of research look at improving the aggregation stage using server-side momentum~\cite{hsu2019measuring} and adaptive optimizers \cite{reddi2020adaptive}, or aggregating task-specific parameters~\cite{smith2017bayesian,briggs2020federated,caldarola2021cluster}.

In this work, we attempt to explain the behavior of the model in federated scenarios by looking at the loss surface and convergence minima, which is, in our opinion, a fundamental perspective to fully understand the reasons behind the degradation of heterogeneous performance relative to centralized and homogeneous settings. To this end, \textit{we focus on explicitly seeking parameters in uniformly low-loss neighborhoods, without any additional communication cost}. By encouraging local convergence towards flatter minima, we show that the generalization capacity of the global model is consequently improved. Moreover, thanks to the cyclical average of stochastic weights - accumulated along the trajectory of SGD during rounds on the server-side - broader regions of the weight space are explored, and wider optima are reached. Referring to the terminology introduced by \cite{yuan2021we}, we aim at bridging the \textit{participation gap} introduced by unseen clients distributions.
Concurrently, \cite{fedsam_icml2022} provide a theoretical analysis of SAM in FL, matching the convergence rates of the existing methods. Unlike our work, they do not explicitly focus on the issue of statistical heterogeneity in vision tasks.

\subsection{Real-world Vision Scenarios in Federated Learning} 
Research on FL has mainly focused on algorithmic aspects, often overlooking its application to real scenarios and vision tasks. Here, we perform an analysis of the following real-world settings.

\noindent{\textit{\textbf{{Large-scale Classification.}}}} Synthetic federated datasets for classification tasks are usually limited in size and do not offer a faithful representation of reality in the data distribution across clients \cite{hsu2020federated}. \cite{hsu2020federated} addresses such issue by adapting the large-scale Google Landmarks v2 \cite{weyand2020google} to the federated context, using authorship information. We employ the resulting \textit{Landmarks-User-160k} in our experiments.

\noindent{\textit{\textbf{{Semantic Segmentation.}}}} A crucial task for real-world applications \cite{garcia2017review,ouahabi2021deep}, \eg autonomous driving \cite{siam2018comparative,tavera2022pixel}, is Semantic Segmentation (SS), which assigns each image pixel to a known category. Most studies of SS in FL focus on medical imaging applications and propose ad hoc techniques to safeguard the patients' privacy \cite{sheller2018multi,li2019privacy,yi2020net,bercea2021feddis}. Differently, \cite{michieli2021prototype} focuses on object segmentation using prototypical representations. %
A recently studied application is FL in autonomous driving, motivated by the large amount of privacy-protected data collected by self-driving cars: the authors of \cite{fantauzzo2022feddrive} propose a new benchmark for analyzing such a scenario, FedDrive. None of those works study the relation between loss landscape and convergence minima of the proposed solution. We apply our approach to the FedDrive benchmark and prove its efficacy in addressing the federated SS task.

\noindent{\textit{\textbf{{Domain Generalization.}}}} When it comes to image data collected from devices around the world, it is realistic to assume there may be different \textit{domains} resulting from the several acquisition devices, light, weather conditions, noise, or viewpoints. With the rising development of FL and the privacy concerns, the problem of Domain Generalization (DG) \cite{blanchard2011generalizing} in a federated setting becomes crucial. DG aims to learn a domain-agnostic model capable of satisfying performances on unseen domains, and its application to federated scenarios is still poorly studied. For instance, \cite{liu2021feddg,tian2021privacy} focus on domain shifts deriving from equipment in the medical field, while \cite{fantauzzo2022feddrive} analyzes the effects of changing landscapes and weather conditions in the setting of autonomous driving. We show that our approach improves generalization to unseen domains both in classification and SS tasks. 

\subsection{Flat Minima and Generalization}
\label{sec:related_min}
To understand neural networks' generalization, several theoretical and empirical studies analyze its relationship with the geometry of the loss surface \cite{hochreiter1997flat,keskar2016large,dziugaite2017computing,visualloss,jiang2019fantastic}, connecting sharp minima with poor generalization. \quotes{\textit{Flatness}}\cite{hochreiter1997flat} is defined as the dimension of the region connected around the minimum in which the training loss remains low. Interestingly, it has been shown \cite{jiang2019fantastic} that sharpness-based measures highly correlate with generalization performance. The above studies lead to the introduction of Sharpness-Aware Minimization (\sam) \cite{foret2020sharpness} which explicitly seeks flatter minima and smoother loss surfaces through a simultaneous minimization of loss sharpness and value during training. As highlighted by \cite{kwon2021asam}, \sam is sensitive to parameter re-scaling, weakening the connection between loss sharpness and generalization gap. \asam \cite{kwon2021asam} solves such issue introducing the concept of adaptive sharpness. Encouraged by their effectiveness across a variety of architectures and tasks\cite{chen2021outperform,bahri2021sharpness}, we ask whether \sam and \asam can improve generalization in FL as well and find it effective even in the most difficult scenarios. In addition, \cite{garipov2018loss,draxler2018essentially} show that local optima found by SGD are connected through a path of near constant loss and that ensambling those points in the weight space leads to high performing networks. Building upon these insights, \cite{izmailov2018averaging} proposes to average the points traversed by SGD to improve generalization and indeed show the model converges towards wider optima. We modify this approach for FL and use it to cyclically ensemble the models obtained with \fedavg on the server side.

%% file: sections/3.heterogeneous_fl.tex
\section{Behind the Curtain of Heterogeneous FL}
\subsection{Federated Learning: Overview}
The standard federated framework is based on a central server exchanging messages with $K$ distributed clients. Each device $k$ has access to a privacy-protected dataset $\mathcal{D}_k$ made of $N_k$ images belonging to the input space $\mathcal{X}$. The goal is to learn a global model $f_{\theta}$ parametrized by $\theta\in \mathcal{W} \subseteq \mathbb{R}^d$, where $f_{\theta}: \mathcal{X} \rightarrow \mathcal{Y}$ when solving the classification task and $f_{\theta}: \mathcal{X} \rightarrow \mathcal{Y}^{N_p}$ in semantic segmentation, with $\mathcal{Y}$ being the output space and $N_p$ the total number of pixels of each image. We assume the structure of $\theta$ to be identical across all devices. The learning procedure spans over $T$ communications rounds, during which a subset of clients $\mathcal{C}$ receives the current model parameters $\theta^t$ with $t\in[T]$ and trains it on  $\mathcal{D}_k \: \forall k\in \mathcal{C}$, minimizing a local loss function $\mathcal{L}_k(\theta^t): \mathcal{W}\times \mathcal{X} \times \mathcal{Y} \rightarrow \mathbb{R}_+$. 
In \fedavg \cite{mcmahan2017communication}, the global model is updated as a weighted average of the clients' updates $\theta^t_k$, aiming at solving the global objective
    $\arg\min_{\theta\in \mathbb{R}^d} \frac{1}{N} \sum_{k\in \mathcal{C}} N_k  \mathcal{L}_k(\theta)$,
with $N = \sum_{k\in \mathcal{C}} N_k$ being the total training images. In particular, from the generalization perspective - defined $\mathcal{D} \triangleq \bigcup_{k\in[K]} \mathcal{D}_k$ the overall clients' data, $\mathfrak{D}$ its distribution and $\mathcal{L}_\mathcal{D} = \nicefrac{1}{\sum_k N_k} \sum_{k\in [K]} N_k \mathcal{L}_k (\theta)$ the training loss - we aim at learning a model having low population loss $\mathcal{L}_\mathfrak{D}(\theta) \triangleq \mathbb{E}_{(x,y)\sim \mathfrak{D}}\big[\mathbb{E}_\mathcal{D}[\mathcal{L}_k(y, f(x, \theta))]\big]$ \cite{yuan2021we}. The difference between the population and training losses defines the \textit{generalization gap}, \textit{i.e.} the ability of the model to generalize to unseen data \cite{foret2020sharpness}.

In realistic scenarios, given two clients $i$ and $j$, $\mathcal{D}_i$ likely follows a different distribution than $\mathcal{D}_j$, \textit{i.e.} $\mathfrak{D}_i \neq \mathfrak{D}_j$, and the loss $\mathcal{L}_i(\theta) \: \forall i\in [K]$ is typically non-convex in $\theta$. The loss landscape  comprehends a multiplicity of local minima leading to models with different generalization performance, \textit{i.e.} significantly different values of $\mathcal{L}_\mathfrak{D}(\theta)$ \cite{foret2020sharpness}. Moreover, at each round, the model is likely not to see the entire distribution, further widening the generalization gap \cite{hendrycks2019benchmarking,hendrycks2021many}.

\subsection{Where Heterogeneous FL Fails at Generalizing}
\label{sec:het_fl}
\captionsetup[subfloat]{font=scriptsize,labelformat=parens}
\begin{figure}[!t]
    \centering
    \subfloat[][]{\includegraphics[width=.25\linewidth]{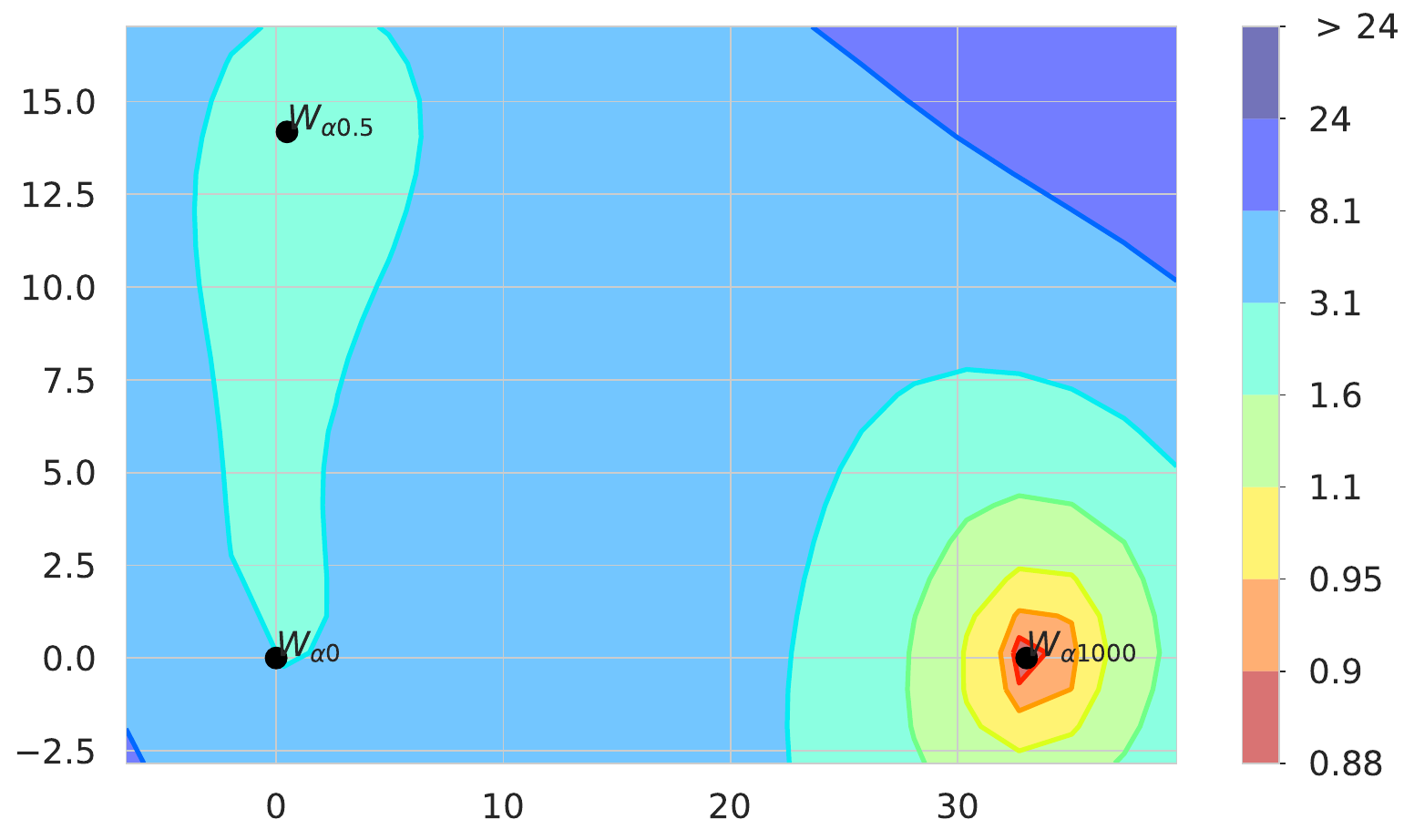}}
    \subfloat[][]{\includegraphics[width=.25\linewidth]{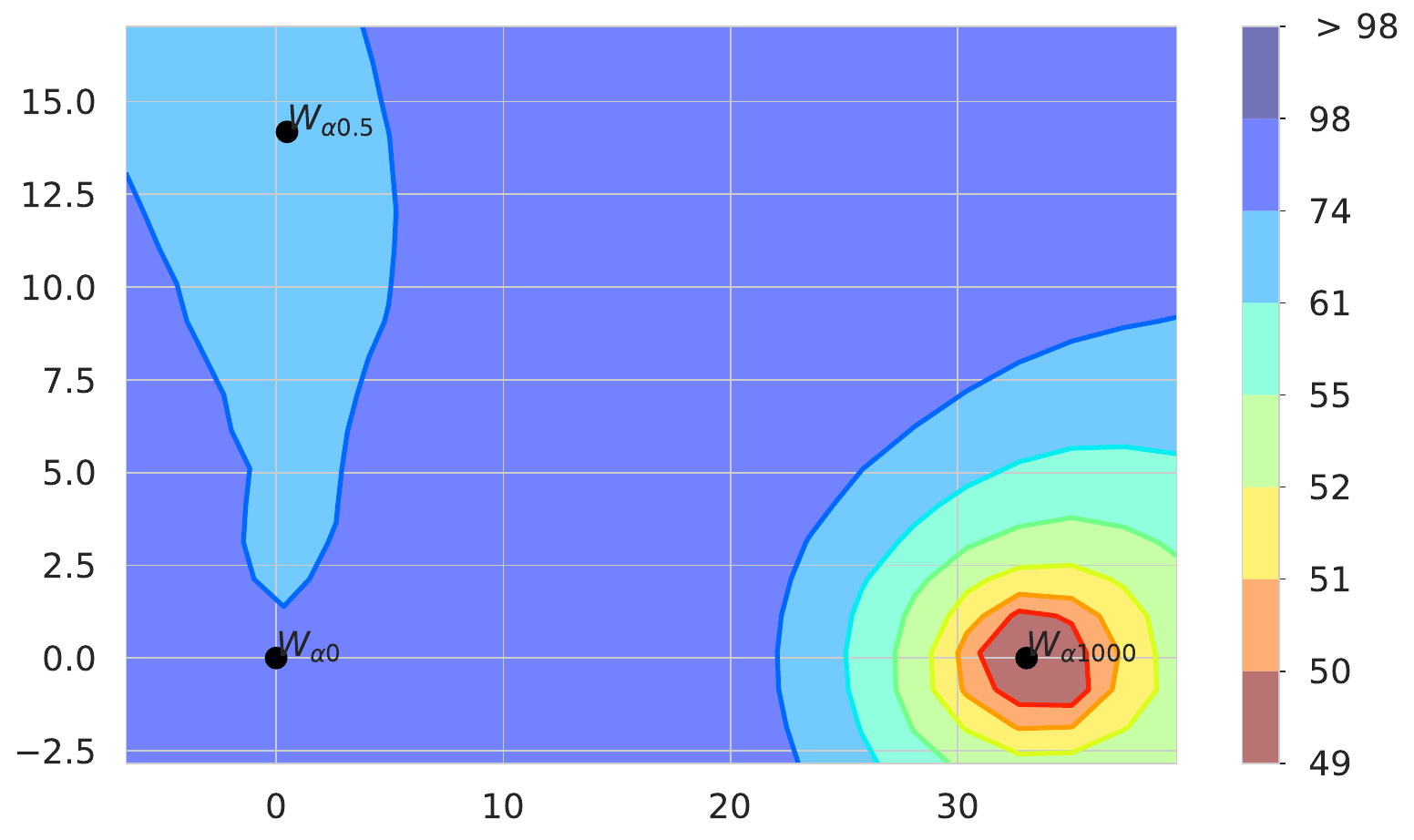}}
    \subfloat[][]{\includegraphics[width=.25\linewidth]{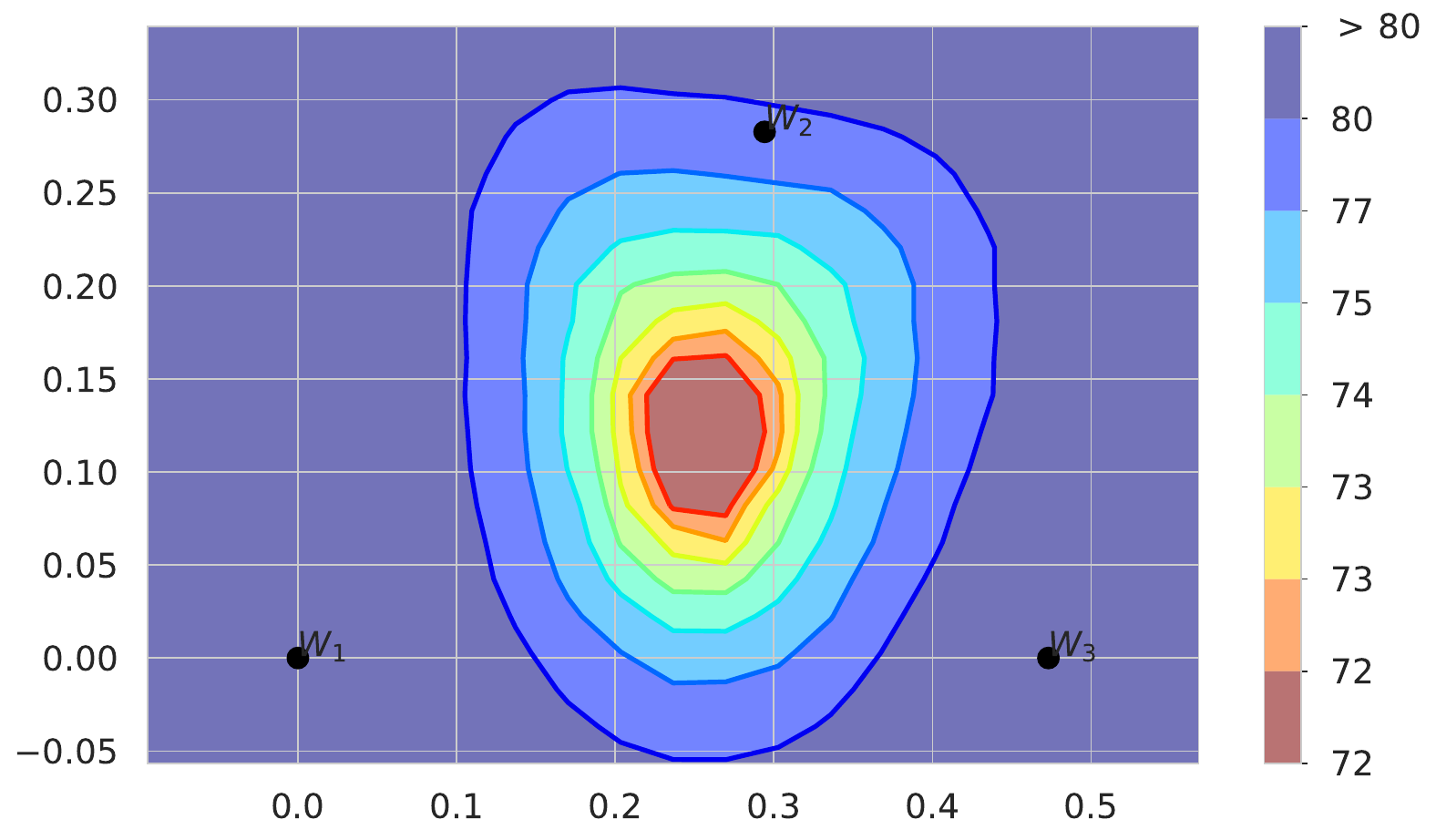}}
    \subfloat[][]{\includegraphics[width=.25\linewidth]{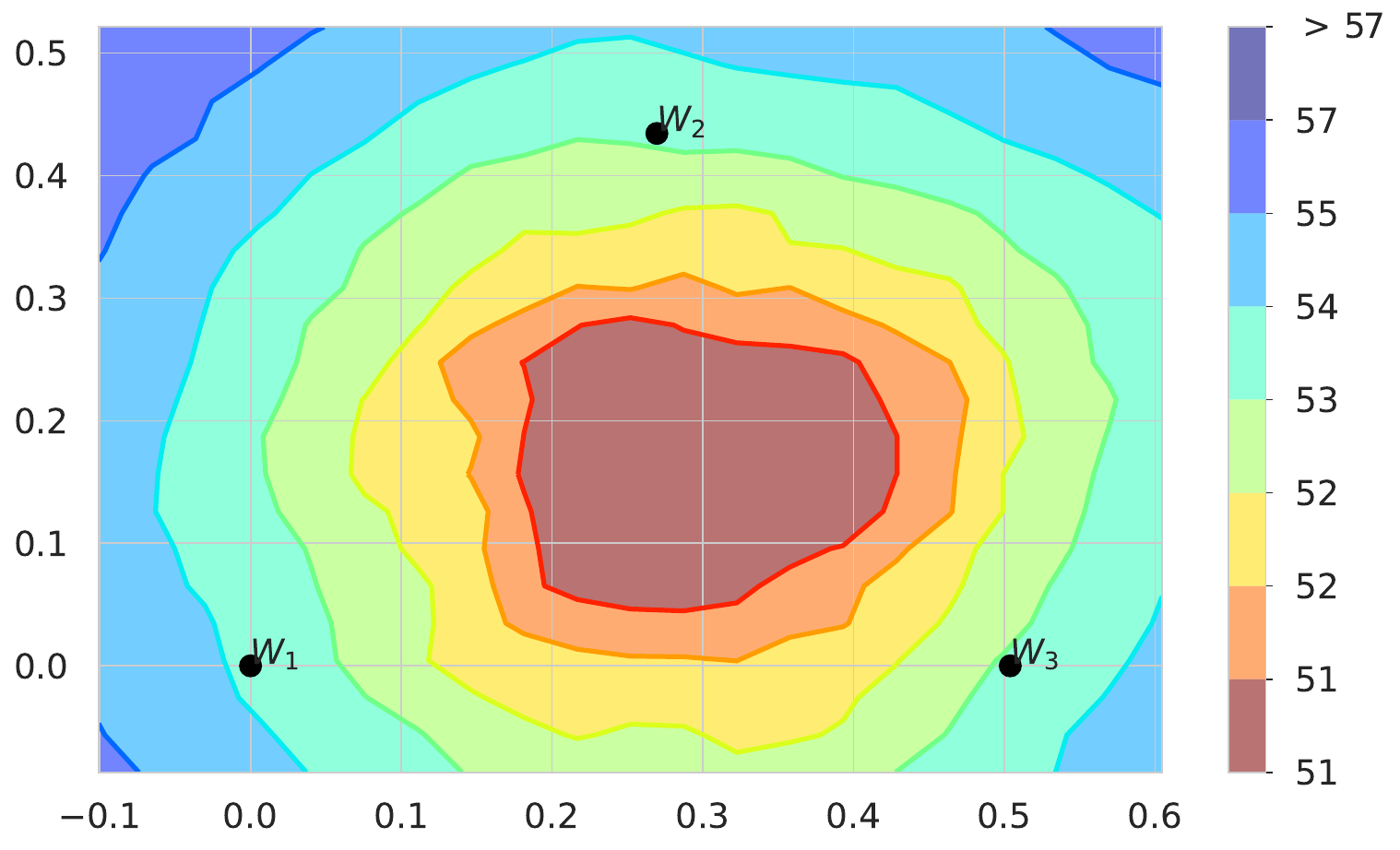}}
    \caption{\footnotesize{\textbf{Left}: CNN convergence points in distinct federated scenarios with $\alpha\in[0,0.5,1k]$ on \textsc{Cifar100}. Please refer to Appendix~\ref{app:exps} for implementation details. \textbf{(a)} Train loss surface showing the weights obtained at convergence. \textbf{(b)} Test error surface of the same models. \textbf{Right}: Test error surfaces computed on \textsc{Cifar100} using three distinct local models after training. \textbf{(c)} When $\alpha=0$, the local models are not able to generalize to the overall data distribution, being too specialized on the local data. \textbf{(d)} When $\alpha=1k$, the resulting models are connected through a low-loss region.}}
    \label{fig:fedavg_convergence}
\end{figure}
In order to fully understand the behavior of a model trained in a heterogeneous federated scenario, we perform a thorough empirical analysis from different perspectives. Our experimental setup replicates that proposed by \cite{hsu2020federated} both as regards the dataset and the network. The \textsc{Cifar100} dataset \cite{krizhevsky2009learning}, widely used as benchmark in FL, is split between $100$ clients, following a Dirichlet distribution with concentration parameter $\alpha$. To replicate a heterogeneous scenario, we choose $\alpha\in\{0,0.5\}$, while $\alpha$ is set to $1000$ for the homogeneous one. The model is trained over $20k$ rounds. Fore more details, please refer to Appendix~\ref{app:exps}. %

\begin{figure}[!t]
\centering
    \subfloat[][]{\includegraphics[width=.25\linewidth]{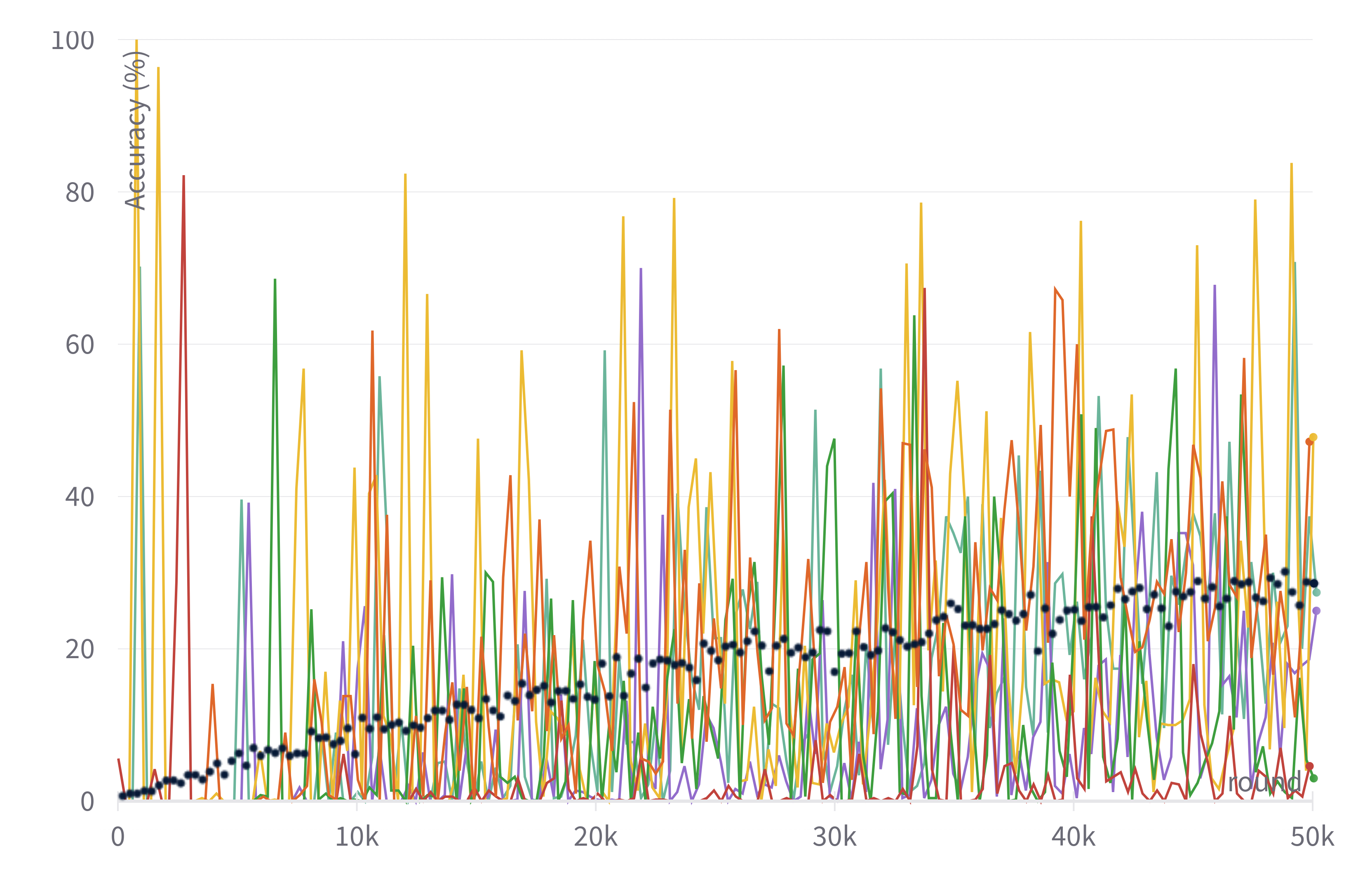}}
    \subfloat[][]{\includegraphics[width=.25\linewidth]{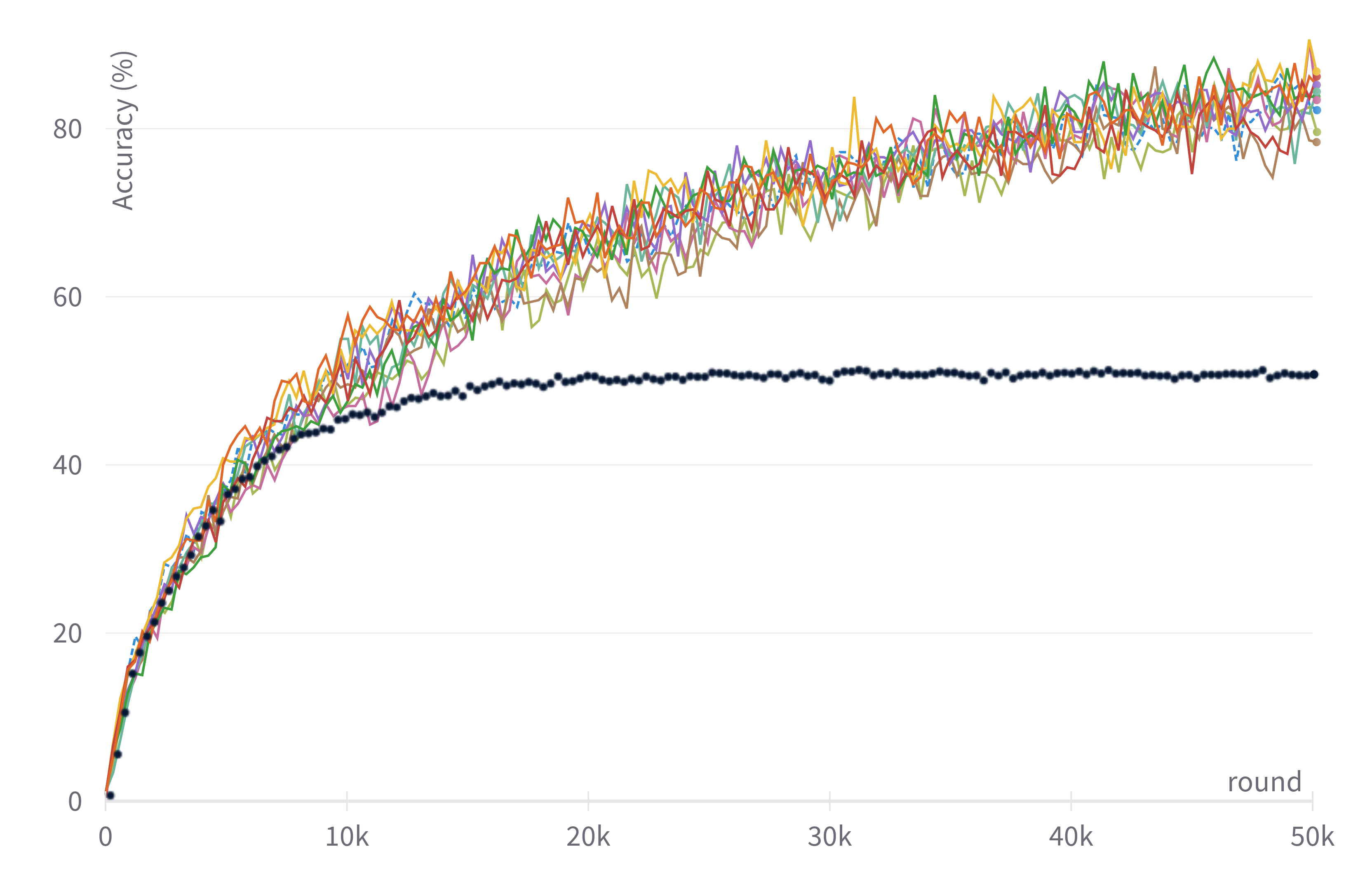}}
    \subfloat[][]{\includegraphics[width=.5\linewidth]{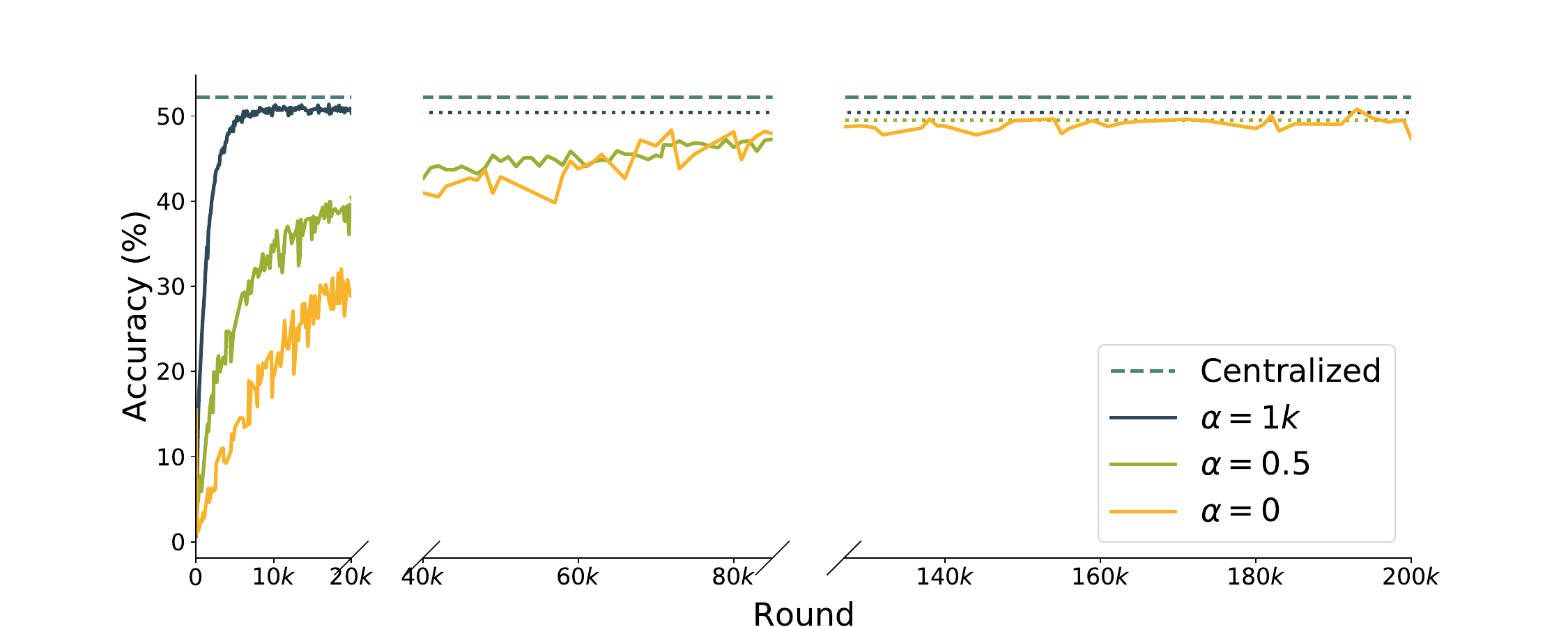}}
    \caption{\footnotesize{\textsc{Cifar100} Accuracy trends. \textbf{Left}: Global model on local distributions with \textbf{(a)} $\alpha=0$ and \textbf{(b)} $1k$ @ $20k$ rounds. Each color represents a local distribution (\textit{i.e.} one class for $\alpha=0$). \textbf{(c)}: $\alpha\in\{0,0.5,1k\}$ with necessary rounds to reach convergence.}}
    \label{fig:local_behavior}
\end{figure}

\noindent{\textit{\textbf{Model Behavior in Heterogeneous and Homogeneous Scenarios.}}}
In Fig. \ref{fig:local_behavior}, we compare the training trends in centralized, homogeneous and heterogeneous federated settings: in the latter, not only is the trend much noisier and more unstable, but the performance gap is considerable. Consequently, we question the causes of such behavior. 
First of all, we wonder if the heterogeneous distribution of the data totally inhibits the model from achieving comparable performances: we find it is only a matter of rounds, \textit{i.e.} with a much larger round budget - 10 times larger in our case - the model reaches convergence (Fig. \ref{fig:local_behavior}). So it becomes obvious the training is somehow slowed down and there is room for improvement. This hypothesis is further validated by the convergence points of the models trained in different settings (Fig. \ref{fig:fedavg_convergence}): when $\alpha=1k$ a low-loss region is reached at the end of training, while the same does not happen with lower values of $\alpha$, meaning that local minima are still to be found. Moreover, the shift between the train and test surfaces suggests us the model trained in the heterogeneous setting ($\alpha=0$) is unable to generalize well to unseen data, finding itself in a high-loss region \cite{izmailov2018averaging}. By analyzing the model behavior, we discover that shifts in client data distribution lead to numerous fluctuations in learning, \textit{i.e.} at each round the model focuses on a subset of the just seen tasks and is unable to generalize to the previously learned ones. This phenomenon is also known as \textit{catastrophic interference} of neural networks \cite{kirkpatrick2017overcoming} and is typical of the world of multitask learning \cite{caruana1997multitask,smith2017federated}. Fig. \ref{fig:local_behavior} highlights this by comparing the accuracy of the global model on the clients' data and the test set when $\alpha=0$ and $\alpha=1k$. In the first case, at each round the model achieves very high performances on one class but forgets about the others and this behavior is only slightly attenuated as the training continues. In the homogeneous scenario, on the other hand, the model behaves very similarly on each client and convergence is easily reached, giving way to overfitting as the number of rounds increases. 

We analyze the clients' local training for further insights from the characteristics of the updated models. By plotting the position of the weights in the loss landscape after training, we find the models easily overfit the local data distribution (Fig. \ref{fig:fedavg_convergence}): when tested on the test set, the clients' updates are positioned in very high-error regions and as a result the global model moves away from the minimum, meaning the clients specialize too much on their own data and are not able to generalize to the overall underlying distribution. Moreover, Fig. \ref{fig:fedavg_convergence} highlights another relevant issue: models trained on homogeneous distributions are connected through a path of low error and can therefore be ensambled to obtain a more meaningful representation \cite{garipov2018loss}, but the same does not hold when $\alpha=0$, where the models are situated in different loss-value regions. Therefore, \fedavg averages models that are too far apart to lead to a meaningful result.

\noindent{\textit{\textbf{{Federated Training Converges to Sharp Minima.}}}}
\captionsetup[subfloat]{font=scriptsize,labelformat=empty}
\begin{table}[!t]
\begin{minipage}{.5\columnwidth}
    \centering
    \subfloat[][\fedavg $\alpha=0$]{\includegraphics[width=.5\linewidth]{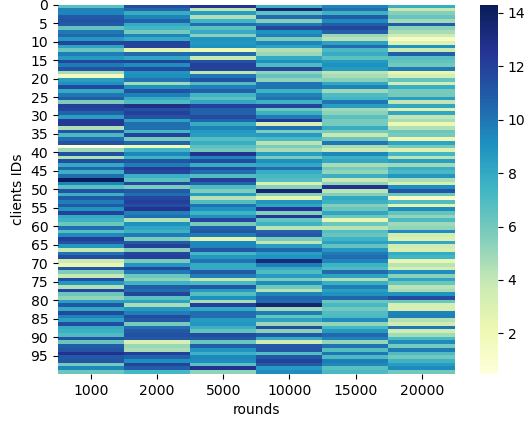}}
    \subfloat[][\asam $\alpha=0$]{\includegraphics[width=.5\linewidth]{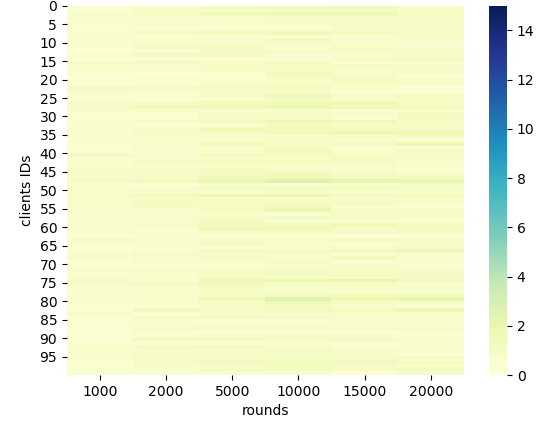}}\\
    \subfloat[][\fedavg $\alpha=1k$]{\includegraphics[width=.5\linewidth]{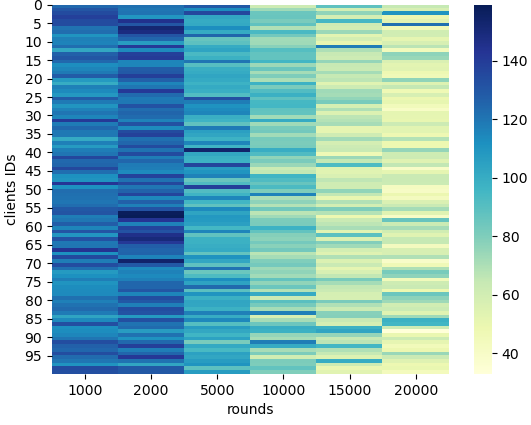}}
    \subfloat[][\asam $\alpha=1k$]{\includegraphics[width=.5\linewidth]{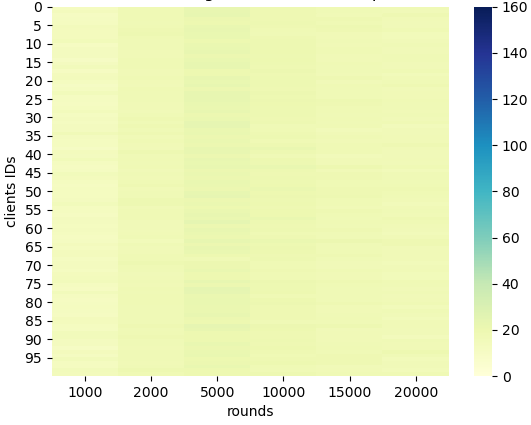}}
    \captionof{figure}{\footnotesize{$\lambda_{max}^k$ for each client $k$ as rounds\\pass}}
    \label{fig:clients_eigs}
\end{minipage}
	\begin{minipage}{0.5\columnwidth}
	\centering
        \tiny
        \caption{\textsc{Cifar100} Hessian eigenvalues.}\label{tab:eigs}
        \begin{tabular}{lcccc}
        \toprule\noalign{\smallskip}
         \multirow{2}{*}{Algorithm} & \multicolumn{2}{c}{$\lambda_{max}$} &\multicolumn{2}{c}{${\nicefrac{\lambda_{max}}{\lambda_5}}$}\\
        \cmidrule(l){2-3} \cmidrule(l){4-5}
        & $\alpha=0$ & $\alpha=1k$ & $\alpha=0$& $\alpha=1k$\\
        \noalign{\smallskip}
        \hline
        \noalign{\smallskip}
        \fedavg \texttt{E=1} & 93.46&106.14& 2.00& 1.31\\
        \fedavg \texttt{E=2} & 110.62  &118.35 &2.32&1.30\\
        \fedsam & 70.29 & 51.28 &\textbf{1.79} &1.48\\
        \fedasam & \textbf{30.11} & \underline{\textbf{20.19}}& 1.80&\underline{\textbf{1.27}}\\\midrule
        \fedavgswa & 97.24 &  120.02& 1.49&1.39\\
        \fedsamswa & 73.16 & 54.20 & 1.56&1.61\\
        \fedasamswa & \underline{\textbf{24.57}}& \textbf{20.49}& \underline{\textbf{1.51}}&\textbf{1.30}\\
        \bottomrule
        \end{tabular}
        \subfloat[][]{\includegraphics[width=.5\linewidth]{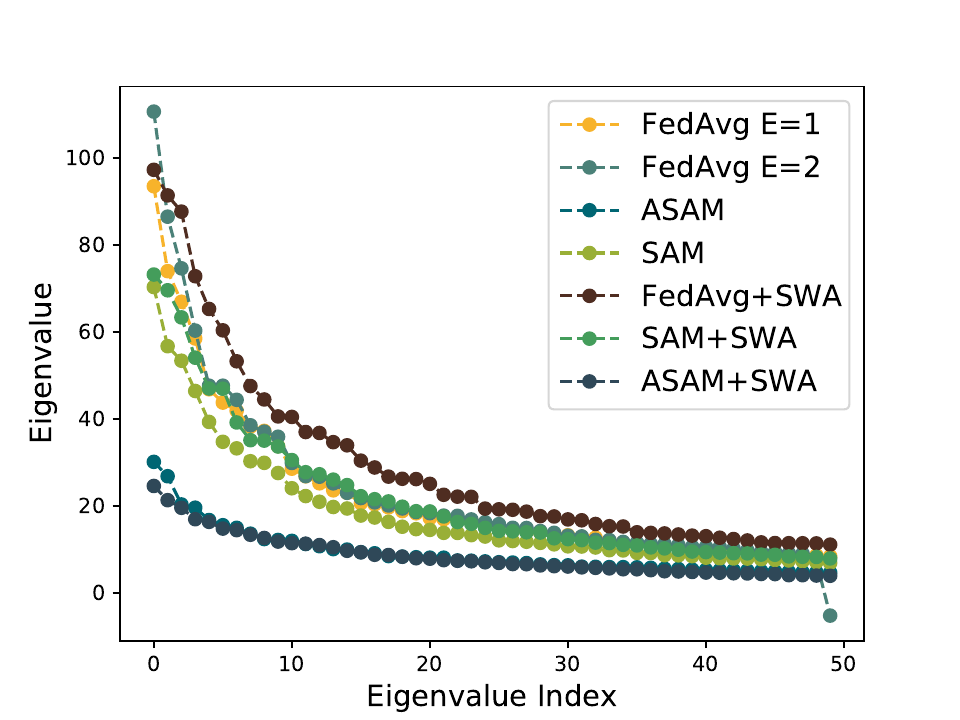}}
        \subfloat[][]{\includegraphics[width=.5\linewidth]{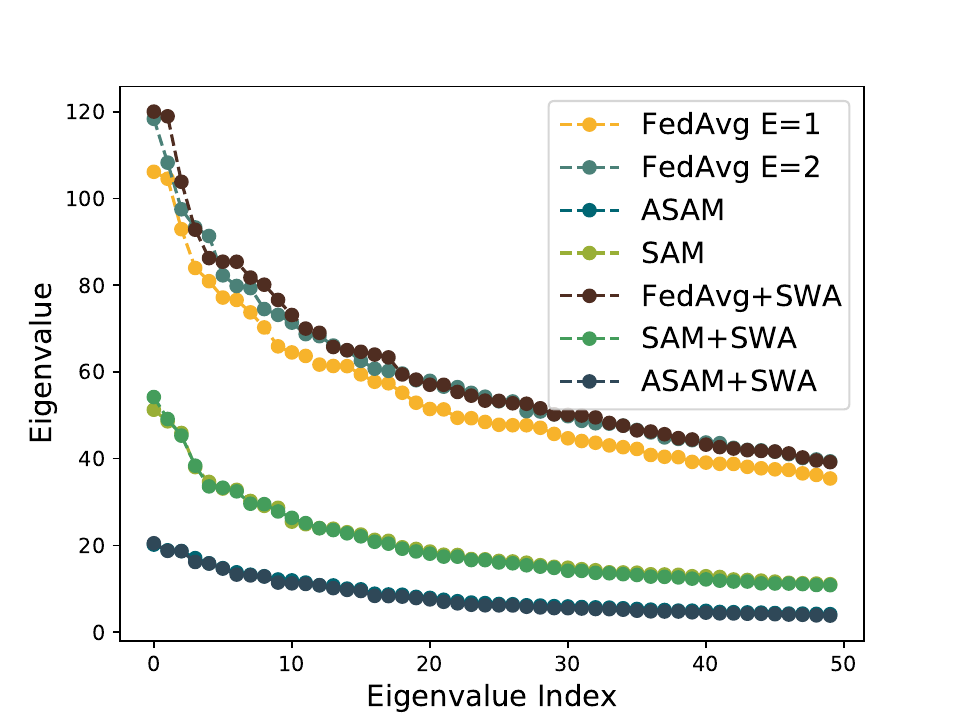}}
        \label{fig:global_eigs}
        \captionof{figure}{Hessian eigenspectra of the global model with $\alpha\in\{0,1k\}$}
	\end{minipage}\hfill
\end{table}
Many works tried to account for this difficulty arising in federated scenarios by enforcing regularization in local optimization not to lead the local model too far apart from the global one \cite{li2020federated,karimireddy2020scaffold,hsu2020federated,acar2021federated,li2021model}, or by using momentum on the server-side \cite{hsu2019measuring}, or learning task-specific parameters keeping distinct models on the server-side \cite{smith2017bayesian,briggs2020federated,caldarola2021cluster}. To the best of our knowledge, this is the first work addressing such behavior by looking at the loss landscape. Inspired by a recent trend in Deep Learning connecting the geometry of the loss and the generalization gap \cite{keskar2016large,dziugaite2017computing,visualloss,jiang2019fantastic,kwon2021asam,izmailov2018averaging}, we investigate the geometry of the loss surface of models trained in non-i.i.d. scenarios with the intention of understanding whether sharp minima may cause the lack of generalization in FL. Following \cite{visualloss}, we plot the loss surfaces obtained with models trained in a heterogeneous and in a homogeneous scenario (Fig. \ref{fig:loss_landscape}) showing that both converge to sharp regions, providing a plausible explanation for the highlighted lack of generalization. Additionally, \cite{keskar2016large} characterizes flatness through the eigenvalues of the Hessian: the dominant eigenvalue $\lambda_{max}$ evaluates the worst-case landscape curvature, \textit{i.e.} the larger $\lambda_{max}$ the greater the change in loss in that direction and the steeper the minimum. Hence, we compute the Hessian eigenspectrum (first $50$ eigenvalues) using the power iteration mode and analyze it both from the global and local perspectives (Fig. \ref{fig:clients_eigs},{\color{red}{5}}). Table \ref{tab:eigs} reports the values of $\lambda_{max}$ and the ratio  $\nicefrac{\lambda_{max}}{\lambda_5}$, commonly used as a proxy for sharpness \cite{jastrzebski2020break}, as the heterogeneity varies. As expected, $\lambda_{max}$ is large in all settings when using \fedavg, implying that such method leads the model towards sharp minima regardless of the data distribution, confirming what was noted in the loss landscapes. 
As for the client-side analysis, we compute the value of $\lambda_{max}^k$ using the locally updated parameters $\theta^t_k$ on the $k$-th device's data $\mathcal{D}_k\: \forall t\in[T]$. Comparing the i.i.d. and non-i.i.d. settings, we note i) the local values of $\lambda_{max}$ are much lower if $\alpha=0$, \ie the clients locally reach wide minima (low Hessian maximum eigenvalue, $\lambda_{max}^k\leq 14$) due to the simplicity of the learned task, \ie a narrow subset of the classes, but the average of the distinct updates drives the model towards sharper minima (high Hessian eigenvalues of the global model, $\lambda_{max}\simeq94$). ii) When $\alpha\in\{0.5,1k\}$, $\lambda_{max}$ decreases as the rounds pass, \ie the global model is moving towards regions with lower curvature, while this is not as evident in the heterogeneous setting. Motivated by these results, we believe that introducing an explicit search for flatter minima can help the model generalize.
\captionsetup[subfloat]{font=tiny,labelformat=empty}

%% file: sections/4.method.tex
\section{Seeking Flat Minima in Federated Learning}
\input{tables/algo2}
Common first-order optimizers (\textit{e.g.} SGD \cite{ruder2016overview}, Adam \cite{kingma2014adam})  are usually non-robust to unseen data distributions \cite{chen2021outperform}, since they only aim at minimizing the training loss $\mathcal{L}_\mathcal{D}$, without looking at higher-order information correlating with generalization (\textit{e.g.} curvature). The federated scenario exacerbates such behavior due to its inherent statistical heterogeneity, resulting in sharp minima and poor generalization. We hypothesize that encouraging the local model to converge towards flatter neighborhoods may help bridging the generalization gap. To this end, we introduce sharpness-aware minimizers, namely \sam \cite{foret2020sharpness} and \asam \cite{kwon2021asam}, on the client-side during local training, and Stochastic Weight Averaging \cite{izmailov2018averaging} on the server-side after the aggregation, adapting the scenario of \cite{izmailov2018averaging} to FL. By minimizing the sharpness of the loss surface and the generalization gap, the local models are more robust towards unseen data distributions and, when averaged, build a more solid central model. Defined the \textit{sharpness} of a training loss $\mathcal{L}_{\mathcal{D}}$ as $\max_{||\epsilon||_p \leq \rho}\mathcal{L}_\mathcal{D}(\theta+\epsilon) - \mathcal{L}_\mathcal{D}(\theta)$, with $\rho$ being the neighborhood size and $p\in[1,\infty)$, \sam aims at minimizing it by solving $\min_{\theta\in\mathbb{R}^d}\max_{||\epsilon||_p \leq \rho}  \mathcal{L}_\mathcal{D}(\theta+\epsilon) + \lambda ||\theta||_2^2$. \swa averages weights proposed by SGD, while using a learning rate schedule to explore regions of the weight space corresponding to high performing networks. For a detailed explanation of \sam, \asam and \swa we refer the reader to Appendix~\ref{app:background}. Algorithm \ref{alg:fl_asam} sums up the details of our approach. %

%% file: tables/algo2.tex
\definecolor{mygreen}{RGB}{77,175,74}
\definecolor{myblue}{RGB}{55,126,184}
\definecolor{skyblue}{RGB}{117,187,253}
\definecolor{myred}{RGB}{228,26,28}
\definecolor{blue_palette}{HTML}{008374}
\definecolor{yellow_palette}{HTML}{f7b32b} %
\newcommand{\sambox}[1]{\colorbox{blue_palette!30}{#1}}
\newcommand{\swabox}[1]{\colorbox{yellow_palette!65}{#1}}

\begin{algorithm}[!t]
\footnotesize
\caption{
 \sambox{\sam/\asam} and \swabox{\swa} applied to \texttt{FedAvg}
} 
\label{alg:fl_asam}
\tiny
\begin{algorithmic}[1]
   \Require Initial random model $f_{\theta}^0$, $K$ clients, $T$ rounds, learning rates $\gamma_1,\gamma_2$, neighborhood size $\rho>0$, $\eta>0$, batch size $|\mathcal{B}|$, local epochs $E$, cycle length $c$
   \For{each round $t=0$ {\bfseries to} $T-1$}
   \If{$t = 0.75 * T$} \Comment{Apply \swa from 75\% of training onwards}
    \State \swabox{$\theta_\text{\swa} \leftarrow \theta^t$} \Comment{Initialize \swa model}
   \EndIf
   \If{$t \geq 0.75 * T$} 
    \State $\gamma = \gamma(t)$ \Comment{Compute LR for the round (Eq.~\ref{math:lr_swa}  in Appendix)} %
   \EndIf
       \State Subsample a set $\mathcal{C}$ of clients 
       \For{each client $k$ in $\mathcal{C}$ in parallel} \Comment{Iterate over subset $\mathcal{C}$ of clients}
       \State $\theta_{k,0}^{t+1} \leftarrow \theta^t$
           \For{$e=0$ {\bfseries to} $E-1$}
               \For{$i=0$ {\bfseries to} $\nicefrac{N_k}{|\mathcal{B}|}$}
                    \State Compute gradient $\nabla_{\theta}  \mathcal{L}_\mathcal{B}(\theta^{t+1}_{k,i})$ on batch $\mathcal{B}$ from $\mathcal{D}_{k}$ 
                    \State {\sambox{Compute $\hat{\epsilon}\big(\theta^{t+1}_{k,i}\big) = \rho \nabla_{\theta}\mathcal{L}_\mathcal{B}\big(\theta^{t+1}_{k,i}\big)\Big/\big|\big|\nabla_{\theta}\mathcal{L}_\mathcal{B}\big(\theta^{t+1}_{k,i}\big)\big|\big|_2 =: \hat{\epsilon}(\theta)$}} \Comment{Solve local maximization (Eq.~\ref{math:approx_sam})} %
                    \State $\theta^{t+1}_{k,i+1} \leftarrow \theta^{t+1}_{k,i} - \gamma\Big(${\sambox{ $\nabla_{\theta}\mathcal{L}_\mathcal{B}(\theta^{t+1}_{k,i})\Big|_{\theta + \hat{\epsilon}(\theta)}$}}\Big) \Comment{Local update with sharpness-aware gradient (Eq.~\ref{math:sam_update})} %
               \EndFor
           \EndFor
           \State Send $\theta^{t+1}_k$ to the server
        \EndFor
        \State $\theta^{t+1} \leftarrow \frac{1}{\sum_{k\in \mathcal{C}} N_k} \sum_{k\in \mathcal{C}} N_k \theta^{t+1}_k$ \Comment{\fedavg}
        \If{$t \geq 0.75 * T$ \textbf{and} mod$(t,c)=0$} \Comment{End of cycle}
            \State {\swabox{\strut $n_\text{models} \leftarrow \nicefrac{t}{c}$}}
            \State {\swabox{\strut $\theta_\text{\swa} \leftarrow \frac{\theta_\text{\swa} \cdot n_\text{models} + \theta^{t+1}}{n_\text{models}+1}$}} \Comment{Update \swa average (Eq.~\ref{math:swa})} %
        \EndIf
   \EndFor
\end{algorithmic}
\end{algorithm}

%% file: sections/5.experiments.tex
\section{Experiments}
\input{tables/cifar}
In this Section, we show the effectiveness of \sam, \asam and \swa in federated scenarios when addressing tasks of image classification (Sec. \ref{sec:cifar}), large-scale classification, SS and DG (Sec. \ref{sec:dg}). Their strength indeed lies in finding flatter minima (Sec. \ref{sec:flat_minima}), which consequently help the model to generalize especially in the heterogeneous scenario. We compare our method with algorithms proper of the FL literature and strong data augmentations (Sec. \ref{sec:cifar}), commonly used to improve generalization in DL, further validating the efficacy of our proposal. We refer to App.~\ref{app:exps} for implementation details and App.~\ref{app:abl} for the ablation studies. %

\subsection{The Effectiveness of the Search for Flat Minima in FL}
\label{sec:cifar}
In Sec.~\ref{sec:het_fl}, we have shown that, given a fixed number of rounds, FL models trained in heterogeneous settings present a considerable performance gap compared to their homogeneous counterparts. Indeed, the gap between the two scenarios can be significant with a difference of up to 20\% points (Table~\ref{table:exp1}). We identify the clients' overspecialization on local data as one of the causes of the poor generalization of the global model to the underlying training distribution. We confirm this by showing the model converges to sharp minima, correlated to a poor generalization capacity. In Table~\ref{table:exp1}, we show that explicitly optimizing for flat minima in both the local training and the server-side aggregation does help improving performances, with evident benefits especially in heterogeneous scenarios. We test \sam, \asam and their combination with \swa on the federated \textsc{Cifar10} and \textsc{Cifar100} \cite{krizhevsky2009learning,hsu2019measuring,hsu2020federated} with several levels of heterogeneity ($\alpha\in\{0,0.05,100\}$ for \textsc{Cifar10} and $\alpha\in\{0,0.5,1k\}$ for \textsc{Cifar100}) and clients participation ($K\in\{5,10,20\}$, \textit{i.e.} 5\%, 10\%, 20\%). As for \textsc{Cifar100}, we additionally test our approach on the setting proposed by \cite{reddi2020adaptive}, later referred to as \textsc{Cifar100-PAM}, where the splits reflect the ``coarse" and ``fine" label structure proper of the dataset. Since both \sam and \asam perform a step of gradient ascent and one of gradient descent for each iteration, they should be compared with \fedavg with $2$ local epochs. However, the results show \fedavg with $E=2$ suffers even more from statistical heterogeneity, so we will compare our baseline with the better-performing \fedavg with $E=1$. Our experiments reveal that applying \asam to \fedavg leads to the best accuracies with a gain of $+6\%$ and $+8\%$ points respectively on \textsc{Cifar100} and \textsc{Cifar10} in the most challenging scenario, \textit{i.e.} $\alpha=0$ and 5 clients per round. This gain is further improved by \fedasamswa with a corresponding increase of $+12\%$ and $+11.5\%$. The stability introduced by \swa especially helps with lower clients participation, where the trend is noisier. Our ablation studies (Appendix~\ref{app:abl_swa}) prove the boost given by \swa is mainly related to the average of the stochastic weights, rather than the cycling learning rate. Table~\ref{tab:lda} shows the results on \textsc{Cifar100-Pam} with ResNet18: here \sam and \samswa help more than \asam. %
\input{tables/cifar_lda2}

\noindent{\textit{\textbf{{\asam and \swa Lead to Flatter Minima in FL.}}}}
\label{sec:flat_minima}
We extend the analysis on the loss landscape and the Hessian eigenspectrum to the models trained with \fedsam, \fedasam and \swa. As expected, both the loss surfaces (Fig.~\ref{fig:loss_landscape}) and the Hessian spectra (Fig. {\color{red}{5}}) indicate us those methods indeed help converging towards flatter minima. The value of $\lambda_{max}$ goes from 93.5 with \fedavg to 70.3 with \fedsam to 30.1 with \fedasam in the most heterogeneous setting (Table~\ref{tab:eigs}). The result is further improved by \fedasamswa, obtaining $\lambda_{max}=24.6$. We notice there is a strict correspondence between the best $\lambda_{max}$ and the best ratio $\nicefrac{\lambda_{max}}{\lambda_5}$. Even if the maximum eigenvalue resulting with \fedavgswa and \fedsamswa is higher than the respective one without \swa, the corresponding lower ratio $\nicefrac{\lambda_{max}}{\lambda_5}$ actually tells us the bulk of the spectrum lies in a lower curvature region \cite{foret2020sharpness}, proving the effectiveness of \swa. Looking at \asam's behavior from each client's perspective (Fig. \ref{fig:clients_eigs}), flat minima are achieved from the very beginning of the training and that reflects positively on the model's performance. 

\noindent{\textit{\textbf{{{\asam and \swa Enable Strong Data Augmentations in FL.}}}}}
\input{tables/cifar_data_augmentations}
Data augmentations usually play a key role in the performance of a neural network and its ability to generalize \cite{zhang2017mixup,xie2020adversarial,bello2021revisiting}, but their design often requires domain expertise and greater computational capabilities, two elements not necessarily present in a federated context. In Table~\ref{tab:lda} and \ref{tab:augms}, we distinctly apply \mixup \cite{zhang2017mixup} and \cutout \cite{devries2017improved} on \textsc{Cifar100-PAM} and \textsc{Cifar100} (\textsc{Cifar10} in Appendix \ref{app:augm}). %
Surprisingly, both lead to worse performances across all algorithms, so instead of helping the model to generalize, they further slow down training. When combined with our methods, the performance improves in the heterogeneous scenarios w.r.t. the corresponding baseline (\fedavg+ data augmentation) and \swa brings a significant boost, enabling the use of data augmentation techniques in FL.

\noindent{\textit{\textbf{{{Heterogeneous FL Benefits Even More from Flat Minima.}}}}}
\input{tables/centralized}
Given the marked improvement brought by \sam, \asam and their combination with \swa, one might wonder if this simply reflects the gains achieved in the centralized scenario. In Table~\ref{tab:centr}, we prove the positive gap obtained in the heterogeneous federated scenario is larger than the centralized one, showing those approaches are actually helping the training. We also note that while \cutout and \mixup improve the performances in the centralized setting, they do not help in FL, where they achieve a final accuracy worse than \fedavg (Appendix~\ref{app:omitted_benefits_sam} for $\alpha \in \{0.5, 1k\}$). %

\noindent{\textit{\textbf{{Comparison with FL SOTA.}}}} We compare our method with \fedprox \cite{li2020federated}, \scaffold \cite{karimireddy2020scaffold}, \fedavgm \cite{hsu2019measuring}, \feddyn \cite{acar2021federated} and \adabest \cite{varno2022minimizing}, both on their own and combined with \sam, \asam and \swa (Table \ref{table:sota1}). \fedprox adds a proximal term to the local objective and, as expected \cite{li2021federated_noniid,varno2022minimizing},  does not bring any notable improvement. \scaffold uses control variates to reduce the client drift, exchanging twice the parameters at each round. While performing on par with \fedavg in the homogeneous scenario (84.5\% on \textsc{Cifar10} and 51.9\% on \textsc{Cifar100}), its performance is heavily affected by the data statistical heterogeneity. The same happens for \fedavgm. \feddyn dynamically aligns global and local stationary points and, %
as highlighted by \cite{varno2022minimizing}, is prone to parameters explosion: while it achieves good results on the simpler \textsc{Cifar10}, it requires heavy gradient clipping and is unable to reach the end of training on \textsc{Cifar100}. As a solution, \adabest is proposed, exceeding \fedavg by a few points. Our results demonstrate the consistent effectiveness of \fedasam w.r.t. the SOTA baselines, improving the accuracy by $\approx6\%$ points on the best SOTA on both datasets. Moreover, by adding \asam, all FL algorithms notably increase their performance. In particular i) we enable \fedavgm and \scaffold to train in most of the settings with highest heterogeneity, 
ii) even if limited by the necessary gradient clipping, the results reached by \feddyn on \textsc{Cifar100} are almost doubled. 
Lastly, the best results are obtained with \asamswa which stabilizes the noisy learning trends and enables models to {converge close to centralized performance} with $\alpha=0$.
\input{tables/sota_fl}

\subsection{ASAM and SWA in Real World Vision Scenarios}
In this Section, we analyze our method in real world scenarios, \textit{i.e.} large scale classification, Semantic Segmentation (SS) for autonomous driving \cite{fantauzzo2022feddrive} and Domain Generalization (DG) applied to both classification and SS.

\noindent{\textit{\textbf{{Large-scale Classification.}}}}
\label{sec:glv2}
We extend our analysis on visual classification tasks to Landmarks-User-160k \cite{hsu2020federated} to validate the effectiveness of \sam, \asam, and \swa in the presence of real-world challenges such as \textit{Non-Identical Class Distribution} (different distribution of classes per device), and \textit{Imbalanced Client Sizes} (varying number of training data per device). Results confirm the benefits of applying client-side sharpness-aware optimizers, especially in combination with server-side weight averaging with an improvement in final accuracy of up to $~7\%$. 

\noindent{\textit{\textbf{{Semantic Segmentation for Autonomous Driving.}}}}
\label{sec:ss}
SS is a fundamental task for applications of autonomous driving. Due to the private nature of the data collected by self-driving cars, it is reasonable to study this task within a federated scenario. We refer to FedDrive \cite{fantauzzo2022feddrive} - a new benchmark for autonomous driving in FL - for both settings and baselines. The employed datasets are Cityscapes \cite{cordts2016cityscapes} and IDDA \cite{alberti2020idda} with both uniform and heterogeneous settings. To test the generalization capabilities of the model when facing both semantic and appearance shift, the test domain of IDDA either contains pictures taken in the countryside, or in rainy conditions. The model is tested on both previously seen and unseen domains. As shown in Table~\ref{tab:ss_idda2}, \asam performs best both on Cityscapes and heterogeneous IDDA. The best performance is obtained combining \asamswa with \texttt{SiloBN} \cite{andreux2020siloed}, keeping the BatchNorm~\cite{ioffe2015batch} statistics local to each client~\cite{li2016revisiting} while sharing the learnable parameters across domains.
\input{tables/glv2_and_ss}

\begin{figure}[]
    \centering
    \subfloat[][\textsc{Cifar10-C}]{\includegraphics[width=.49\linewidth]{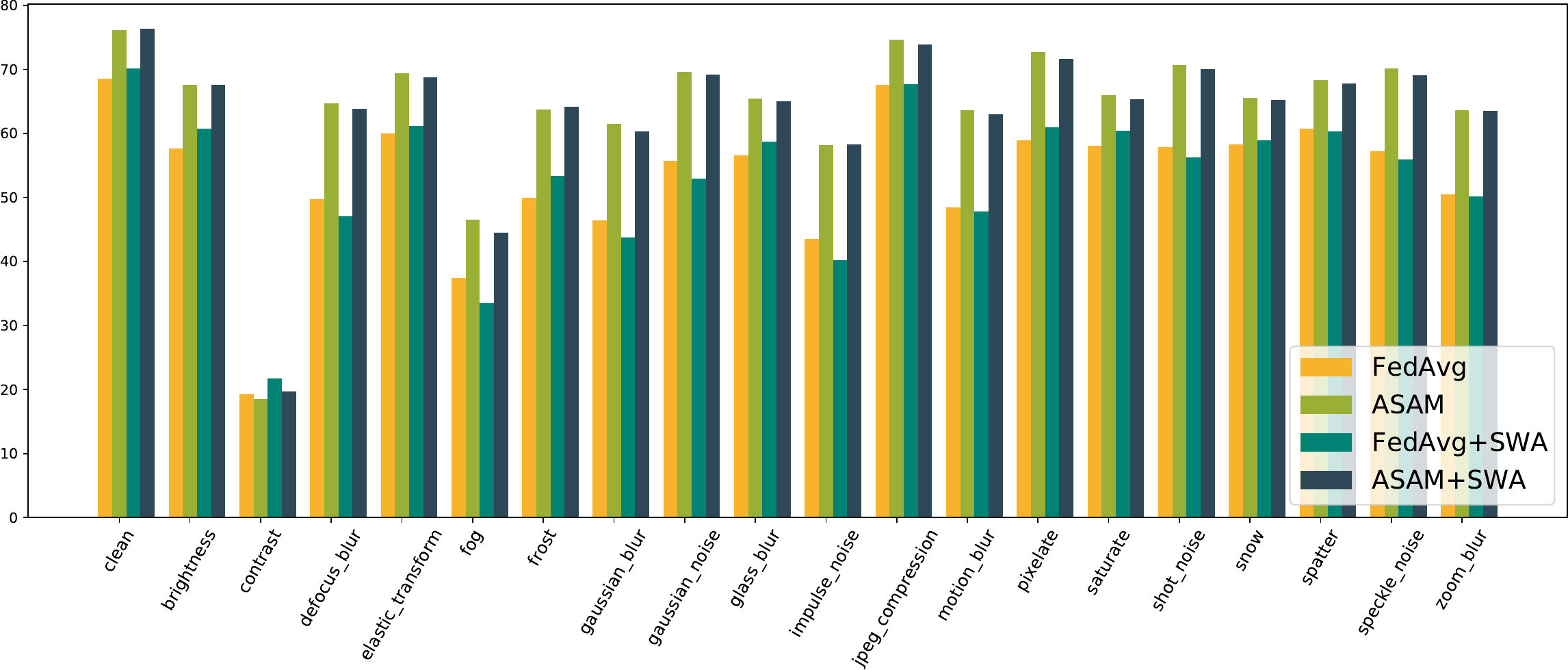}}
    \subfloat[][\textsc{Cifar100-C}]{\includegraphics[width=.49\linewidth]{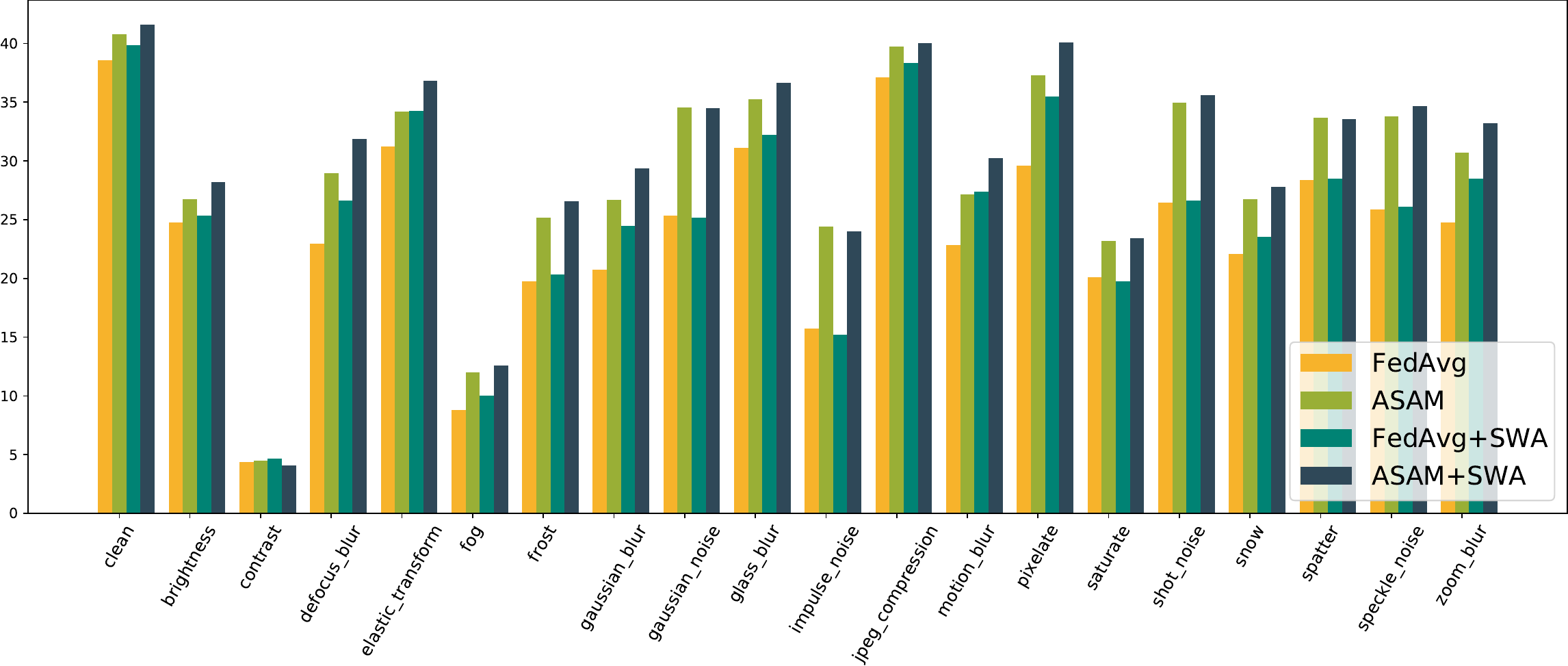}}
    \caption{\footnotesize{Domain generalization in FL. Results with $\alpha=0$, 20 clients, severity level 5.}}
    \label{fig:cifar-c}
\end{figure}

\noindent{\textit{\textbf{{Domain Generalization.}}}}
\label{sec:dg}
To further show the generalization performance acquired by the model trained with \sam, \asam and \swa, we test it on the corrupted \textsc{Cifar} datasets %
\cite{hendrycks2019benchmarking}. The test images are altered by 19 corruptions each with $5$ levels of severity. Fig.~\ref{fig:cifar-c} shows the results on the highest severity and once again validate the efficacy of seeking flat minima in FL (complete results in App.~\ref{app:corr}). %

%% file: tables/cifar.tex
\setlength{\tabcolsep}{4pt}
\begin{table}[!t]
\begin{center}
\caption{\fedsam, \fedasam and \swa on \textsc{Cifar100} %
and \textsc{Cifar10}} %
\label{table:exp1}
\tiny
\begin{tabular}{llccccccccc}
\toprule\noalign{\smallskip}
 & \multirow{2}{*}{Algorithm} & \multicolumn{3}{c}{$\alpha=0$} &\multicolumn{3}{c}{$\alpha=0.5/0.05$} & \multicolumn{3}{c}{$\alpha=1000/100$}\\
\cmidrule(l){3-5} \cmidrule(l){6-8} \cmidrule(l){9-11}
& & $5cl$& $10cl$ & $20cl$ & $5cl$& $10cl$ & $20cl$ & $5cl$& $10cl$ & $20cl$\\
\noalign{\smallskip}
\hline
\noalign{\smallskip}
\multirow{7}{*}{\rotatebox[origin=c]{90}{\textsc{Cifar100}}} & \fedavg \texttt{E=1} & 30.25 & 36.74 & 38.59 & 40.43 & 41.27 &  42.17 & 49.92 & 50.25& 50.66\\
& \fedavg \texttt{E=2} & 24.94 & 31.81 & 35.18 & 38.21 & 39.59 & 40.94  &48.72 & 48.64& 48.45\\
& \fedsam & 31.04 & 36.93 & 38.56 & 44.73 & 44.84 & 46.05  &54.01 & 53.39 &53.97\\
& \fedasam & \textbf{36.04} & \textbf{39.76} & \textbf{40.81} & \textbf{45.61} &\textbf{46.58} & \textbf{47.78}  & \underline{\textbf{54.81}} & \underline{\textbf{54.97}}&\underline{\textbf{54.50}}\\\cmidrule{2-11}
& \fedavgswa & 39.34 &39.74  & 39.85 & 43.90 &   44.02  &42.09 & 50.98&50.87 &50.92\\
& \fedsamswa & 39.30 & 39.51 & 39.24 & 47.96 & 46.76  &46.47 &\textbf{53.90} &53.67 &\textbf{54.36}\\
& \fedasamswa & \underline{\textbf{42.01}} & \underline{\textbf{42.64}} & \underline{\textbf{41.62}}  &\underline{\textbf{49.17}} &  \underline{\textbf{48.72}} & \underline{\textbf{48.27}}& 53.86& \textbf{54.79}&54.10\\
\midrule
\multirow{7}{*}{\rotatebox[origin=c]{90}{\textsc{Cifar10}}} & \fedavg \texttt{E=1} & 65.00 & 65.54 & 68.52 & 69.24 & 72.50 & 73.07  &84.46 & 84.50& 84.59\\
& \fedavg \texttt{E=2} & 61.49 & 62.22 & 66.36 & 69.23 & 69.77 & 73.48  & 83.93& 84.10&84.21 \\
& \fedsam & 70.16 & 71.09 & 72.90 & 73.52 & 74.81 & 76.04  &84.58 & 84.67 &\underline{\textbf{84.82}}\\
& \fedasam & \textbf{73.66} & \textbf{74.10} & \textbf{76.09} & \textbf{75.61} & \textbf{76.22} & \underline{\textbf{76.98}}  & \textbf{84.77} &\textbf{84.72} &84.75\\\cmidrule{2-11}
& \fedavgswa & 69.71 & 69.54 & 70.19 & 73.48 &  72.80   &73.81 &84.35 & 84.32&84.47\\
& \fedsamswa & 74.97 &73.73  & 73.06 & \underline{\textbf{76.61}} & 75.84  & 76.22&84.23 & 84.37&84.63\\
& \fedasamswa & \underline{\textbf{76.44}} & \underline{\textbf{75.51}} & \underline{\textbf{76.36}}  & 76.12 & \underline{\textbf{76.16}}  & \textbf{76.86}& \underline{\textbf{84.88}}&\underline{\textbf{84.80}} &{\textbf{84.79}}\\
\bottomrule
\end{tabular}
\end{center}
\end{table}
\setlength{\tabcolsep}{1.4pt}

%% file: tables/cifar_lda2.tex
\setlength{\tabcolsep}{3pt}
\begin{table}[!t]
\begin{center}
\caption{Accuracy results on \textsc{Cifar100-PAM} with ResNet18}
\label{tab:lda}
\tiny
\begin{tabular}{lcccc|ccc|ccc|ccc}\toprule
\multirow{3}{*}{Algorithm} &\multirow{3}{*}{Aug} & \multicolumn{6}{c}{$E=1$} & \multicolumn{6}{c}{$E=2$}\\ \cmidrule(l){3-8} \cmidrule(l){9-14}
& & \multicolumn{3}{c}{10 clients} &\multicolumn{3}{c}{20 clients} & \multicolumn{3}{c}{10 clients} &\multicolumn{3}{c}{20 clients}\\
\cmidrule(l){3-5} \cmidrule(l){6-8} \cmidrule(l){9-11} \cmidrule(l){12-14}
\textbf{} & &@$5k$ &@$10k$ &w/ \swa &@$5k$ &@$10k$ &w/ \swa &@$5k$ &@$10k$ &w/ \swa &@$5k$ &@$10k$ &w/ \swa\\\midrule

\fedavg &\multirow{3}{*}{} &46.60 &47.03 &52.70 &46.51 &45.83 &50.28 &44.58 &43.90 &51.10 &43.31 &42.88 &47.95\\
\fedsam & &50.71 &\textbf{53.10} &\textbf{55.44} &52.96 &53.41 &\textbf{54.67} &\textbf{52.36} &52.04 &55.23 &51.41 &51.35 &53.41\\
\fedasam & &49.31 &51.10 &54.25 &47.21 &\textbf{53.50} &54.29 &49.03 &49.33 &53.01 &\textbf{53.88} &52.94 &54.18\\
\midrule %
\fedavg &\multirow{3}{*}{\rotatebox[origin=c]{90}{\mixup}} &43.47 &49.25 &56.71 &50.33 &49.89 &55.74 &\textbf{44.76} &46.44 &57.15 &47.10 &47.59 &54.40\\
\fedsam & &42.83 &\textbf{51.92} &53.96 &49.66 &\textbf{55.77} &57.70 &42.17 &51.04 &56.54 &\textbf{53.50} &54.75 &\textbf{58.88}\\
\fedasam & &43.13 &51.09 &56.31 &50.51 &52.62 &56.89 &\textbf{44.74} &50.14 &\textbf{58.31} &49.87 &50.87 &55.86\\
\midrule %
\fedavg &\multirow{3}{*}{\rotatebox[origin=c]{90}{\cutout}} &48.64 &48.59 &55.40 &47.00 &46.96 &51.70 &45.19 &45.46 &55.40 &44.68 &44.25 &49.39\\
\fedsam & &48.28 &53.53 &57.25 &52.06 &\textbf{54.37} &\textbf{56.70}&\textbf{49.39} &51.88 &\textbf{57.32} &\textbf{52.16} &52.37 &55.45 \\
\fedasam & &47.52 &\textbf{52.13} &57.01 &50.01 &50.66 &53.54 &48.99 &50.09 &55.77 &48.48 &48.77 &52.00\\
\bottomrule
\end{tabular}
\end{center}
\end{table}
\setlength{\tabcolsep}{1.4pt}

%% file: tables/cifar_data_augmentations.tex
\setlength{\tabcolsep}{4pt}
\begin{table}[!t]
\begin{center}
\caption{\fedavg, \sam, \asam and \swa w/ strong data augmentations (\mixup, \cutout)}
\label{tab:augms}
\tiny
\begin{tabular}{llccccccccccc}
\toprule\noalign{\smallskip}
 & \multirow{2}{*}{Algorithm} & \multirow{2}{*}{SWA}&\multirow{2}{*}{Aug} &\multicolumn{3}{c}{$\alpha=0$} &\multicolumn{3}{c}{$\alpha=0.5/0.05$} & \multicolumn{3}{c}{$\alpha=1000/100$}\\
\cmidrule(l){5-7} \cmidrule(l){8-10} \cmidrule(l){11-13}
& & &&$5cl$& $10cl$ & $20cl$ & $5cl$& $10cl$ & $20cl$ & $5cl$& $10cl$ & $20cl$\\
\noalign{\smallskip}
\hline
\noalign{\smallskip}
\multirow{12}{*}{\rotatebox[origin=c]{90}{\textsc{Cifar100}}}&\fedavg&\ding{55}&\multirow{6}{*}{\rotatebox[origin=c]{90}{\mixup}}&29.91&33.67&35.67&35.10&37.80&39.34&55.34&\textbf{55.81}&\textbf{55.98}\\
&\fedsam&\ding{55}&&30.46&34.10&35.89&38.76&40.31&42.03&54.21&54.94&55.24\\
&\fedasam&\ding{55}&&34.04&36.82&36.97&40.71&42.24&\textbf{44.45}&49.75&49.87&49.68\\
&\fedavg&\ding{51}&&35.56&36.07&36.08&39.21&39.22&38.31&\textbf{55.43}&55.37&55.39\\
&\fedsam&\ding{51}&&35.62&36.25&35.66&42.13&41.95&42.03&52.9&53.14&53.48\\
&\fedasam&\ding{51}&&\textbf{40.08}&\textbf{38.74}&\textbf{37.47}&\textbf{44.53}&\textbf{43.97}&44.22&46.97&47.24&46.93\\
\cmidrule{2-13}
&\fedavg&\ding{55}&\multirow{6}{*}{\rotatebox[origin=c]{90}{\cutout}}&24.24&31.55&32.44&37.72&38.45&39.48&53.48&53.83&52.90\\
&\fedsam&\ding{55}&&23.51&30.92&33.12&40.33&40.31&42.58&\textbf{54.27}&\textbf{54.75}&\textbf{54.76}\\
&\fedasam&\ding{55}&&30.05&33.62&34.51&41.86&41.84&43.33&51.88&51.78&53.03\\
&\fedavg&\ding{51}&&33.65&34.40&35.03&40.43&40.12&39.32&53.87&54.09&52.75\\
&\fedsam&\ding{51}&&34.00&34.08&34.26&43.09&42.81&42.85&53.78&54.28&53.93\\
&\fedasam&\ding{51}&&\textbf{39.30}&\textbf{37.46}&\textbf{36.27}&\textbf{44.76}&\textbf{43.48}&\textbf{43.95}&50.00&49.65&50.81\\
\bottomrule
\end{tabular}
\end{center}
\end{table}
\setlength{\tabcolsep}{1.4pt}

%% file: tables/centralized.tex
\definecolor{rosso}{HTML}{cc0000}
\definecolor{verde}{HTML}{279738}
\setlength{\tabcolsep}{4pt}
\begin{table}[!t]
\begin{center}
\setlength\tabcolsep{0.25cm}
\caption{\footnotesize{Comparison of improvements (\%) in centralized and heterogeneous federated scenarios ($\alpha=0$, 5 clients) on \textsc{Cifar100}, computed w.r.t. the reference at the bottom}}
\label{tab:centr}
\tiny
\begin{tabular}{lcccccc}
\toprule\noalign{\smallskip}
\multirow{2}{*}{Algorithm} & \multicolumn{2}{c}{Accuracy}& \multicolumn{2}{c}{Absolute Improvement} & \multicolumn{2}{c}{Relative Improvement}\\
\cmidrule(l){2-3} \cmidrule(l){4-5} \cmidrule(l){6-7}
 & Centr. & $\alpha=0$  & Centr. & $\alpha=0$ & Centr. & $\alpha=0$\\
\midrule
\sam & 55.22&31.04&+3.02&+0.79&+5.79&+2.61\\
\asam & 55.66&36.04&+3.46&{\color{verde}{+5.79}}&+6.63&{\color{verde}{+19.14}}\\
\swa&52.72&39.34&+0.52&{\color{verde}{+9.09}}&+1.00&{\color{verde}{+30.05}}\\
\samswa&55.75&39.30&+0.55&{\color{verde}{+9.05}}&+1.06&{\color{verde}{+29.92}}\\
\asamswa&55.96&42.01&+3.76&{\color{verde}{\textbf{+11.76}}}&+7.20&{\color{verde}{\textbf{+38.88}}}\\
\mixup&58.01&29.91&+5.81&{\color{rosso}{-0.34}}&+11.13&{\color{rosso}{-1.12}}\\
\cutout&55.30&24.24&+3.10&{\color{rosso}{-6.01}}&+5.94&{\color{rosso}{-19.87}}\\
\midrule
\multicolumn{7}{l}{\texttt{Centralized}: \textbf{52.20} - \fedavg: \textbf{30.25}}\\
\bottomrule
\end{tabular}
\end{center}
\end{table}
\setlength{\tabcolsep}{1.4pt}

%% file: tables/sota_fl.tex
\renewcommand{\arraystretch}{0.7}
\setlength{\tabcolsep}{5pt}
\begin{table}[!t]
\begin{center}
\caption{SOTA comparison on \textsc{Cifar10} and \textsc{Cifar100} (\underline{centralized performance})}
\label{table:sota1}
\tiny
\begin{tabular}{llcccccccc}
\toprule\noalign{\smallskip}
 & \multirow{3}{*}{Algorithm} & \multicolumn{4}{c}{\textsc{\textbf{w/o SWA}}} & \multicolumn{4}{c}{\textsc{\textbf{w/ SWA}}}\\ \cmidrule(l){3-6} \cmidrule(l){7-10}
 
 & & \multicolumn{2}{c}{$\alpha=0$} &\multicolumn{2}{c}{$\alpha=0.05/0.5$}  & \multicolumn{2}{c}{$\alpha=0$} &\multicolumn{2}{c}{$\alpha=0.05/0.5$}\\
\cmidrule(l){3-4} \cmidrule(l){5-6} \cmidrule(l){7-8} \cmidrule(l){9-10}
& & $5cl$&  $20cl$ & $5cl$&  $20cl$ & $5cl$& $20cl$ & $5cl$& $20cl$\\
\noalign{\smallskip}
\hline
\noalign{\smallskip}

\multirow{13}{*}{\rotatebox[origin=c]{90}{\textsc{Cifar10}}} & \fedavg  & 65.00 &  68.52 & 69.24 &  73.07  & 69.71 &  70.19 & 73.48   &73.81\\
& \fedsam  & 70.16 &  72.90 & 73.52 & 76.04  & 74.97   & 73.06 & {{76.61}} & 76.22\\
& \fedasam  & \textbf{{73.66}} &  \textbf{{76.09}} & \textbf{{75.61}} & {{76.98}}  & {{{76.44}}} &  \textbf{{76.36}}  & 76.12 &  {76.86}\\

& \fedavgm & \textcolor{red}{10.00} & \textcolor{red}{10.00} & \textcolor{red}{10.00} & \textbf{78.51}&  \textcolor{red}{10.00}  & \textcolor{red}{10.00} & \textcolor{red}{10.00} & \underline{\textbf{84.00}}\\
& \fedprox  & 62.72 &  68.44 & 68.38 &  73.02  & 70.56  & 70.08 & 74.27  & 73.67\\
& \scaffold  & 32.25 &  15.56 & 54.46 &  44.76 & 11.98& \textcolor{red}{10.00}& 33.25 &  24.11\\
& \feddyn  & 67.69 & 73.81 & 71.36 &  75.20& 77.00   & 74.00 & {77.99} &  75.12\\
& \adabest  & 66.77 & 72.29& 69.84 & 75.89 & \textbf{78.94} & 76.12&\textbf{ 80.35} & 79.35\\\cmidrule{2-10}

& \fedavgmasam & 77.30 &  \underline{\textbf{84.89}} & 77.06  &  \underline{\textbf{84.92}}  & 80.88  & \underline{{85.98}} & 78.29 & \underline{\textbf{86.03}} \\

& \fedproxasam  & 73.74 &  75.76 & 75.32 &  77.03 & 76.89 &  75.92 & 76.65 &  76.95 \\
& \scaffoldasam & \textbf{77.78} &  77.93 & 77.59 &  77.80&  75.66 &  75.30 & 75.32 &  75.29 \\

& \feddynsam  & 77.38 & 81.00 & \textbf{79.18}&  81.70  & {\textbf{83.81}} &  \underline{\textbf{86.07}} & {\textbf{83.18}} &  \underline{85.57}\\

& \adabestasam & 77.48 & 78.43  & 78.41& 79.72 & 82.00 & 80.80&81.87 & 80.81\\
\midrule

\multirow{13}{*}{\rotatebox[origin=c]{90}{\textsc{Cifar100}}} & \fedavg & 30.25 &  38.59 & 40.43 &   42.17  & 39.34   & 39.85 & 43.90   &42.09\\
& \fedsam  & 31.04  & 38.56 & 44.73  & 46.05 & 39.30 & 39.24 & 47.96   &46.47\\
& \fedasam  & \textbf{36.04}  & \textbf{40.81} & \textbf{45.61}  & \textbf{47.78}   & {{42.01}}  & {{41.62}}  &\textbf{{49.17}} &   {{48.27}}\\

& \fedavgm & \textcolor{red}{1.00}  & 40.64 & 4.60  & \textbf{47.88}  & \textcolor{red}{1.00} & \underline{\textbf{53.50}} & 4.60  & \underline{\textbf{53.69}}\\
& \fedprox  & 31.20  & 38.59  & 39.53 & 42.17& 39.06  & 39.68 & 43.98  & 41.84\\
& \scaffold  & \textcolor{red}{1.00} & \textcolor{red}{1.00} & 33.26  & \textcolor{red}{1.00}  & \textcolor{red}{1.00} & \textcolor{red}{1.00} & 5.76 & \textcolor{red}{1.00} \\
& \feddyn  & \textcolor{red}{1.00}  & \textcolor{red}{1.40} & 22.03   & 24.75  & \textcolor{red}{1.00} & \textcolor{red}{1.40} & 8.27 & 35.15\\
& \adabest  & 29.90 & 39.11 & 36.93&  43.25 & \textbf{44.48}& 44.21 & 48.20 & 44.51\\
\cmidrule{2-10}

& \fedavgmasam &  \textcolor{red}{1.00}  & 39.61 & 4.60  & \underline{\textbf{51.65}}   & \textcolor{red}{1.00} & \textbf{\underline{51.58}} & 4.60 & \underline{\textbf{56.19}} \\
& \fedproxasam  & 36.10  & 40.91 & 44.81  & 48.17  & 43.90 &  42.06 & 48.66  & 48.19  \\
& \scaffoldasam  & \textbf{43.65} & 42.61 & \textbf{46.50} & 46.76 & 40.63 &  39.07 & 44.87  & 44.28 \\
& \feddynasam & 22.16  & 23.51 & 38.43 & 38.60 & 17.51& 19.22& 38.60 & 31.06\\
& \adabestasam  & 39.75 &  \textbf{45.00}& 45.25& {49.56} & \underline{\textbf{51.75}} & {{47.42}}& \underline{\textbf{51.89}} & \underline{51.47}\\

\bottomrule
\end{tabular}
\end{center}
\end{table}
\setlength{\tabcolsep}{1.4pt}

%% file: tables/glv2_and_ss.tex
\begin{table}[!t]
    \begin{minipage}{.5\columnwidth}
        \scriptsize
        \caption{Accuracy Results (\%) on Landmarks-User-160k}
        \label{tab:gldv2_2}
        \begin{tabular}{lcccc}\toprule
            &@$5k$ rounds &w/ \swa 75 &w/ \swa 100 \\\midrule
            \fedavg &61.91 &66.05 &67.52 \\
            \fedsam &63.72 &67.11 &68.12 \\
            \fedasam &64.23 &67.17 &\textbf{68.32} \\
            \midrule
            \texttt{Centralized} &\multicolumn{3}{c}{74.03} \\
            \bottomrule
        \end{tabular}
    \end{minipage}\hfill
    \begin{minipage}{.5\columnwidth}
        \caption{Federated SS on Cityscapes and IDDA. Results in mIoU (\%) @ $1.5k$ rounds}
    \label{tab:ss_idda2}
    \tiny
\begin{tabular}{lcccccccc}
\toprule\noalign{\smallskip}
  \multirow{2}{*}{Algorithm }& \multirow{2}{*}{Uniform} & &\multicolumn{2}{c}{Country} & \multicolumn{2}{c}{Rainy}& &\multirow{2}{*}{mIoU}\\
 &&& seen & unseen & seen & unseen\\
\midrule
\fedavg &\multicolumn{1}{c|}{\ding{51}} &\multirow{15}{*}{\rotatebox[origin=c]{90}{\textsc{IDDA}}}& \multicolumn{1}{|c}{63.31}&48.60&\underline{\textbf{65.16}}&\multicolumn{1}{c|}{27.38}&\multirow{15}{*}{\rotatebox[origin=c]{90}{\textsc{Cityscapes}}}&\multicolumn{1}{|c}{43.61}\\
\fedsam &\multicolumn{1}{c|}{\ding{51}} && \multicolumn{1}{|c}{\underline{\textbf{64.22}}}&\underline{\textbf{49.74}}&64.81&\multicolumn{1}{c|}{30.00}&&\multicolumn{1}{|c}{44.58}\\
\fedasam&\multicolumn{1}{c|}{\ding{51}} & &\multicolumn{1}{|c}{62.74}&48.73&64.74&\multicolumn{1}{c|}{\textbf{31.32}}&&\multicolumn{1}{|c}{\underline{\textbf{45.86}}}\\
\cdashline{1-2} \cdashline{4-7} \cdashline{9-9}
\fedavgswa&\multicolumn{1}{c|}{\ding{51}} &&\multicolumn{1}{|c}{\textbf{63.91}}&43.28&63.24&\multicolumn{1}{c|}{47.72}&&\multicolumn{1}{|c}{45.64}\\
\fedsamswa&\multicolumn{1}{c|}{\ding{51}} &&\multicolumn{1}{|c}{62.26}&\textbf{46.26}&\textbf{63.69}&\multicolumn{1}{c|}{48.40}&&\multicolumn{1}{|c}{45.29}\\
\fedasamswa&\multicolumn{1}{c|}{\ding{51}} &&\multicolumn{1}{|c}{60.78}&44.23&63.18&\multicolumn{1}{c|}{\underline{\textbf{51.76}}}&&\multicolumn{1}{|c}{\textbf{45.69}}\\
\cmidrule(l){1-2} \cmidrule(l){4-7} \cmidrule(l){9-9}
\fedavg & \multicolumn{1}{c|}{\ding{55}}&&\multicolumn{1}{|c}{42.06}&36.04&39.50&\multicolumn{1}{c|}{24.59}&&\multicolumn{1}{|c}{38.65}\\
\fedsam &\multicolumn{1}{c|}{\ding{55}}&&\multicolumn{1}{|c}{43.28}&{\textbf{37.83}}&39.65&\multicolumn{1}{c|}{29.27}&&\multicolumn{1}{|c}{41.22}\\
\fedasam&\multicolumn{1}{c|}{\ding{55}}&&\multicolumn{1}{|c}{\textbf{43.67}}&36.11&\textbf{41.68}&\multicolumn{1}{c|}{\textbf{30.07}}&&\multicolumn{1}{|c}{\textbf{42.27}}\\
\cdashline{1-2} \cdashline{4-7} \cdashline{9-9}
\fedavgswa&\multicolumn{1}{c|}{\ding{55}} &&\multicolumn{1}{|c}{37.16}&37.48&37.06&\multicolumn{1}{c|}{42.33}&&\multicolumn{1}{|c}{42.48}\\
\fedsamswa&\multicolumn{1}{c|}{\ding{55}} &&\multicolumn{1}{|c}{44.26}&\underline{\textbf{40.45}}&38.15&\multicolumn{1}{c|}{45.25}&&\multicolumn{1}{|c}{\textbf{43.42}}\\
\fedasamswa&\multicolumn{1}{c|}{\ding{55}} &&\multicolumn{1}{|c}{\textbf{45.23}}&39.72&\textbf{42.09}&\multicolumn{1}{c|}{\underline{\textbf{45.40}}}&&\multicolumn{1}{|c}{43.02}\\
\cdashline{1-2} \cdashline{4-7} \cdashline{9-9}
\texttt{SiloBN}& \multicolumn{1}{c|}{\ding{55}}&&\multicolumn{1}{|c}{45.86}&32.77&48.09&\multicolumn{1}{c|}{39.67}&&\multicolumn{1}{|c}{45.96}\\
\texttt{SiloBN + SAM}&\multicolumn{1}{c|}{\ding{55}}&&\multicolumn{1}{|c}{\underline{\textbf{46.88}}}&33.71&48.22&\multicolumn{1}{c|}{40.08}&&\multicolumn{1}{|c}{49.10}\\
\texttt{SiloBN + ASAM}&\multicolumn{1}{c|}{\ding{55}}&&\multicolumn{1}{|c}{46.57}&\textbf{35.22}&\underline{\textbf{48.33}}&\multicolumn{1}{c|}{{\textbf{40.76}}}&&\multicolumn{1}{|c}{\underline{\textbf{49.75}}}\\
\bottomrule
\end{tabular}
\end{minipage}
\end{table}

%% file: sections/6.conclusions.tex
\section{Conclusions}
Heterogeneous Federated Learning suffers from degraded performances and slowdown in training due to the poor generalization of the learned global model. Inspired by recent trends in deep learning connecting the loss landscape and the generalization gap, we analyzed the behavior of the model through the lens of the geometry of the loss surface and linked the lack of generalization to convergence towards sharp minima. As a solution, we introduced Sharpness-Aware Minimization, its adaptive version and Stochastic Weight Averaging in FL for encouraging convergence towards flatter minima. We showed the effectiveness of this approach in several vision tasks and datasets.\\
{\noindent{\textbf{Acknowledgments.}} We thank L. Fantauzzo for her help with the SS experiments. We acknowledge the \textsc{Cineca} HPC infrastructure. Work funded by \textsc{Cini}.}

%% file: supplementary.tex
{\noindent
\textbf{\Large{Appendix}}}

\section{Background}
\label{app:background}
In this section, we briefly review the details of Sharpness-Aware Minimization (SAM) \cite{foret2020sharpness}, its adaptive version (ASAM) \cite{kwon2021asam} and Stochastic Weight Averaging (SWA) \cite{izmailov2018averaging}.
\subsection{SAM and ASAM: Overview}
\subsubsection{SAM} aims at finding the solution $\theta$ surrounded by a neighborhood having uniform low training loss $\mathcal{L}_\mathcal{D}(\theta)$, \textit{i.e.} located in a flat minimum. The \textit{sharpness} of a training loss function is defined as:
\begin{equation}
    \max_{||\epsilon||_p \leq \rho}  \mathcal{L}_\mathcal{D}(\theta+\epsilon) - \mathcal{L}_\mathcal{D}(\theta)
    \label{math:sharpness}
\end{equation}
where $\rho$ is an hyper-parameter defining the neighborhood size and $p\in[1,\infty)$. SAM aims at minimizing the sharpness of the loss solving the following minmax objective:
\begin{equation}
    \min_{\theta\in\mathbb{R}^d}\max_{||\epsilon||_p \leq \rho}  \mathcal{L}_\mathcal{D}(\theta+\epsilon) + \lambda ||\theta||_2^2
    \label{math:minmax_sam}
\end{equation}
where $\lambda$ is a hyper-parameter weighing the importance of the regularization term. In \cite{foret2020sharpness}, it is shown that $p=2$ is typically the optimal choice, hence, without loss of generality, we use the ${\ell}_2$-norm in the maximization over $\epsilon$ and omit the regularization term for simplicity. %
In order to obtain the exact solution of the inner maximization problem $\epsilon^* \triangleq \arg\max_{||\epsilon||_2 \leq \rho} \mathcal{L}(\theta+\epsilon)$, the authors propose to employ a first-order approximation of $\mathcal{L}(\theta+\epsilon)$ around $0$:
\begin{equation}
    \epsilon^* \approx \arg\max_{||\epsilon||_2 \leq \rho} \mathcal{L}_\mathcal{D}(\theta) + \epsilon^T \nabla_{\theta}\mathcal{L}_\mathcal{D}(\theta) =\rho\frac{\nabla_{\theta}\mathcal{L}_\mathcal{D}(\theta)}{||\nabla_{\theta}\mathcal{L}_\mathcal{D}(\theta)||_2} =: \hat{\epsilon}(\theta)
    \label{math:approx_sam}
\end{equation}
Under this computationally efficient approximation, 
$\hat{\epsilon}(\theta)$ is nothing more than a scaled gradient of the current parameters $\theta$. The sharpness-aware gradient is then defined as $\nabla_{\theta}\mathcal{L}_\mathcal{D}(\theta)|_{\theta + \hat{\epsilon}(\theta)}$ and used to update the model as 
\begin{equation}
\theta_{t+1} \leftarrow \theta_{t} - \gamma \nabla_{\theta_t}\mathcal{L}_\mathcal{D}(\theta_t)|_{\theta_t + \hat{\epsilon}_t}, \label{math:sam_update}
\end{equation}
where $\gamma$ is an appropriate learning rate and $\hat{\epsilon}_t = \hat{\epsilon}(\theta_t)$. This two-steps procedure is iteratively applied to solve Eq. \ref{math:minmax_sam}. Intuitively, SAM performs a first step of gradient ascent to estimate the point $(\theta_t + \hat{\epsilon}_t)$ at which the loss is approximately maximized and then applies gradient descent at $\theta_t$ using the just computed gradient.

\subsubsection{ASAM} In \cite{kwon2021asam}, the authors point out that sharpness defined in a rigid region with a fixed radius $\rho$ (Eq. \ref{math:sharpness}) is sensitive to parameter re-scaling, negatively affecting the connection between sharpness and generalization gap. %
If $A$ is a scaling operator acting on the parameters space without changing the loss function,
two neural networks with weights $\theta$ and $A\theta$ can have different values of sharpness while maintaining the same generalization gap, \textit{i.e.} the sharpness is \textit{scale-dependent}. As a solution, they introduce the concept of \textit{adaptive} sharpness, defined as  %
\begin{equation}
    \max_{||T_{\theta}^{-1}\epsilon||_p \leq \rho}  \mathcal{L}_\mathcal{D}(\theta+\epsilon) - \mathcal{L}_\mathcal{D}(\theta)
    \label{math:adaptive_sharpness}
\end{equation}
where $T_{\theta}^{-1}$ is the normalization operator of $\theta$ such that $T_{A\theta}^{-1}A = T_{\theta}^{-1}$. Eq. \ref{math:minmax_sam} can be rewritten to define the Adaptive Sharpness-Aware Minimization (ASAM) problem as follows:
\begin{equation}
    \min_{\theta\in\mathbb{R}^d}\max_{||T_{\theta}^{-1}\epsilon||_p \leq \rho}  \mathcal{L}_\mathcal{D}(\theta+\epsilon) + \lambda ||\theta||_2^2
    \label{math:asam}
\end{equation}
For improving stability, $T_{\theta}$ is substituted by $T_{\theta} + \eta I_w$, where $\eta>0$ is a hyper-parameter controlling the trade-off between stability and adaptivity, while $w$ is the number of weight parameters of the model.

\subsection{Stochastic Weight Averaging: Overview}
SWA averages weights proposed by SGD, while using a learning rate schedule to explore regions of the weight space corresponding to high performing networks. At each step $i$ of a cycle of length $c$, the learning rate is decreased from $\gamma_1$ to $\gamma_2$:
\begin{equation}
    \gamma(i) = \big(1-t(i)\big) \gamma_1 + t(i) \gamma_2, \quad t(i) = \frac{1}{c}\big(mod(i-1,c)+1\big) 
    \label{math:lr_swa}
\end{equation}
If $c=1$ the learning rate is constant ($\gamma_1$), otherwise for $c>1$ the learning schedule is cyclical. Starting from a pre-trained model $f_{\hat{\theta}}$, SWA captures all the updates $\theta$ at the end of each cycle and averages them as:
\begin{equation}
    \theta_{\text{SWA}} \leftarrow \frac{\theta_{\text{SWA}}\cdot n_{\text{models}} + \theta}{n_{\text{models}}+1}
    \label{math:swa}
\end{equation}
obtaining the final model $f_{\theta_{\text{SWA}}}$, where $n_\text{models}$ keeps track of the number of completed cycles.

In our method, SWA is applied on the server-side to make the learning process more robust. 
Adapting the scenario of \cite{izmailov2018averaging} to FL, from $75\%$ of the training onwards, the server keeps two models, $f_\theta$ and $f_{\theta_{\text{\swa}}}$ ($f$ and $f_{\text{\swa}}$ to simplify the notation). $f$ follows the standard FedAvg paradigm, while $f_{\text{\swa}}$ is updated every $c$ rounds (Eq. \ref{math:swa}). At each round, the cycling learning rate is computed (Eq. \ref{math:lr_swa}) and used for the clients' local training.

\subsection{Mixup and Cutout: Overview}
Mixup and Cutout are recent methods for data augmentation, aiming to improve the learned models' generalization. We apply one of the two in the client-side training.

\subsubsection{mixup} \cite{zhang2017mixup} trains the neural network on convex combinations of images and their labels, exploiting the prior knowledge that linear interpolation of features leads to linear interpolations of their corresponding targets. Given two input images $(x_i,x_j)$ and their corresponding one-hot label encodings $(y_i,y_j)$ drawn from the $k$-th client's training data $\mathcal{D}_k$, virtual training examples are constructed as follows:
\begin{equation}
\begin{split}
    \Bar{x} = \lambda x_i + (1-\lambda) x_j\\
    \Bar{y} = \lambda y_i + (1-\lambda) y_j
\end{split}
\end{equation}
with $\lambda\sim \text{Beta}(\alpha,\alpha)$ for $\alpha\in(0,\infty)$. 
\subsubsection{Cutout} \cite{devries2017improved} regularizes learning by randomly masking out square regions of the input during training. At the implementation level, this corresponds to applying a fixed-size zero-mask to a random location of the image. 

\section{Training in Heterogeneous Scenarios - Additional Material}
\begin{figure}[t]
    \centering
    \subfloat[][]{\includegraphics[width=.4\linewidth]{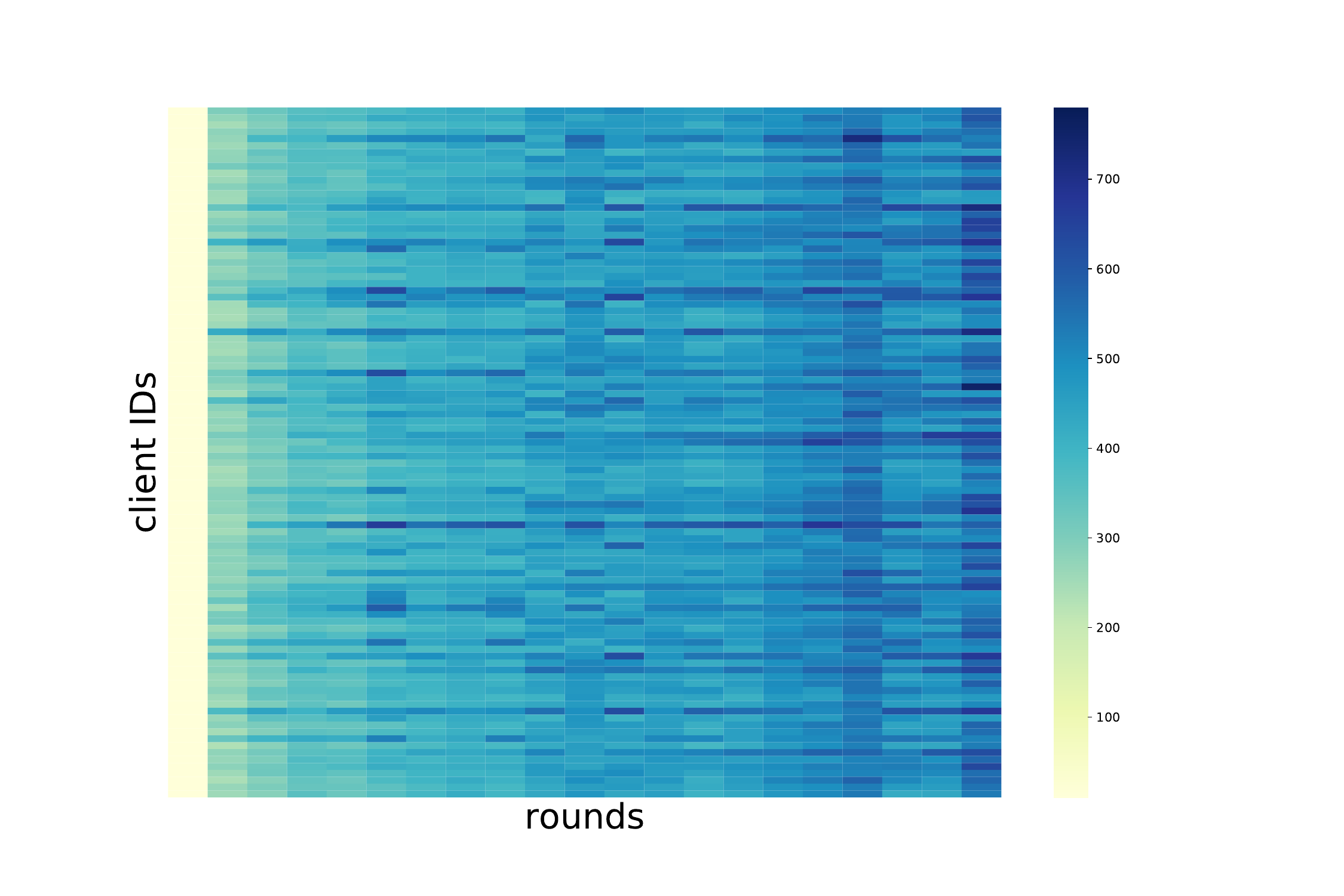}}
    \subfloat[][]{\includegraphics[width=.4\linewidth]{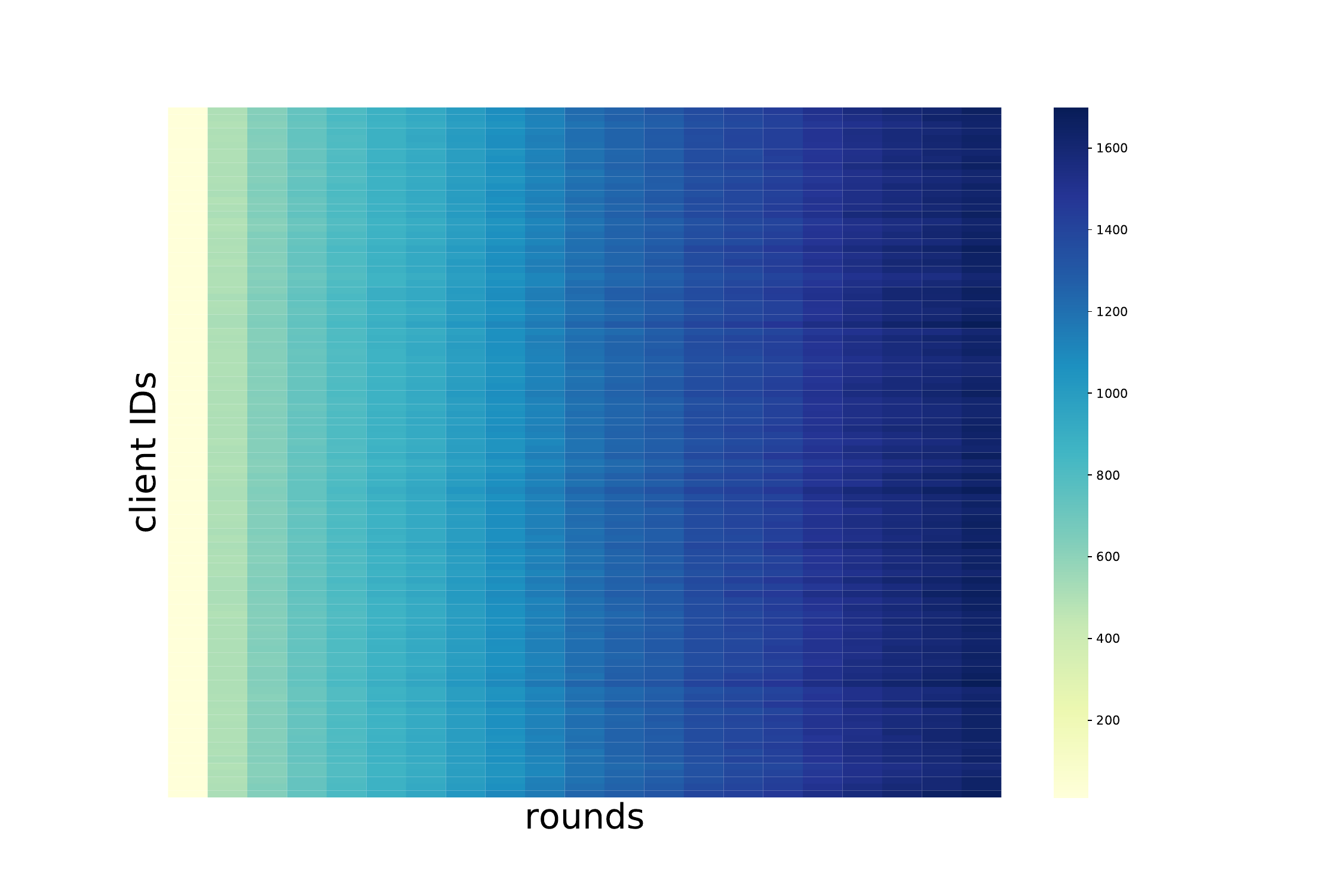}}
    \caption{\textsc{Cifar100}. L2-norm of global classifier output features as rounds pass, after receiving as input each client's local data. \textbf{(a)} with $\alpha=0$, the model tends to focus on a different client's distribution, \textit{i.e.} on a single class, at each round. \textbf{(b)} when $\alpha=1000$, the model gives the same attention to each distribution.}
    \label{fig:clf_feats}
\end{figure}
In this section, we provide further analysis of the model's behavior in heterogeneous and homogeneous federated scenarios. As explained in Sec. \ref{sec:het_fl}, the model trained under a condition of statistical heterogeneity is subject to oscillations and loss in performance and generalization. Fluctuations in model predictions can also be noted by looking at its output features, defined as $f_\theta(x) \: \forall x\in\mathcal{X}$. Fig. \ref{fig:clf_feats} shows the L2-norm of the output features computed using the current global model $f_{\theta}^t \: \forall t\in[T]$, given as input the local clients' data $\mathcal{D}_k \: \forall k\in[K]$, where a higher norm value corresponds to greater attention paid to that class by the network. The uniformity of the features obtained in the homogeneous setting contrasts with the chaotic distribution of the ones resulting when $\alpha=0$, which significantly vary over time without following a constant trend.

\section{Experiments Details}
\label{app:exps}
Here we provide a detailed description of the datasets and models used in the paper, together with information regarding the chosen hyper-parameters and their fine-tuning intervals. All results presented in both the main text and the Appendix are averaged over the last 100 rounds for increased robustness and reliability. Unless otherwise specified, the framework is PyTorch \cite{NEURIPS2019_9015} and experiments were run on one NVIDIA GeForce GTX 1070.

\subsection{Datasets and Models}
 Table \ref{tab:stats} summarizes the tasks and the statistics of the number of clients and examples for each dataset.

\begin{table}[!t]\centering
\caption{Datasets statistics}\label{tab:stats}
\scriptsize
    \begin{tabular}{lccccc}
    \toprule
    Dataset & Task & Train clients & Size imbalance & Train samples & Test samples\\\midrule
    \textsc{Cifar10} &  Classification  & 100 & \ding{55} & 50,000 & 10,000\\
    \textsc{Cifar100} & Classification & 100 & \ding{55} & 50,000 & 10,000\\
    \textsc{Cifar100-Pam} &  Classification  &500 & \ding{55} & 50,000 & 10,000\\
    \textsc{Cifar10-C} & DG & - & - & - & 10,000\\
    \textsc{Cifar100-C} & DG & - & - & - & 10,000\\
    \textsc{Landmarks-User-160k} &  Classification  & 1,262 &  \ding{51} & 164,172 & 19,526\\
    \textsc{Cityscapes} (uniform) & SS & 146 &  \ding{51} & \multirow{2}{*}{2,975} & \multirow{2}{*}{500}\\
    \textsc{Cityscapes} (heterogeneous) &  SS & 144 & \ding{51}& & \\
    \textsc{Idda} (country) & SS+DG & 90 &  \ding{55} &  4,320 &1,920\\
    \textsc{Idda} (rainy) & SS+DG & 69 &  \ding{55} &  3,312 &2,928\\
    \bottomrule
    \end{tabular}
\end{table}

\subsubsection{CIFAR10 and CIFAR100} We replicate the federated version of the \textsc{Cifar} datasets proposed by \cite{hsu2019measuring}. Each dataset is split among 100 clients, receiving 500 images each according to the latent Dirichlet distribution (LDA) applied to the labels. The client's examples are selected following a multinomial distribution drawn from a symmetric Dirichlet distribution with parameter $\alpha$. The higher the value of $\alpha$ the larger the number of classes locally seen , \textit{i.e.} the more similar and homogeneous the clients' distributions are. We test $\alpha\in\{0,0.05,100\}$ on \textsc{Cifar10} and $\alpha\in\{0,0.5,1000\}$ on \textsc{Cifar100}. The task is image classification on 10 (\textsc{Cifar10}) and 100 (\textsc{Cifar100}) classes. 
\paragraph{Model:} We train a Convolutional Neural Network (CNN) similar to LeNet5 \cite{lecun1998gradient} on both datasets, following the setting of \cite{hsu2020federated}. The network has two 64-channels convolutional layers with kernel of size $5\times5$, each followed by a $2\times2$ max-pooling layer, ended by two fully connected layers with 384 and 192 channels respectively and a linear classifier. 
\paragraph{Data pre-processing:} The $32\times32$ input images are pre-processed following the standard pipeline: the training images are randomly cropped applying padding 4 with final size $32\times32$, randomly horizontally flipped with probability 0.5 and finally the pixel values are normalized with the dataset's mean and standard deviation; normalization is applied to test images as well.

\subsubsection{CIFAR100-PAM} We further extend our experiments to a more complex version of \textsc{Cifar100}, \textit{i.e.} \textsc{Cifar100-Pam} proposed by \cite{reddi2020adaptive}, reflecting the ``coarse" and ``fine" label structure of the dataset for a more realistic partition. The dataset is split among 500 clients - with 100 images each - following the Pachinko Allocation Method (PAM) \cite{li2006pachinko}, on the result of which LDA is applied. 
\paragraph{Model:} We train a modified ResNet18, replacing Batch Normalization~\cite{ioffe2015batch} layers with group normalization (GN) ones \cite{wu2018group}, as suggested by \cite{hsieh2020non}. We use two groups for each GN layer. Experiments have been run using \texttt{FedJAX}~\cite{ro2021fedjax} on a cluster with NVIDIA V100 GPUs.

\paragraph{Data pre-processing:} \textsc{Cifar100-Pam} images are pre-processed as the \textsc{Cifar} LDA versions described above.

\subsubsection{CIFAR10-C and CIFAR100-C} are the corrupted versions of the \textsc{Cifar} datasets. They are part of the benchmark proposed by \cite{hendrycks2019benchmarking}, used for testing the image classifiers' robustness. The $10k$ images-test set is modified according to a given \textit{corruption} and a corresponding level of \textit{severity}. There are 19 possible corruptions (brightness, contrast, elastic blur, elastic transform, fog, frost, Gaussian blur, Gaussian noise, glass blur, impulse noise, JPEG compression, motion blur, pixelate, saturate, short noise, snow, spatter, speckle noise, zoom blur), while the severity ranges from 1 (low) to 5 (high). 
\paragraph{Model:} The same model described for \textsc{Cifar10} and \textsc{Cifar100} is used here. To test the generalization ability of our method, we test the model trained with \textsc{Cifar10/100} on the corresponding corrupted dataset.
 
\subsubsection{Landmarks-User-160k} Introduced by \cite{hsu2020federated}, the Landmarks-User-160k dataset comprises 164,172 training images belonging to 2,028 landmarks. The dataset is created according to the authorship information from the large-scale dataset Google Landmarks v2 (GLv2) \cite{weyand2020google}. Each author owns at least 30 pictures depicting $5$ or more landmarks, while each location is depicted by at least 30 images and was visited by no less than $10$ users. The authors in the test set do not overlap with the ones appearing in the training split. 
\paragraph{Model:} 
We follow a setting similar to the one proposed by~\cite{hsu2020federated} and use a MobileNetV2~\cite{sandler2018mobilenetv2} network pre-trained on ImageNet~\cite{deng2009imagenet} with with GroupNorm layers in place of BatchNorm. Since no details on the model are available, we set the network feature multiplier $\alpha=1$ and use $8$ groups for the GN layers. We did not apply a bottleneck layer before the classifier as specified in~\cite{hsu2020federated}.
To reduce training time, we use \texttt{Flax}~\cite{flax2020github} for both pre-training and centralized baselines, and \texttt{FedJAX}~\cite{ro2021fedjax} for the implementation of the federated algorithms. Both libraries are based on \texttt{JAX}~\cite{jax2018github} and allow for efficient data parallelization. Implementation of the MobileNetV2 backbone used for all the experiments is available here\footnote{\url{https://github.com/rwightman/efficientnet-jax/tree/a65811fbf63cb90b9ad0724792040ce93b749303}}. All large-scale classification experiments have been performed using an NVIDIA DGX A100 40GB. 

The model trained on ImageNet reaches $\approx 68\%$ top-$1$ accuracy on the validation set. In our experience, GroupNorm tends to perform slightly worse than BatchNorm when trained on ImageNet. However, since we did not extensively tune the hyper-parameters, getting better final performance is possible. For the ImageNet training, we used $8$ GPUs with a total batch size of $2048$ images.

\paragraph{Data pre-processing:} We applied the same data augmentation for training the model on ImageNet and fine-tuning on GLv2: we crop and resize the input images to $224\times224$ with random scale and aspect ratio as described in \cite{szegedy2015going}. 
The data augmentation pipeline used for the experiments can be found here\footnote{\url{https://github.com/google/flax/blob/571018d16b42ce0a0387515e96ba07130cbf79b9/examples/imagenet/input_pipeline.py}}.
We also adapted the GLv2 TensorFlow Federated data pipeline\footnote{
\url{https://www.tensorflow.org/federated/api_docs/python/tff/simulation/datasets/gldv2/load_data}} to be compatible with \texttt{FedJAX}.

\subsubsection{Cityscapes} \cite{cordts2016cityscapes} is a popular dataset for Semantic Segmentation and contains 2,975 real photos taken in the streets of 50 different cities under good weather conditions. Annotations are provided for 19 semantic classes. We refer to the federated splits proposed in the FedDrive benchmark \cite{fantauzzo2022feddrive}. The \textit{uniform} version of the dataset randomly assigns each image to one of the 146 users. In order to account for the distribution heterogeneity appearing in real-world scenarios, an ulterior version is proposed, referred to as \textit{heterogeneous}: every client only accesses images from one of the 18 training cities. In both cases, the test set contains pictures of unseen cities.
\paragraph{Model:} As proposed by the authors of FedDrive, we employ the lightweight network BiSeNetv2 \cite{yu2021bisenet} for training, accounting for possible lower computational capabilities of the edge devices. 
\paragraph{Data pre-processing:} The images are randomly scaled in the range (0.5, 1.5) and cropped to a $512\times1024$ shape. 
 
\subsubsection{IDDA} \cite{alberti2020idda} is a synthetic dataset for semantic segmentation, specific for the field of autonomous driving. In addition to the annotations for 16 semantic classes, the driving conditions are further characterized by three axes: a city among the 7 available, ranging from Urban to Rural environments; one of 5 viewpoints, simulating different vehicles; an atmospheric condition among 3 possible choices (Noon, Sunset, Rainy), for a total of 105\textit{ domains}. As done for Cityscapes, we refer to FedDrive \cite{fantauzzo2022feddrive} for the federated splits. In the \textit{uniform} distribution of IDDA, each client has access to 48 images randomly drawn from the whole dataset. The \textit{heterogeneous} version is built so that every user only sees a single domain. Two distinct testing scenarios are proposed to assess the generalization abilities of the learned model: one with images belonging to domains likely already seen at training time (``seen'' in Table \ref{tab:ss_idda2} of the main text) and another one containing a never-seen one (``unseen''). The unseen domain either contains images taken in the countryside (``country'') to analyze the \textit{semantic} shift or in rainy conditions (``rainy'') for studying the shift in \textit{appearance}.

\paragraph{Model:} As done for Cityscapes, BiSeNetv2 is the model of choice.

\paragraph{Data pre-processing:} The images are randomly scaled in the range (0.5, 2.0) and cropped to a $512\times928$ shape. 

\subsection{Hyper-parameters Tuning}
We consider a different hyper-parameters setup for each dataset. The final choices of training hyper-parameters are summarized in Table \ref{tab:best_params}. Table \ref{tab:abl_sam} and \ref{tab:params_swa} respectively show the values used for \sam/\asam and \swa.

\begin{table}[t]\centering
\caption{Best performing training parameters}\label{tab:best_params}
\scriptsize
    \begin{tabular}{lccccccc}
    \toprule
    \multirow{2}{*}{Dataset} & Client &\multirow{2}{*}{Batch size} & \multirow{2}{*}{Weight decay} & \multirow{2}{*}{Epochs} & Client &\multirow{2}{*}{Rounds} & Clients\\
    & learning rate& & &&momentum& & per round\\
    \midrule
    \textsc{Cifar10} &  0.01 & 64 & $4\cdot10^{-4}$ & 1 & 0 &$10k$ & $\{5,10,20\}$ \\
    \textsc{Cifar100} & 0.01 & 64 & $4\cdot10^{-4}$ & 1 & 0 &$20k$ & $\{5,10,20\}$\\
    \textsc{Cifar100-Pam} & 0.01 &20 &$4\cdot10^{-4}$ &1-2 &0.9 &$10k$ & $\{10,20\}$\\
    \textsc{Landmarks-User-160k} & 0.1 &64 &$4\cdot10^{-5}$ &5 &0 &$5k$ &10 \\
    \textsc{Cityscapes} (unif.) & 0.05& 8& $5\cdot10^{-4}$& 2& 0.9&$1.5k$ & 5\\
    \textsc{Cityscapes} (het.) &  0.05& 8& $5\cdot10^{-4}$& 2&0.9& $1.5k$ & 5\\
    \textsc{Idda} (country) & 0.1& 8& 0&2 &0.9& $1.5k$ & 5\\
    \textsc{Idda} (rainy) & 0.1& 8& 0&2 &0.9& $1.5k$ & 5\\
    \bottomrule
    \end{tabular}
\end{table}

\subsubsection{CIFAR10 and CIFAR100} For both datasets, the training hyper-parameters follow the choice of \cite{hsu2020federated}. The client learning rate is tuned between the values $\{0.01, 0.1\}$ and set to 0.01, the batch size is 64, $E\in\{1,2\}$ is tested for the number of local epochs and the former is chosen. As for the weight decay the value $4\cdot10^{-4}$ leads to better performances than 0. The local optimizer is SGD with no momentum. No learning rate scheduler is used for simplicity. We optimize the cross-entropy loss. As for the server-side, we compare the behavior of different optimizers (\textit{i.e.} SGD, Adam, AdaGrad) with learning rates in $\{0.001, 0.01, 0.1, 1\}$ (results in Appendix \ref{app:server_optims}), following the setup of \cite{reddi2020adaptive}, and find out that \fedavg, \textit{i.e.} SGD with learning rate 1, is the best choice. When testing \fedavgm, the server-side momentum $\beta=0.9$. As for the other SOTAs, we choose $\mu=0.1$ in \fedprox and $\alpha=0.01$ in \feddyn from $\{0.001,0.01,0.1\}$; in \adabest, we tune $\beta\in\{0.8,0.9\}$ and $\mu\in\{0.01,0.02\}$ and pick $(0.9,0.02)$ for \textsc{Cifar10} and $(0.8,0.02)$ for \textsc{Cifar100}. The training proceeds for $10k$ rounds on \textsc{Cifar10} and $20k$ rounds on \textsc{Cifar100}.

\paragraph{Mixup/Cutout:} Following the setup of \cite{zhang2017mixup}, we fix $\alpha_\text{mixup}=1$, resulting in $\lambda$ uniformly distributed between 0 and 1. As for Cutout instead, we select a cutout size of $16\times16$ pixels for \textsc{Cifar10} and $8\times8$ for \textsc{Cifar100}, as done by \cite{devries2017improved}.  

\paragraph{SAM/ASAM:} The parameter $\rho$ of \sam is searched in $\{0.01,0.02,0.05,0.1,0.2,\\0.5\}$. As for \asam, the value of $\rho$ is tuned in $\{0.05,0.1,0.2,0.5,0.7,1.0,2.0\}$ and $\eta\in\{0.0,0.01,0.1,0.2\}$. The choices made for each dataset and $\alpha$ are shown in Table \ref{tab:abl_sam}. There is no distinction of values as clients vary per round.

\paragraph{SWA:} We test \swa's starting round in $\{5\%, 25\%, 50\%, 75\%\}$ of the rounds budget and as expected \cite{izmailov2018averaging} the best contribution is given if applied from $75\%$ of the training onwards (see Appendix \ref{app:abl}). We set the value of the learning rate $\gamma_1$ to 0.01 and test $\gamma_2\in \{10^{-5}, 10^{-4}, 10^{-3}\}$, selecting $\gamma_2=10^{-4}$. The cycle length $c$ is tested in $\{5,10,20\}$ and set to $10$ for \textsc{Cifar10} and $20$ for \textsc{Cifar100}. Table \ref{tab:params_swa} summarizes the choices. 

\begin{table}[t]\centering
\caption{\fedsam and \fedasam hyper-parameters}\label{tab:abl_sam}
\scriptsize
\setlength\tabcolsep{0.25cm}
    \begin{tabular}{llcccc}
    \toprule
    \multirow{2}{*}{Dataset} & \multirow{2}{*}{Distribution}&\sam & \multicolumn{2}{c}{\asam}\\
    \cmidrule(l){3-3} \cmidrule(l){4-5}
    && $\rho$ & $\rho$ & $\eta$\\
    \midrule
    \multirow{3}{*}{\textsc{Cifar10}} & $\alpha=0$&0.1&0.7&0.2\\
    & $\alpha=0.05$&0.1&0.7&0.2\\
    & $\alpha=100$&0.02&0.05&0.2\\
    \multirow{3}{*}{\textsc{Cifar100}} & $\alpha=0$&0.02&0.5&0.2\\
    & $\alpha=0.5$&0.05&0.5&0.2\\
    & $\alpha=1000$&0.05&0.5&0.2\\
    \textsc{Cifar100-Pam} &$\alpha=0.1$& 0.05&0.5&0/0.2\\
    \textsc{Landmarks-User-160k} &-& 0.05&0.5&0/0.2\\
    \textsc{Cityscapes} & het/unif& 0.01&0.1&0.2\\
    \textsc{Idda} & het/unif&0.01&0.5&0.2\\
    \bottomrule
    \end{tabular}
\end{table}

\subsubsection{CIFAR100-PAM} The hyper-parameters follow the same choice of~\cite{reddi2020adaptive} (see Table~\ref{tab:best_params}). We report accuracy at $5K$ and $10K$ communication rounds.

\paragraph{Mixup/Cutout:} Same as \textsc{Cifar100}.

\paragraph{SAM/ASAM:} We search hyperpameters in the same values as \textsc{Cifar100}. For $\rho$ we found $0.05$ and $0.5$ to be the best values respectively for \sam and \asam in all configurations. For \asam we found that $\eta=0.2$ is working fine when cutout or no augmentations are applied, while $\eta=0$ works best in the case of Mixup.

\paragraph{SWA:} Same as \textsc{Cifar100}.

\subsubsection{Landmarks-User-160k} We start from the hyper-parameters proposed by~\cite{hsu2020federated}. In contrast with the original paper, we found that \fedavgm with momentum $\beta=0.9$ is unstable with $10$ participating clients and requires reducing the server learning rate to $0.1$ to train the model. Better performance and faster convergence can be obtained with $50$ clients per round and $\beta=0.9$. However, we use $10$ clients per round and \fedavg as the baseline because of our limited resources and to maintain consistency with other experiments. All hyper-parameters are described in Table~\ref{tab:best_params}.

\paragraph{SAM/ASAM:} The parameter $\rho$ of \sam is searched in $\{0.01,0.05,0.1\}$. As for \asam, the value of $\rho$ is tuned in $\{0.1,0.3,0.5\}$ and $\eta\in\{0.0,0.1,0.2\}$.

\paragraph{SWA:} We tested both \swa starting at the 75\% and 100\% of training, \textit{i.e.} the $3750$-\textit{th} and $5000$-\textit{th} rounds. We tested different combinations of cycle lengths $c \in \{5, 10, 20\}$ and learning rate $\gamma_2 \in \{ 10^{-2}, 10^{-3} ,10^{-4}\}$. The best performing learning rates $(\gamma_1,\gamma_2)$ are respectively $(10^{-1},10^{-3})$ and the cycle length is $5$.

\subsubsection{Cityscapes and IDDA} For both Cityscapes and IDDA, we maintain the choice of hyper-parameters of \cite{fantauzzo2022feddrive}. The clients' initial learning rate is 0.05 on Cityscapes and 0.1 on IDDA, the weight decay is $5\cdot10^{-4}$ on Cityscapes, while it is not used on IDDA, 2 local epochs, the client optimizer is SGD with momentum 0.9. Differently from \cite{fantauzzo2022feddrive}, we do not use mixed precision, thus the batch size is reduced from 16 to 8. A polynomial learning rate scheduler is applied locally, following \cite{yu2021bisenet}. The optimization is based on the Online Hard-Negative Mining \cite{shrivastava2016training}, which selects the 25\% of the pixels having the highest cross-entropy loss. The training is spanned across $1.5k$ rounds. 

\paragraph{SAM/ASAM:} The parameter $\rho$ of \sam is searched in $\{0.01,0.05,0.1\}$. As for \asam, the value of $\rho$ is tuned in the set $\{0.05,0.1,0.5\}$ and $\eta\in\{0.0,0.1,0.2\}$.

\paragraph{SWA:} Following the setup established for the \textsc{Cifar} datasets, \swa starts at the 75\% of training, \textit{i.e.} the 1125\textit{th} round. The learning rates $(\gamma_1,\gamma_2)$ are respectively $(10^{-1},10^{-3})$ for IDDA and $(5\cdot10^{-2}, 5\cdot10^{-4})$ for Cityscapes. The cycle length is 5 for both datasets.

\begin{table}[t]\centering
\caption{\swa hyper-parameters}\label{tab:params_swa}
\scriptsize
\setlength\tabcolsep{0.25cm}
    \begin{tabular}{lcccc}
    \toprule
    Dataset & $c$ & $\gamma_1$ & $\gamma_2$ & Start round\\
    \midrule
    \textsc{Cifar10} & 10 & $10^{-2}$ & $10^{-4}$ & 7500\\
    \textsc{Cifar100} & 20 & $10^{-2}$ & $10^{-4}$ & 15000\\
    \textsc{Cifar100-Pam} & 5 & $10^{-2}$ & $10^{-4}$ & 15000\\
    \textsc{Landmarks-User-160k} & 5 & $10^{-1}$ & $10^{-3}$ & 3750/5000\\
    \textsc{Cityscapes} & 5&$5\cdot10^{-2}$& $5\cdot10^{-4}$ & 1125\\
    \textsc{Idda} &5& $10^{-1}$ &$10^{-3}$& 1125\\
    \bottomrule
    \end{tabular}
\end{table}

\subsection{Plotting the Loss Landscapes}
In the main text, we introduced both 2-D (Fig. \ref{fig:fedavg_convergence} of the main text) and 3-D plots of the loss landscapes (Fig. \ref{fig:loss_landscape} of the main text). Implementation details follow.

\subsubsection{2D Loss Landscape} Following the indications of \cite{garipov2018loss,mirzadeh2020linear}:
\begin{enumerate}
    \item We choose three weight vectors $\theta_1, \theta_2, \theta_3$ and use them to obtain two basis vectors $\vec u = (\theta_2-\theta_1)$ and $\vec{v} = (\theta_3-\theta_1) - \frac{\langle\theta_3-\theta_1, \theta_2-\theta_1\rangle}{||\theta_2-\theta_1||^2}\cdot(\theta_2-\theta_1)$.
    \item Then, the normalized vectors $\hat{u}=\nicefrac{u}{||u||}$ and $\hat{v}=\nicefrac{v}{||v||}$ form an orthonormal basis in the plain containing $\theta_1, \theta_2, \theta_3$. 
    \item We now define a Cartesian grid of $N\times N$ points in the basis $\hat{u},\hat{v}$. In our case, $N=21$. 
    \item For each point of the grid, the corresponding weights are computed and the loss is consequently evaluated with the resulting network. For each point $P$ of the grid having coordinates $(x,y)$, the corresponding weights are computed as $P = \theta_1 + x \cdot \hat{u} + y \cdot \hat{v}$. As a consequence, $\theta_1$ is the reference and can be found in the origin $(0,0)$.
\end{enumerate}
We adapted the code of \cite{garipov2018loss}\footnote{\url{https://github.com/timgaripov/dnn-mode-connectivity}} to our scenario.

\subsubsection{3D Loss Landscape} The plots in Fig. \ref{fig:loss_landscape} in the main text are generated using the code of \cite{visualloss}\footnote{\url{https://github.com/tomgoldstein/loss-landscape}}, modified to fit our datasets and models. Given a network architecture and its pre-trained parameters, the loss surface is computed along random directions near the optimal parameters.  

\subsection{Computing Hessian Eigenvalues}
We refer to \cite{hessian-eigenthings} for computing both the local and the top 50 Hessian eigenvalues (Figs. \ref{fig:clients_eigs},{\color{red}{5}}
in the main text) with Stochastic Power Iteration method \cite{xu2018accelerated} with maximum 20 iterations per run.

\section{Results on Corrupted CIFAR10 and CIFAR100}
\label{app:corr}
\begin{figure}[!t]
    \centering
    \subfloat[][\textsc{Cifar10-C} $\alpha=0$]{\includegraphics[width=.49\linewidth]{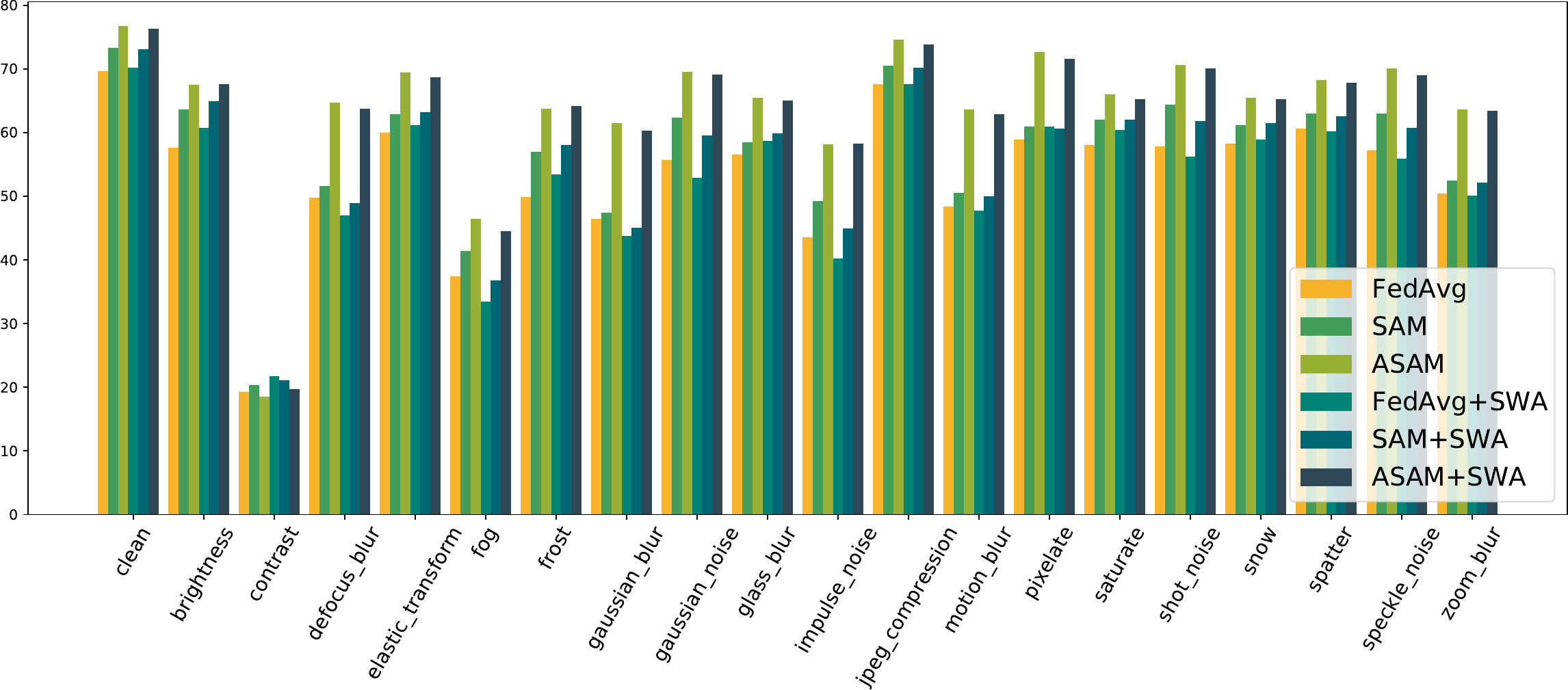}}
    \subfloat[][\textsc{Cifar100-C} $\alpha=0$]{\includegraphics[width=.49\linewidth]{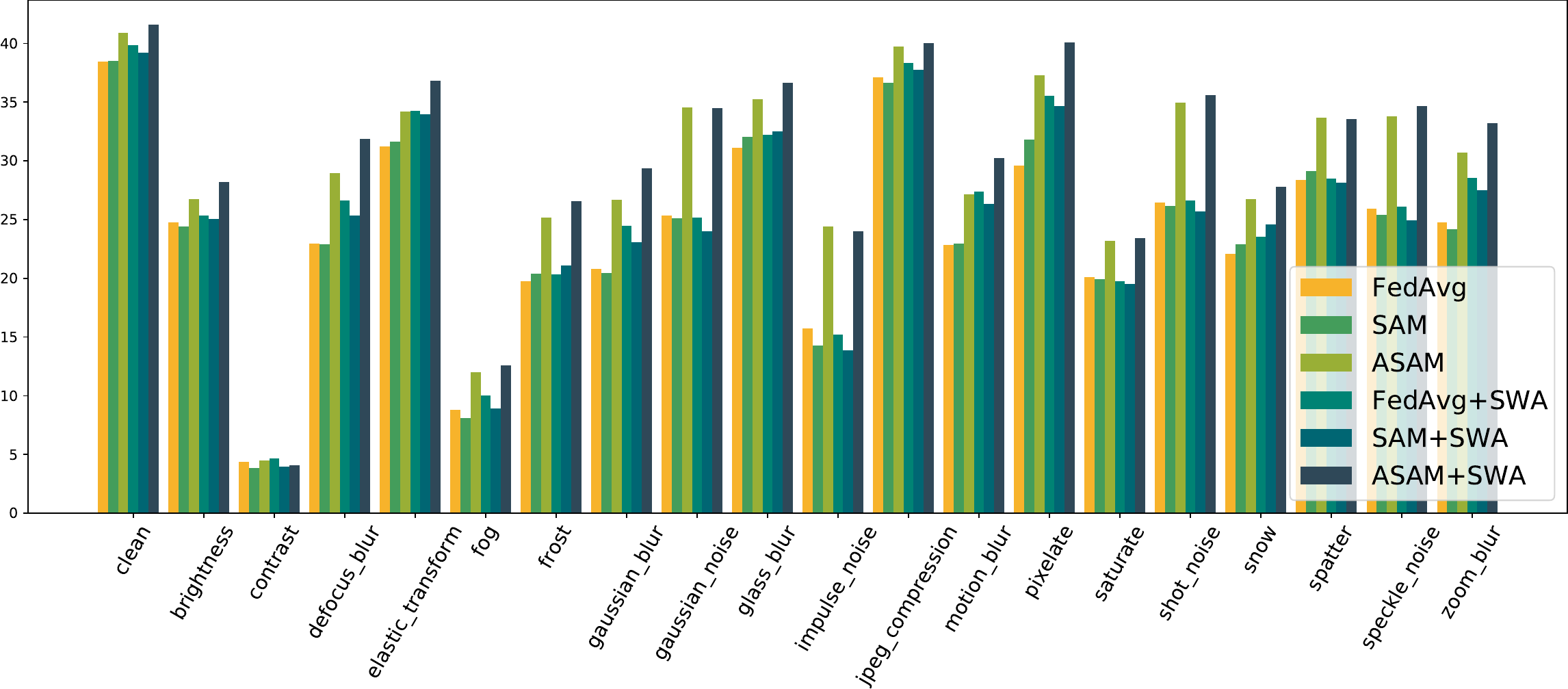}}\\
    \subfloat[][\textsc{Cifar10-C} $\alpha=0.5$]{\includegraphics[width=.49\linewidth]{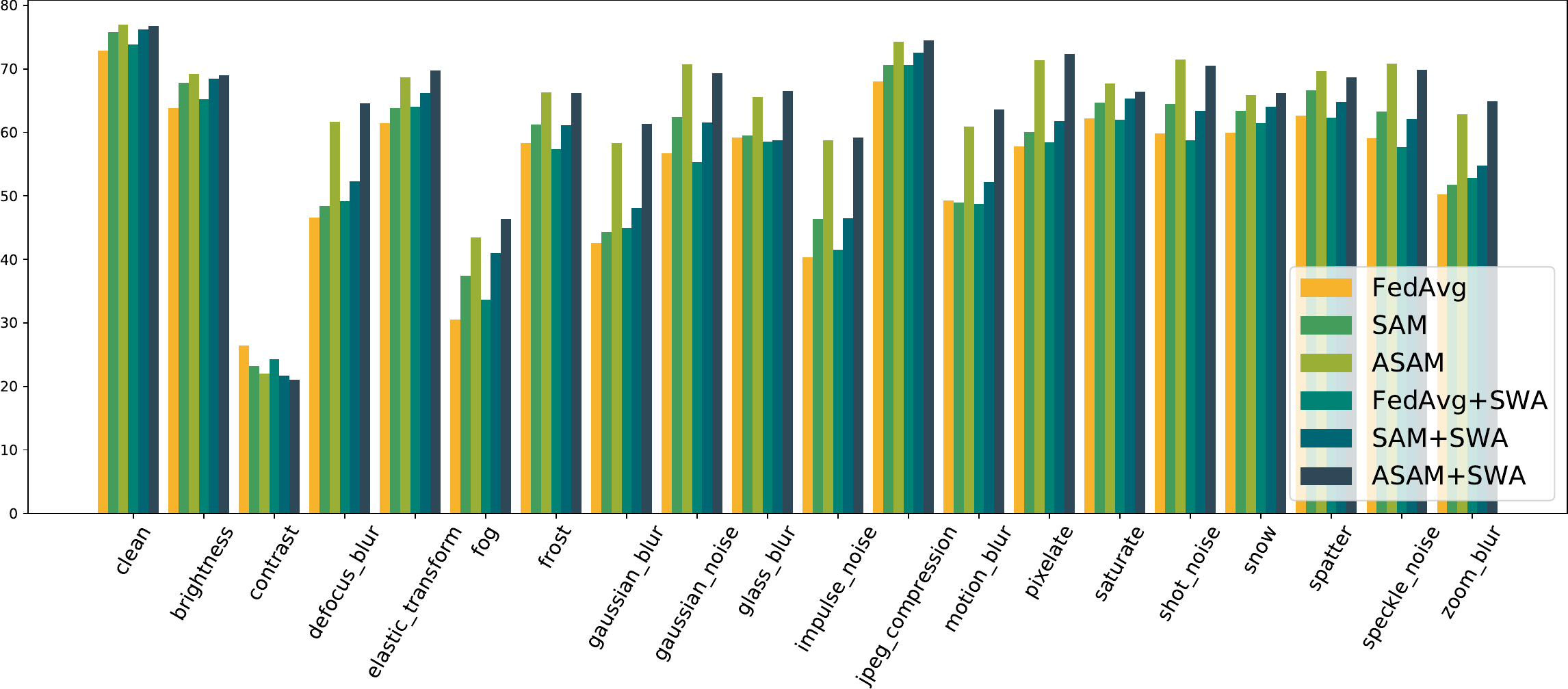}}
    \subfloat[][\textsc{Cifar100-C} $\alpha=0.5$]{\includegraphics[width=.49\linewidth]{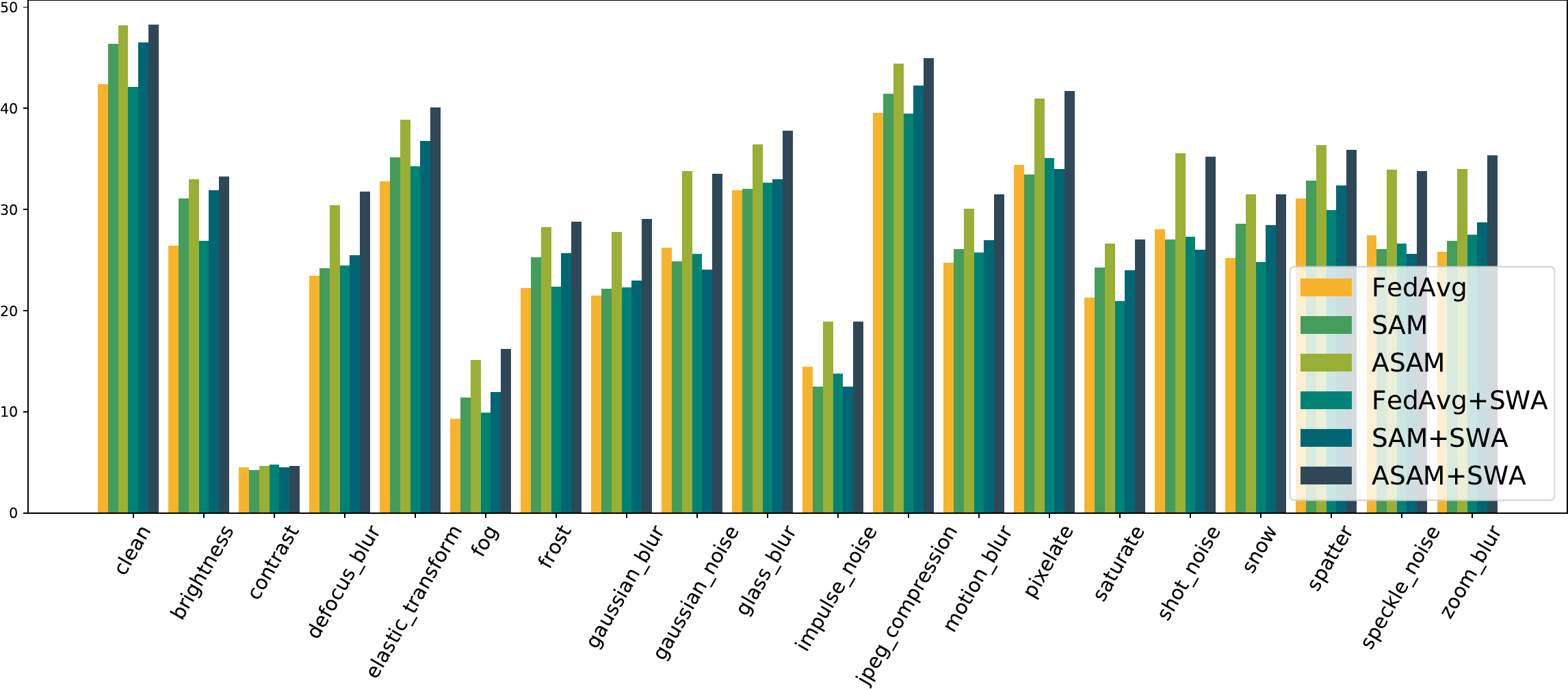}}\\
    \subfloat[][\textsc{Cifar10-C} $\alpha=100$]{\includegraphics[width=.49\linewidth]{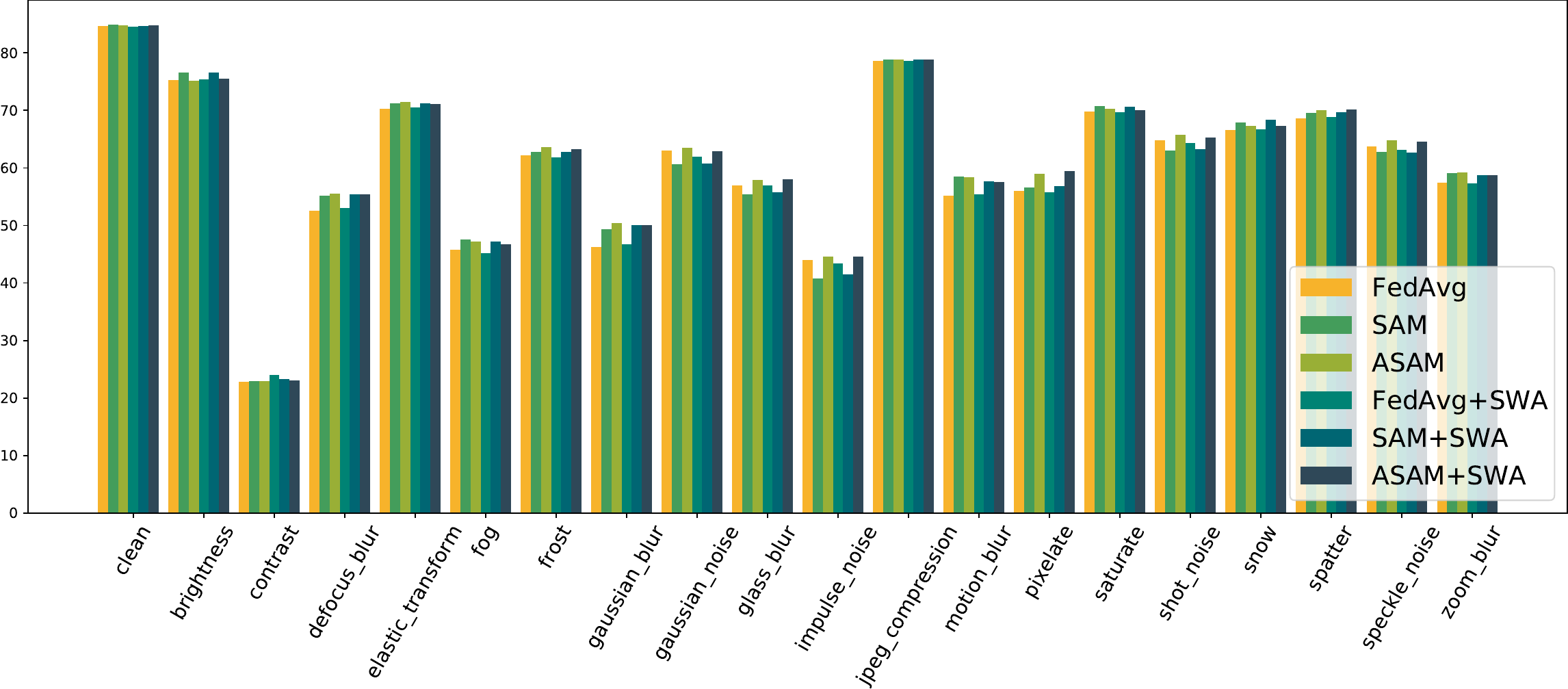}}
    \subfloat[][\textsc{Cifar100-C} $\alpha=1000$]{\includegraphics[width=.49\linewidth]{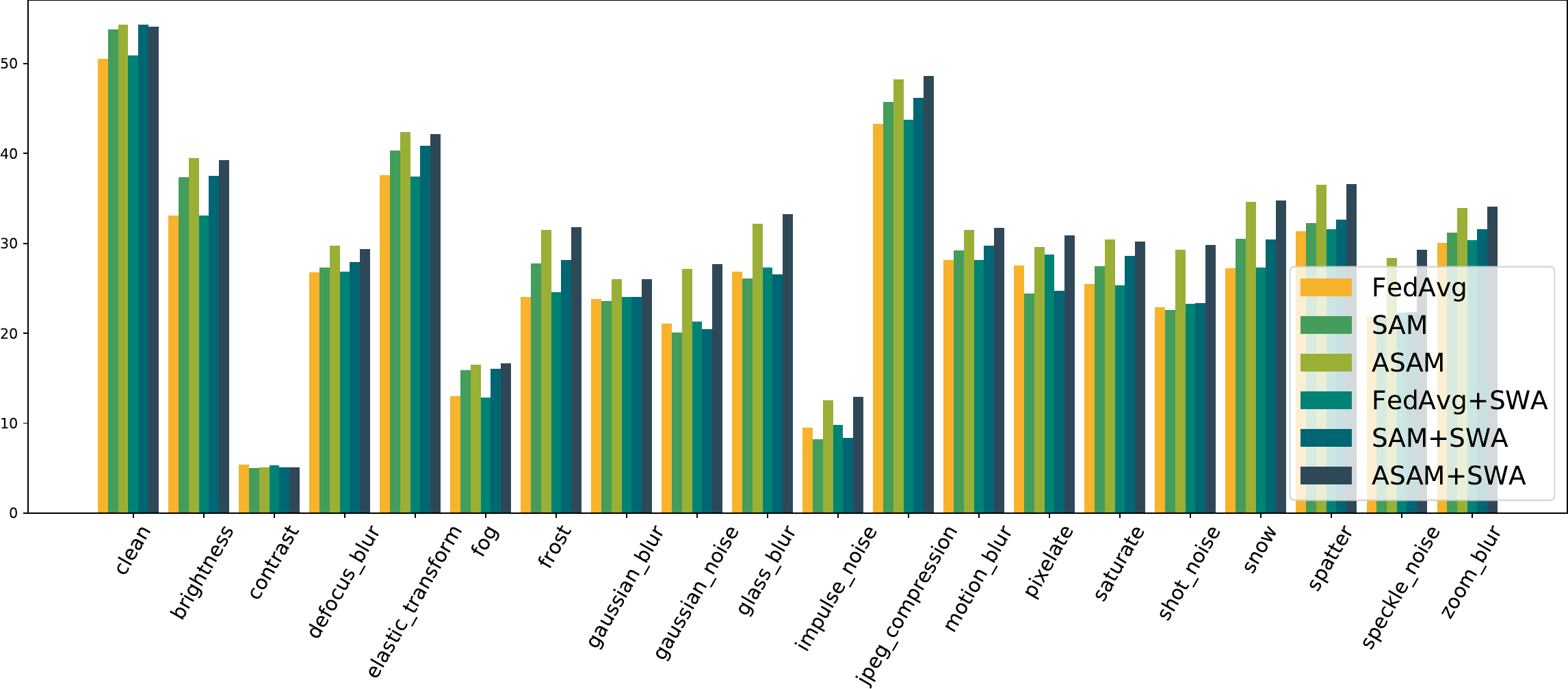}}
    \caption{\footnotesize{Domain generalization in FL. Results with 20 clients, severity level 5 on \textsc{Cifar10-C} and \textsc{Cifar100-C}.}}
    \label{fig:cifar-c-supp}
    \vspace{-0.5cm}
\end{figure}

In Fig. \ref{fig:cifar-c-supp}, we compare the performance obtained by \fedavg, \fedsam, \fedasam, \fedavg+ \swa, \fedsam+ \swa and \fedasam+ \swa on \textsc{Cifar10-C} and \textsc{Cifar100-C} as $\alpha$ varies. All results tell us that \asam (alone or combined with \swa) is the algorithm with the best generalization capabilities, as already seen in Sec. \ref{sec:dg} of the main text. 
\captionsetup[subfloat]{font=scriptsize,labelformat=empty}

\section{Ablation Studies}
\label{app:abl}
In this Section, we present our ablation studies on server-side optimizers, \sam, \asam and \swa, moved from the main text due to space constraints.

\subsection{Ablation Study on Server-Side Optimizers}
\label{app:server_optims}
\input{tables/server_opts}
To choose the best server-side optimizer, we test SGD, Adam and AdaGrad on the heterogeneous ($\alpha=0$) and homogeneous ($\alpha=1k$) versions of \textsc{Cifar100} with 5 clients per round. Following \cite{reddi2020adaptive}, we set $\beta_1=\beta_2=0$ for AdaGrad and $\beta_1=0.9, \beta_2=0.99$ for Adam. As Table \ref{tab:server_optim} shows, SGD with learning rate 1, \textit{i.e.} \fedavg, is certainly the best choice to have acceptable performances both in the homogeneous scenario and above all in the heterogeneous one.

\subsection{Ablation Study on \sam and \asam}
\begin{figure}[]
    \centering
    \subfloat[][]{\includegraphics[width=.33\linewidth]{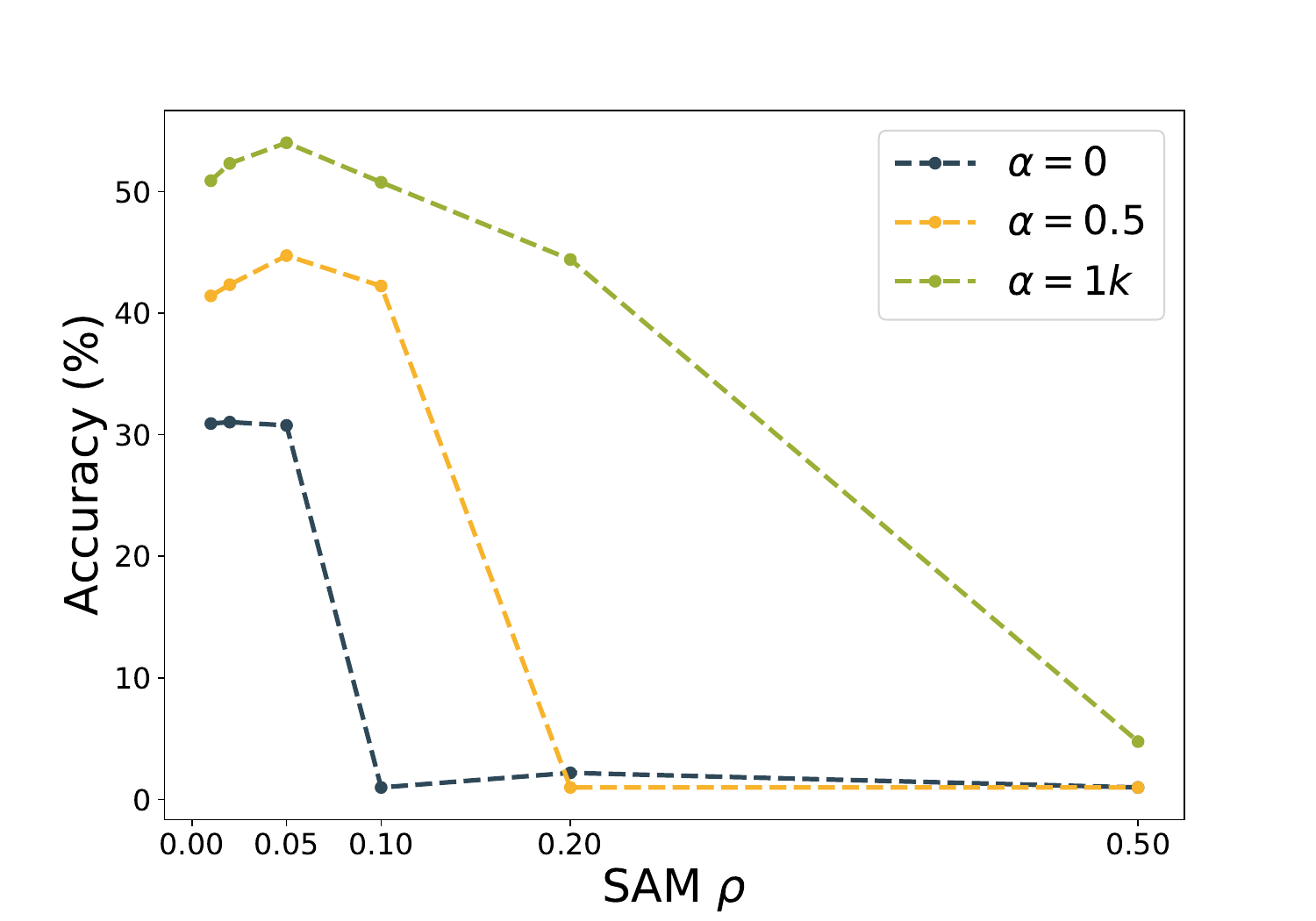}}
    \subfloat[][]{\includegraphics[width=.33\linewidth]{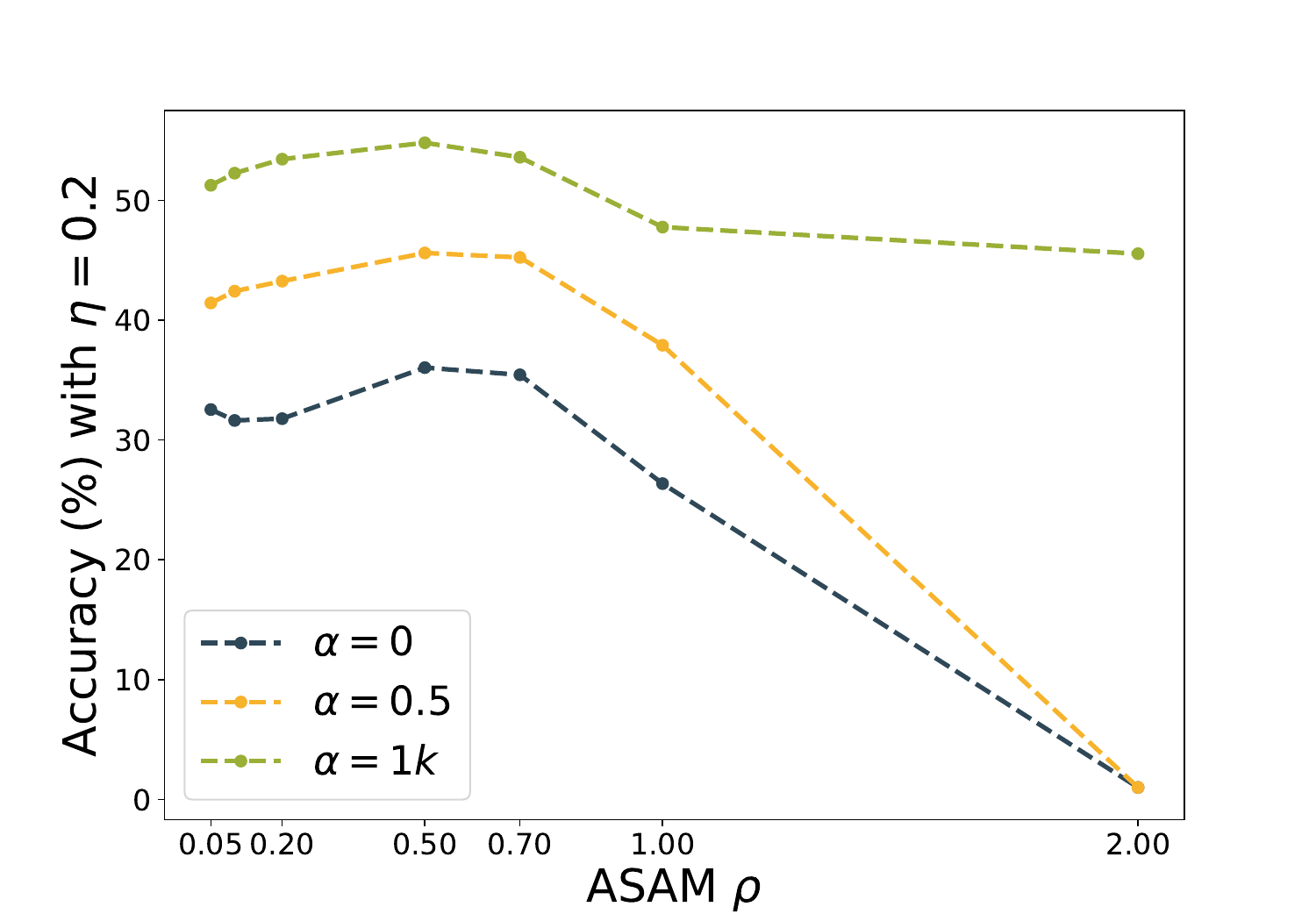}}
    \subfloat[][]{\includegraphics[width=.33\linewidth]{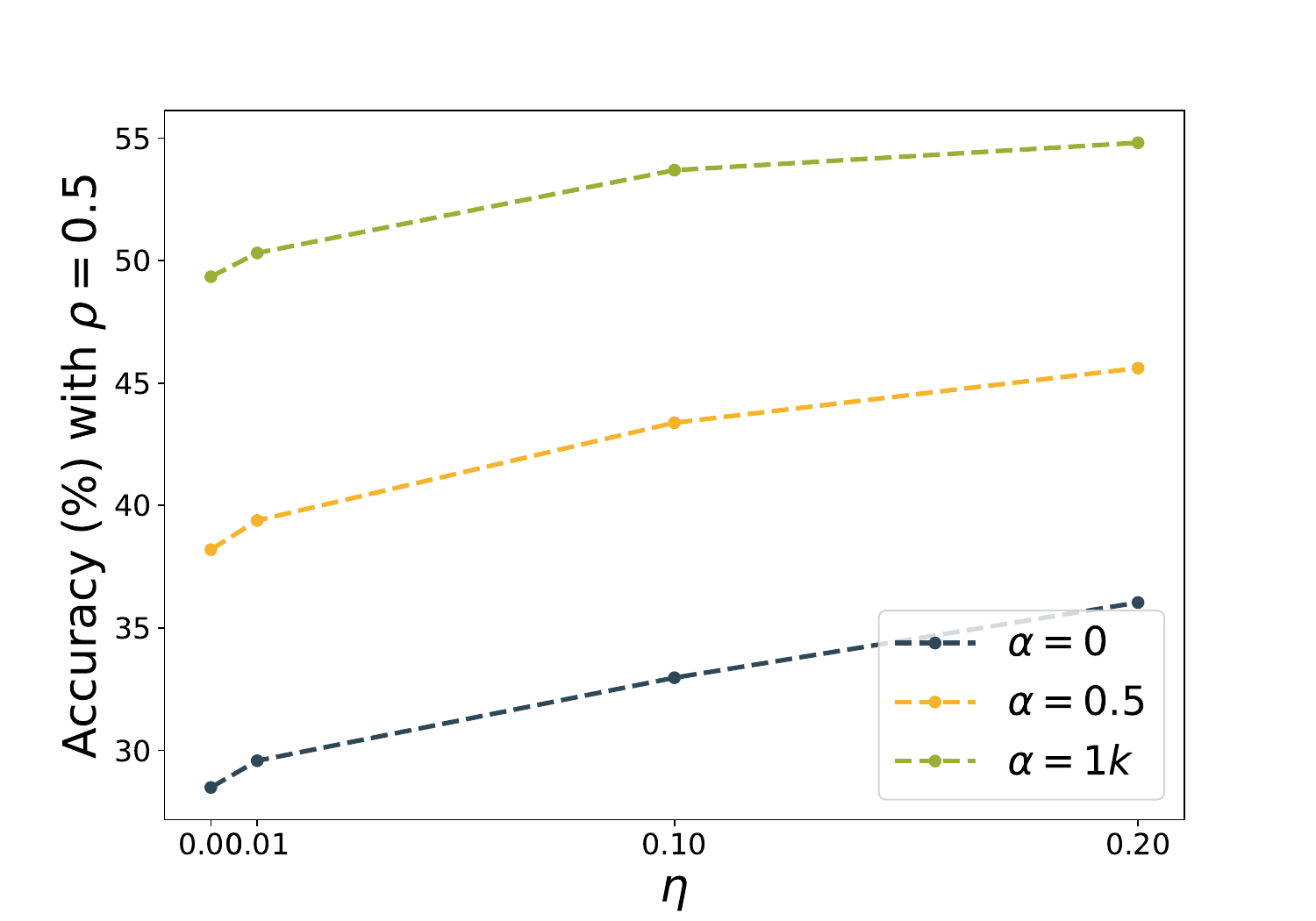}}
    \caption{Results on \textsc{Cifar100}, 5\% clients participation. \textbf{(a)} Sensitivity to \sam's parameter $\rho$. \textbf{(b)-(c)} Sensitivity to \asam's parameters $\rho$ (with fixed $\eta=0.2$) and $\eta$ (with fixed $\rho=0.5$) as $\alpha$ varies.}
    \label{fig:sensitivity}
\end{figure}

We present here an analysis on the sensitivity of the model to the hyper-parameters $\rho$ and $\eta$ in \asam and $\rho$ in \sam (Fig. \ref{fig:sensitivity}), having as a reference the setting with 5\% clients participation on \textsc{Cifar100}. Regardless of the distribution, we can see that high values of \sam's $\rho$ lead to a fast decline in performance (Fig. \ref{fig:sensitivity}{\color{red}{a}}), meaning that the algorithm handles smaller neighborhoods better. On the other hand, \asam allows us to have more freedom and expand the size of the neighborhood up to the value of $\rho=0.5$ (Fig. \ref{fig:sensitivity}{\color{red}{b}}), index of the greater robustness of the method. In Fig. \ref{fig:sensitivity}{\color{red}{c}}, we notice that the performances improve linearly as $\eta$ increases, where $\eta$ is a hyper-parameter balancing the trade-off between stability and adaptivity.

\captionsetup[subfloat]{font=scriptsize,labelformat=parens}

\subsection{Ablation Study on \swa}
\label{app:abl_swa}
\swa adds two new concepts to the standard federated training: the average of stochastic weights collected along the trajectory of SGD (Eq. \ref{math:swa}) and the cyclical learning rate (Eq. \ref{math:lr_swa}), which decreases from $\gamma_1$ to $\gamma_2$ according to the cycle length $c$, transmitted as additional information to the clients of each round. Our ablation studies aim to understand which of these two components has the greatest impact on the achieved stability and increased model performance. We compare the results obtained by \swa with $c>1$ with those reached when the learning rate is kept constant, \textit{i.e.} $c=1$, and when the server-side average of the collected weights is not applied while maintaining $c>1$, \textit{i.e.} changing only the clients' learning rate cyclically (Table \ref{tab:abl_swa}). We point out that using $c=1$ and not applying the average brings us back to the standard federated setting. We discover that the server-side average gives the major contribution, which helps in stabilizing learning, while the cycle length does not particularly affect the results. Since the best results in the most difficult scenarios (\textit{i.e.} low value of both $\alpha$ and number of participating clients on \textsc{Cifar100}) are reached when $c>1$, we prefer the cyclical learning rate to the constant one in further experiments.

In addition, in Table \ref{tab:abl_start} we report the differences in results when applying \swa from $\{5\%, 25\%, 50\%, 75\%\}$ of the training onwards on \fedavg with 5 clients per round, showing that a longer pre-training of the network leads to the greater effectiveness of this algorithm.

\begin{table}[t]\centering
\caption{\swa ablation study: comparison between cyclical ($c>1$) and constant learning rate ($c=1$) and contribution given by averaging stochastic weights. Highlighted in bold the best result for each combination (Algorithm, $\alpha$, participating clients).}\label{tab:abl_swa}
\scriptsize
    \begin{tabular}{llcccccccccccc}
    \toprule
    \multirow{2}{*}{Dataset} & \multirow{2}{*}{Algorithm} & \multirow{2}{*}{WeightsAvg} & \multirow{2}{*}{$c$} & \multicolumn{3}{c}{$\alpha=0$} & \multicolumn{3}{c}{$\alpha=0.5/0.05$} & \multicolumn{3}{c}{$\alpha=1k/100$}\\
    \cmidrule(l){5-7} \cmidrule(l){8-10} \cmidrule(l){11-13}
    & & && $5cl$ & $10cl$ & $20cl$ & $5cl$ & $10cl$ & $20cl$ & $5cl$ & $10cl$ & $20cl$\\
    \midrule
    \multirow{12}{*}{\textsc{Cifar100}} & \fedavg & \multirow{3}{*}{\ding{51}} & \multirow{3}{*}{20} & \textbf{39.34 }&39.74  & 39.85 & \textbf{43.90} &   \textbf{44.02}  &42.09 & 50.98&50.87 &50.92\\
    & \fedsam &  &  &  \textbf{39.30} &\textbf{ 39.51} & 39.24 & \textbf{47.96} & \textbf{46.76}  &\textbf{46.47} &{53.90} &53.67 &\textbf{54.36}\\
    & \fedasam & && {{42.01}} & {\textbf{42.64}} & {{41.62}}  &{\textbf{49.17}} &  {\textbf{48.72}} & {{48.27}}& 53.86& {54.79}&54.10\\\cmidrule{2-13}
    & \fedavg & \multirow{3}{*}{\ding{51}} & \multirow{3}{*}{1} & 38.86 & \textbf{39.82} & \textbf{40.19} & 43.86&43.93&\textbf{42.67}&\textbf{51.33}&\textbf{51.05}&\textbf{51.11}\\
    & \fedsam & & & 38.58 & 39.20 & \textbf{39.37}&47.29&46.34&46.40&53.88&\textbf{53.70}&\textbf{54.36}\\
    & \fedasam & & &\textbf{42.50}&42.40&\textbf{41.76}&48.67&48.50&47.95&{54.16}&\textbf{55.07}&{54.19}\\\cmidrule{2-13}
    & \fedavg & \multirow{3}{*}{\ding{55}}& \multirow{3}{*}{20} & 30.68&34.86&37.42&40.34&42.40&41.89&50.06&50.21&50.81\\
    & \fedsam & & & 31.51&35.87&37.81&44.08&45.80&46.43&53.76& 53.46&54.28\\
    & \fedasam & & & 36.85&39.76&41.03&46.34&48.06&\textbf{48.38}&{54.21}&55.06&{54.22}\\\cmidrule{2-13}
    & \fedavg & \multirow{3}{*}{\ding{55}} & \multirow{3}{*}{1} & 30.25 & 36.74 & 38.59 & 40.43 & 41.27 &  42.17 & 49.92 & 50.25& 50.66\\
    & \fedsam & & & 31.04 & 36.93 & 38.56 & 44.73 & 44.84 & 46.05  &\textbf{54.01} & 53.39 &53.97\\
    & \fedasam & & & {36.04} & {39.76} & {40.81} & {45.61} &{46.58} & {47.78}  & {\textbf{54.81}} & 54.97&{\textbf{54.50}}\\\midrule
    \multirow{12}{*}{\textsc{Cifar10}} & \fedavg & \multirow{3}{*}{\ding{51}} & \multirow{3}{*}{10} & 69.71 & 69.54 & 70.19 & 73.48 &  72.80   &\textbf{73.81} &84.35 & 84.32&84.47\\
    & \fedsam & & & 74.97 &73.73  & 73.06 & {{76.61}} & 75.84  & 76.22&84.23 & 84.37&84.63\\
    & \fedasam & & & {{76.44}} & {\textbf{75.51}} & {{76.36}}  & 76.12 & {{76.16}}  & {76.86}& {{84.88}}&{{84.80}} &{\textbf{84.79}}\\\cmidrule{2-13}
    & \fedavg & \multirow{3}{*}{\ding{51}} & \multirow{3}{*}{1} & \textbf{69.88}&\textbf{69.83}&\textbf{70.72}&\textbf{73.91}&\textbf{73.12}&73.07&\textbf{84.90}&84.47&\textbf{84.67}\\
    & \fedsam & & & \textbf{75.17}&\textbf{74.00}&\textbf{73.53}&\textbf{76.93}&\textbf{76.06}&\textbf{76.55}&\textbf{84.53}&84.54&84.77\\
    & \fedasam & & & \textbf{76.80}&75.48&\textbf{76.84}&\textbf{76.87}&\textbf{76.30}&\textbf{77.55}&\textbf{85.09}&\textbf{85.06}&84.73\\\cmidrule{2-13}
    & \fedavg & \multirow{3}{*}{\ding{55}} & \multirow{3}{*}{10} & 61.41&63.96&67.39&67.17&69.88&72.19&84.18&84.15&84.45\\
    & \fedsam & & & 70.66&71.14&73.04&73.93&74.96&76.20&84.23&84.40&84.69\\
    & \fedasam & & & 75.07&74.87&76.37&75.37&76.17&77.14&84.68&84.72&84.71\\\cmidrule{2-13}
    & \fedavg & \multirow{3}{*}{\ding{55}} & \multirow{3}{*}{1} &  65.00 & 65.54 & 68.52 & 69.24 & 72.50 & 73.07  &84.46 & \textbf{84.50}& 84.59\\
    & \fedsam & & & 70.16 & 71.09 & 72.90 & 73.52 & 74.81 & 76.04  &84.58 & \textbf{84.67} &{\textbf{84.82}}\\
    & \fedasam & & &{73.66} & {74.10} & {76.09} & {75.61} & {76.22} & {{76.98}}  & {84.77} &{84.72} &84.75\\
    \bottomrule
    \end{tabular}
\end{table}
 
\begin{table}[]\centering
\caption{\swa ablation study: comparison between \swa starting rounds when using \fedavg with 5 clients per round}\label{tab:abl_start}
\scriptsize
\setlength\tabcolsep{0.3cm}
    \begin{tabular}{lccccc}
    \toprule
    \multirow{2}{*}{Dataset} &\multirow{2}{*}{$c$} & \multirow{2}{*}{Start round}  &   \multicolumn{3}{c}{Test Accuracy (\%)} \\
    \cmidrule(l){4-6} 
    & & & $\alpha=0$ &  $\alpha=0.5/0.05$ &  $\alpha=1k/100$\\
    \midrule
    \multirow{4}{*}{\textsc{Cifar100}} & \multirow{4}{*}{20} & 1000& 24.53&34.52&49.38\\
    & & 5000&30.66&39.71&\textbf{51.52}\\
    & & 10000&36.21&42.55&51.01\\
    & & 15000&\textbf{39.34}&\textbf{43.90}&50.98\\
    \midrule
    \multirow{4}{*}{\textsc{Cifar10}} & \multirow{4}{*}{10} & 500& 55.57&60.50&79.09\\
    & & 2500&60.34&65.72&81.49\\
    & & 5000&66.22&70.55&83.79\\
    & & 7500&\textbf{69.71}&\textbf{73.48}&\textbf{84.35}\\
    \bottomrule
    \end{tabular}
\end{table}

\section{Tables Omitted in the Main Text}
\subsection{Heterogeneous FL Benefits Even More from Flat Minima - Additional Material}
\label{app:omitted_benefits_sam}
Table \ref{tab:centr2} completes the analysis introduced in Sec. \ref{sec:cifar} regarding the gains obtained in the federated scenario w.r.t. the centralized one. Here we report the results for $\alpha\in\{0,1k\}$. As noted for $\alpha=0$ (Table \ref{tab:centr} in the main text), data augmentations fail in the federated heterogeneous scenarios ($\alpha\in\{0,0.5\}$), but reasonably work in the homogeneous ones. 
\input{tables/centralized2}

\subsection{Data Augmentations with CIFAR10}
\label{app:augm}
Here we show the results obtained when applying Mixup and Cutout to \textsc{Cifar10} as the value of $\alpha$, clients participation and algorithm change (Table \ref{tab:augms_cifar10}). As demonstrated for \textsc{Cifar100} (Sec. \ref{sec:cifar}), data augmentations do not improve generalization in a federated context, but on the contrary they seem to inhibit learning, leading to sometimes even worse results than \fedavg. 
\input{tables/cifar10_data_augms}

\section{Figures Omitted in the Main Text}
\begin{figure}
    \centering
    \subfloat[][\textsc{Cifar100}]{\includegraphics[width=.5\linewidth]{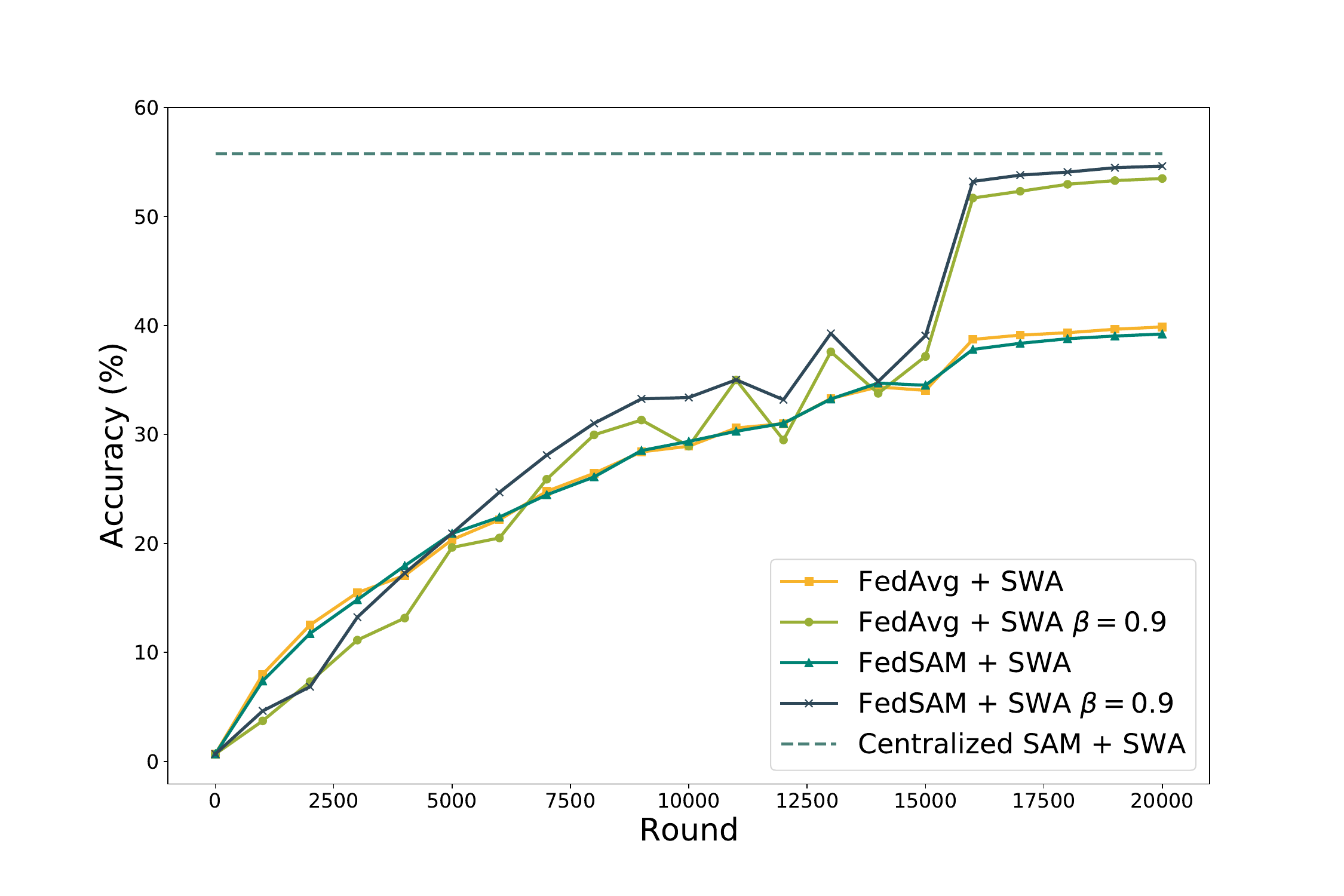}}
    \subfloat[][\textsc{Cifar10}]{\includegraphics[width=.5\linewidth]{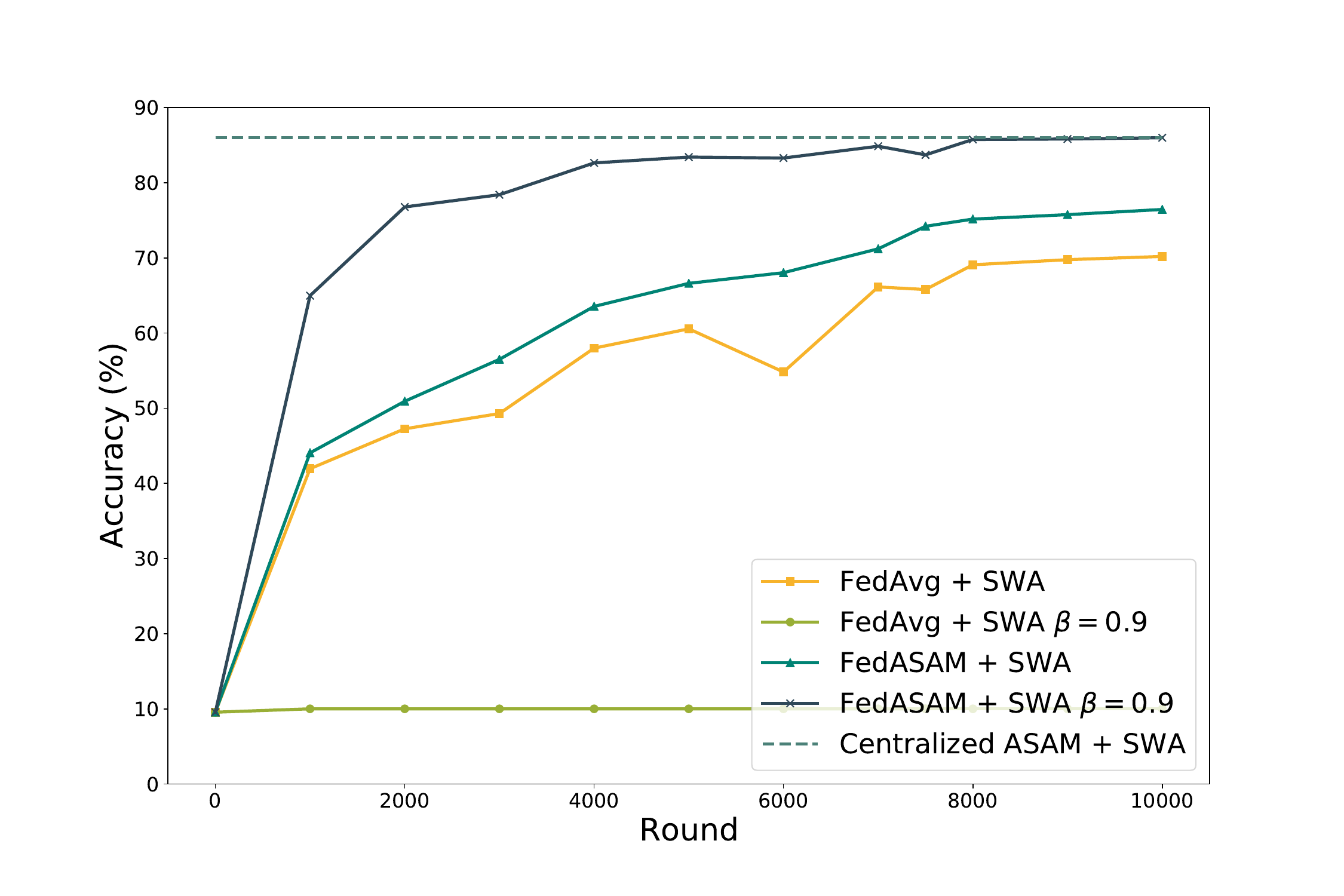}}
    \caption{Convergence plots with $\alpha=0$, 20 clients. When combining \fedavgm or \fedsam (\textsc{Cifar100})/\fedasam (\textsc{Cifar10}) with \swa, convergence is reached even in the most heterogeneous scenarios. \fedavgmswa applied to \textsc{Cifar10} fails to learn, while adding momentum to \fedasam significantly speeds up training.}
    \label{fig:convergence}
\end{figure}

\begin{figure}
    \centering
    \subfloat[][\textsc{Cifar100}]{\includegraphics[width=.5\linewidth]{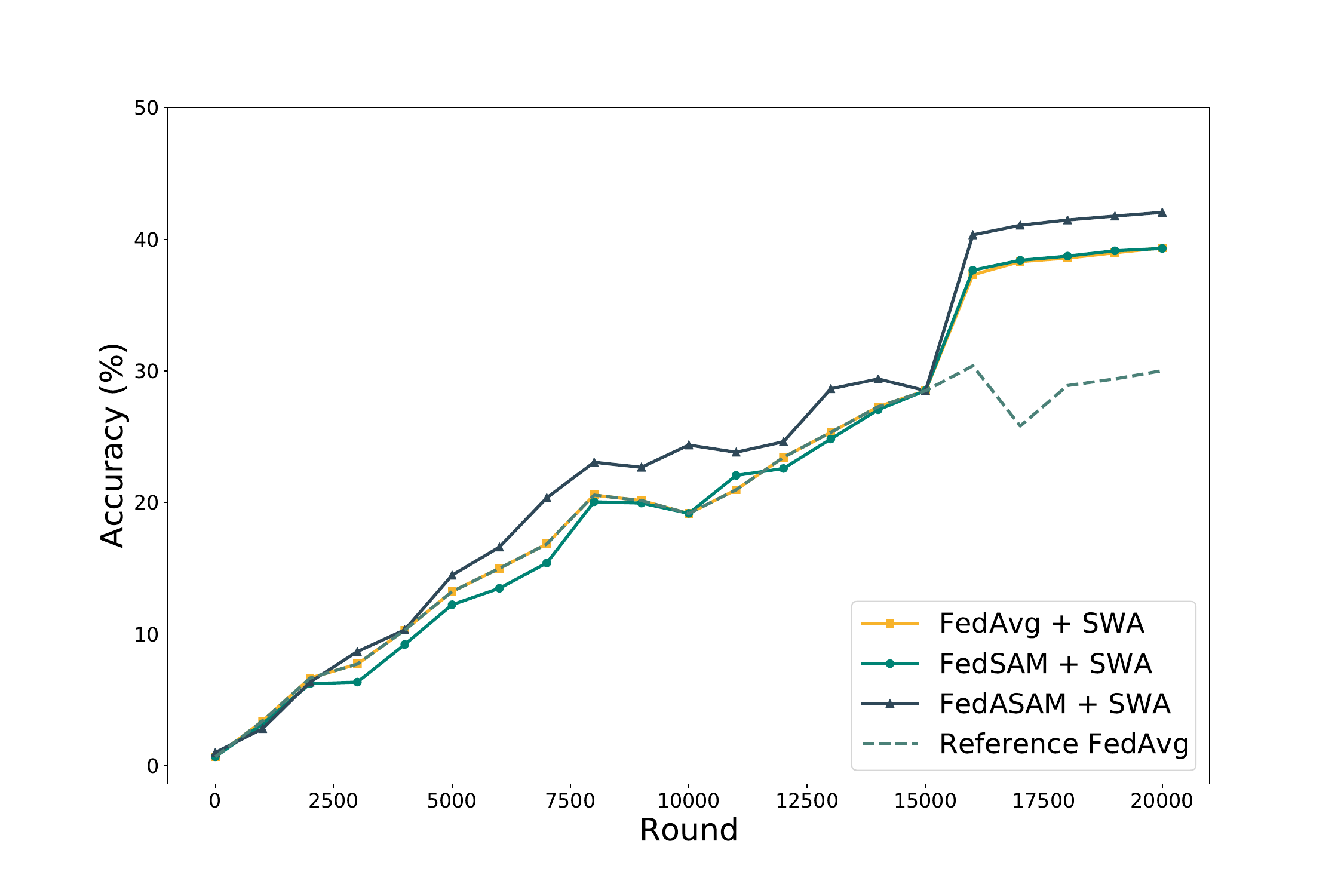}}
    \subfloat[][\textsc{Cifar10}]{\includegraphics[width=.5\linewidth]{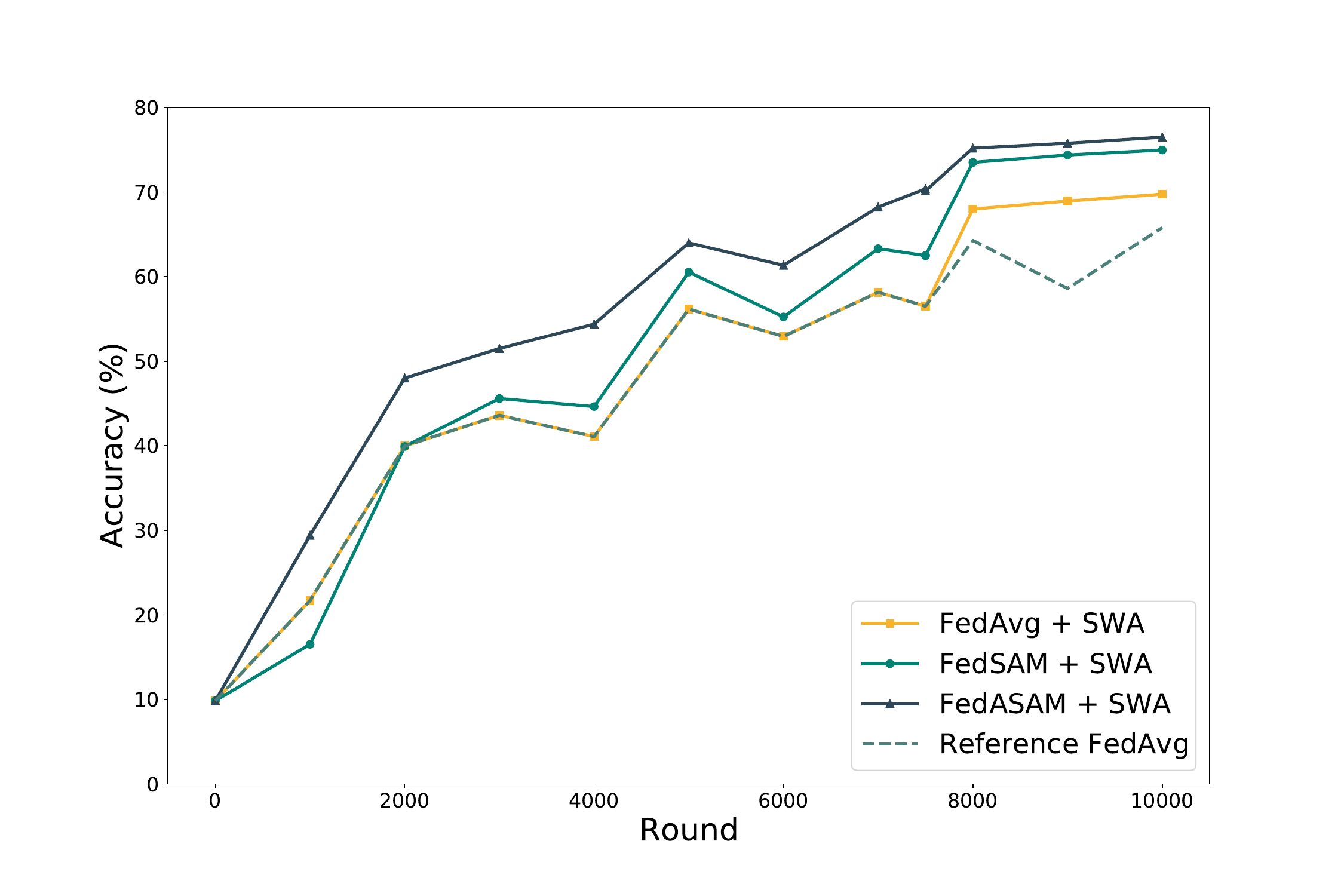}}
    \caption{Convergence plots with $\alpha=0$, 5 clients, highlighting the positive gap in performance and the stability introduced by \swa (both if applied on \fedavg but especially on \fedasam) in the most difficult setting.}
    \label{fig:conv_k5}
\end{figure}

\begin{figure}
    \centering
    \includegraphics[width=.4\linewidth]{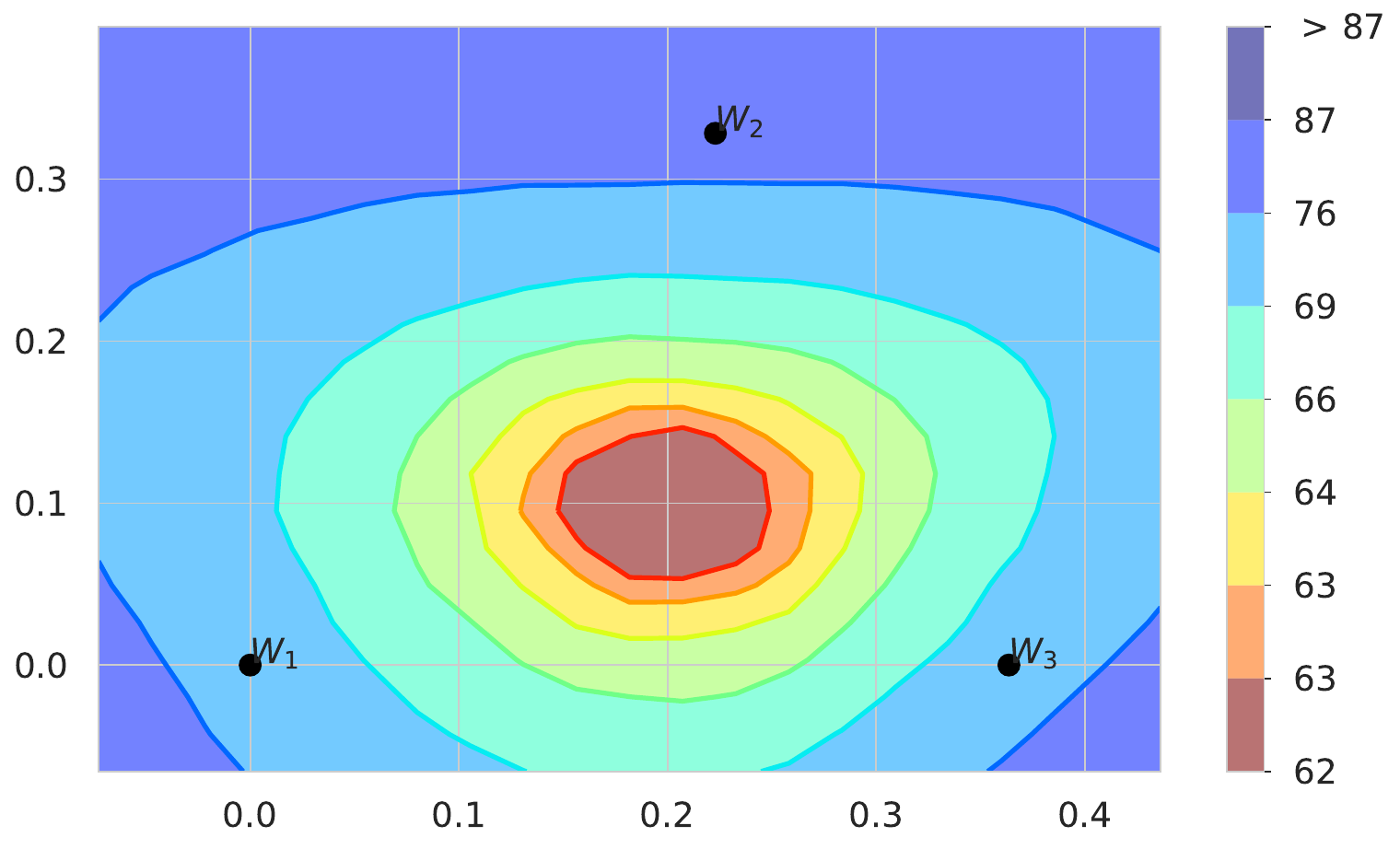}
    \caption{Test error surface computed on \textsc{Cifar100} using three distinct local models trained with $\alpha=0.5$ for $20k$ rounds.}
    \label{fig:a05_client_conv}
\end{figure}

\begin{figure}[!t]
    \centering
    \subfloat[][]{\includegraphics[width=.33\linewidth]{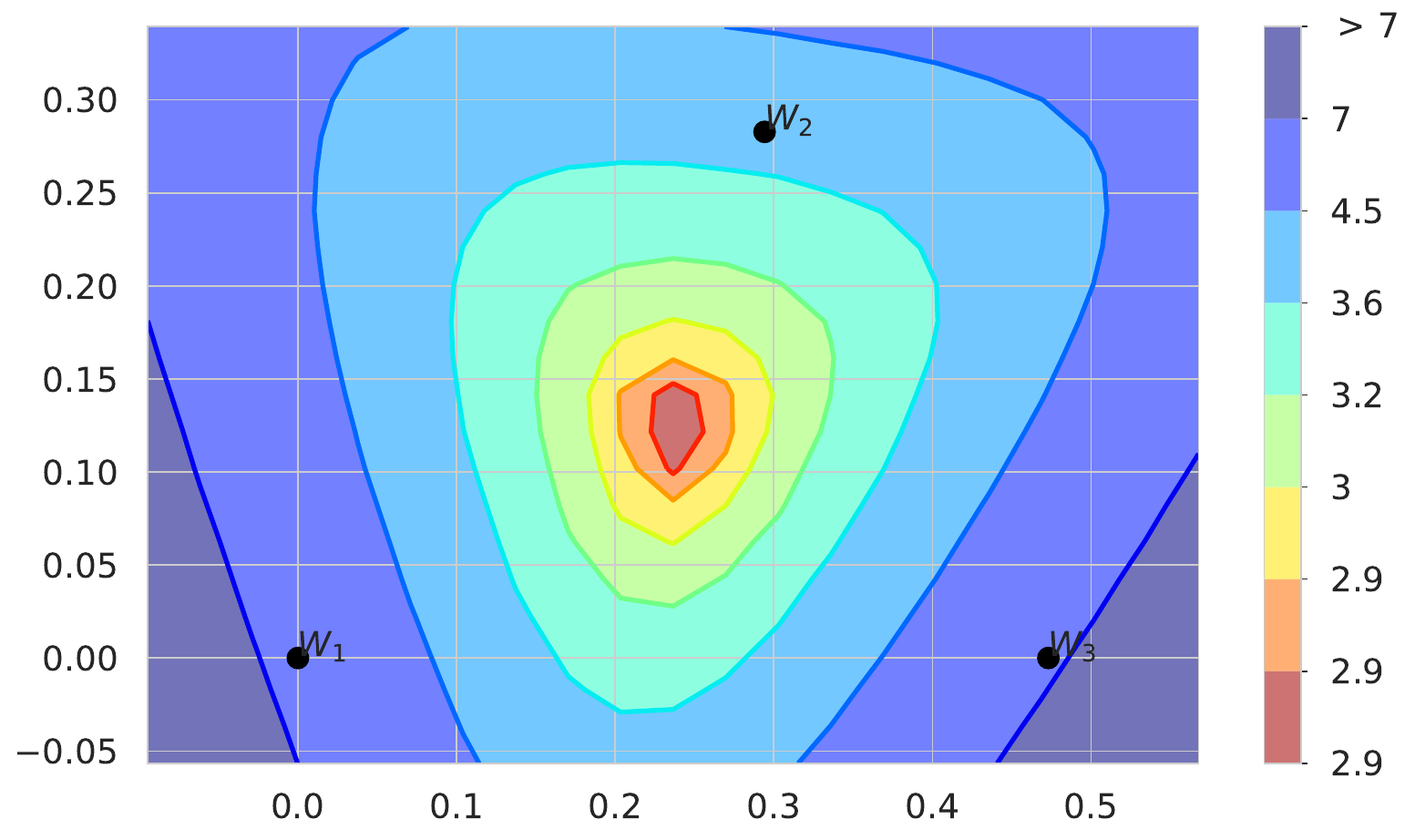}}
    \subfloat[][]{\includegraphics[width=.33\linewidth]{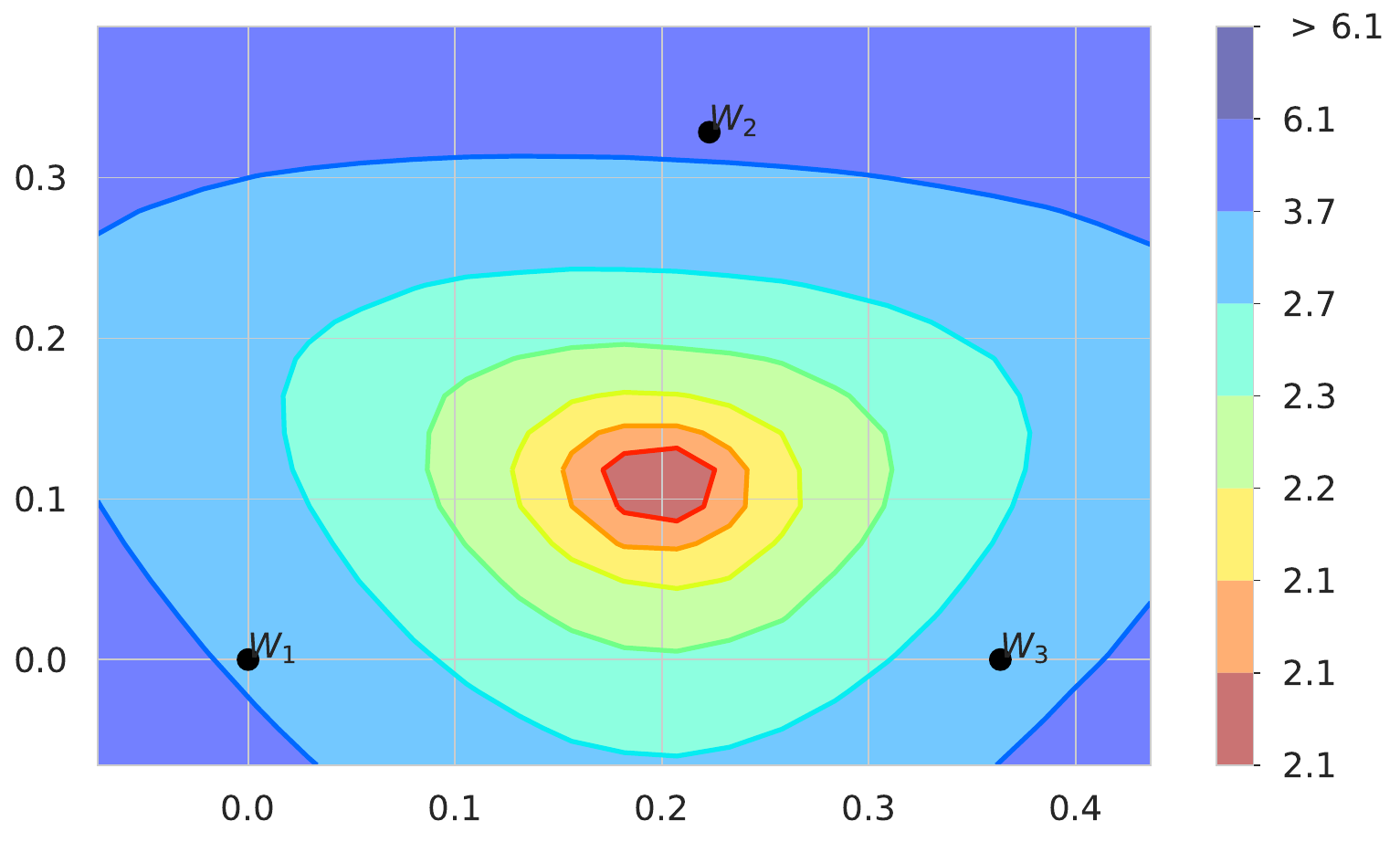}}
    \subfloat[][]{\includegraphics[width=.33\linewidth]{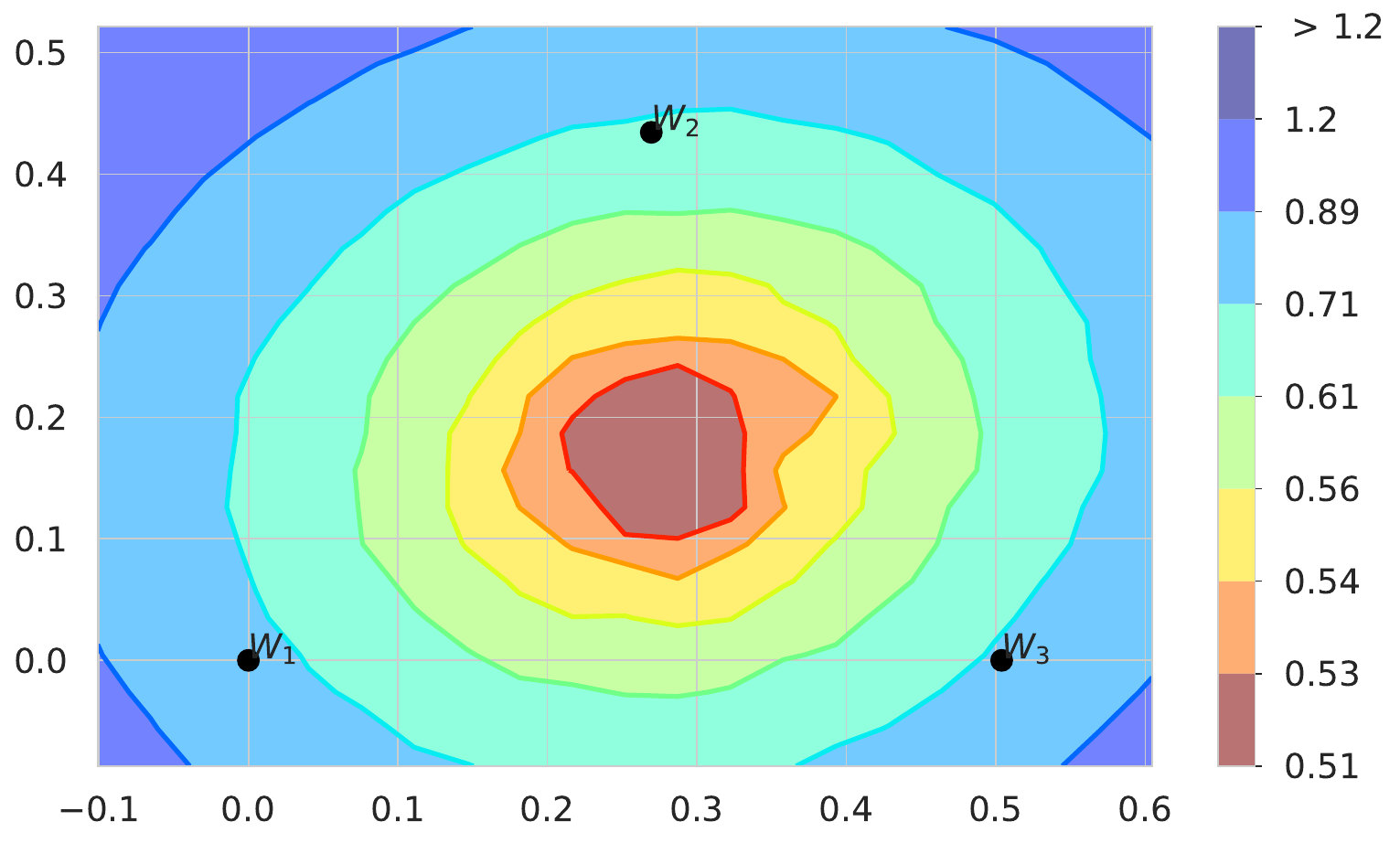}}
    \caption{Train cross-entropy loss surfaces computed with three local models after $20k$ training rounds on \textsc{Cifar100}. \textbf{(a)} $\alpha=0$ \textbf{(b)} $\alpha=0.5$ \textbf{(c)} $\alpha=1000$.}
    \label{fig:train_loss_plane}
\end{figure}

\begin{figure}[!t]
    \centering
    \subfloat[][]{\includegraphics[width=.25\linewidth]{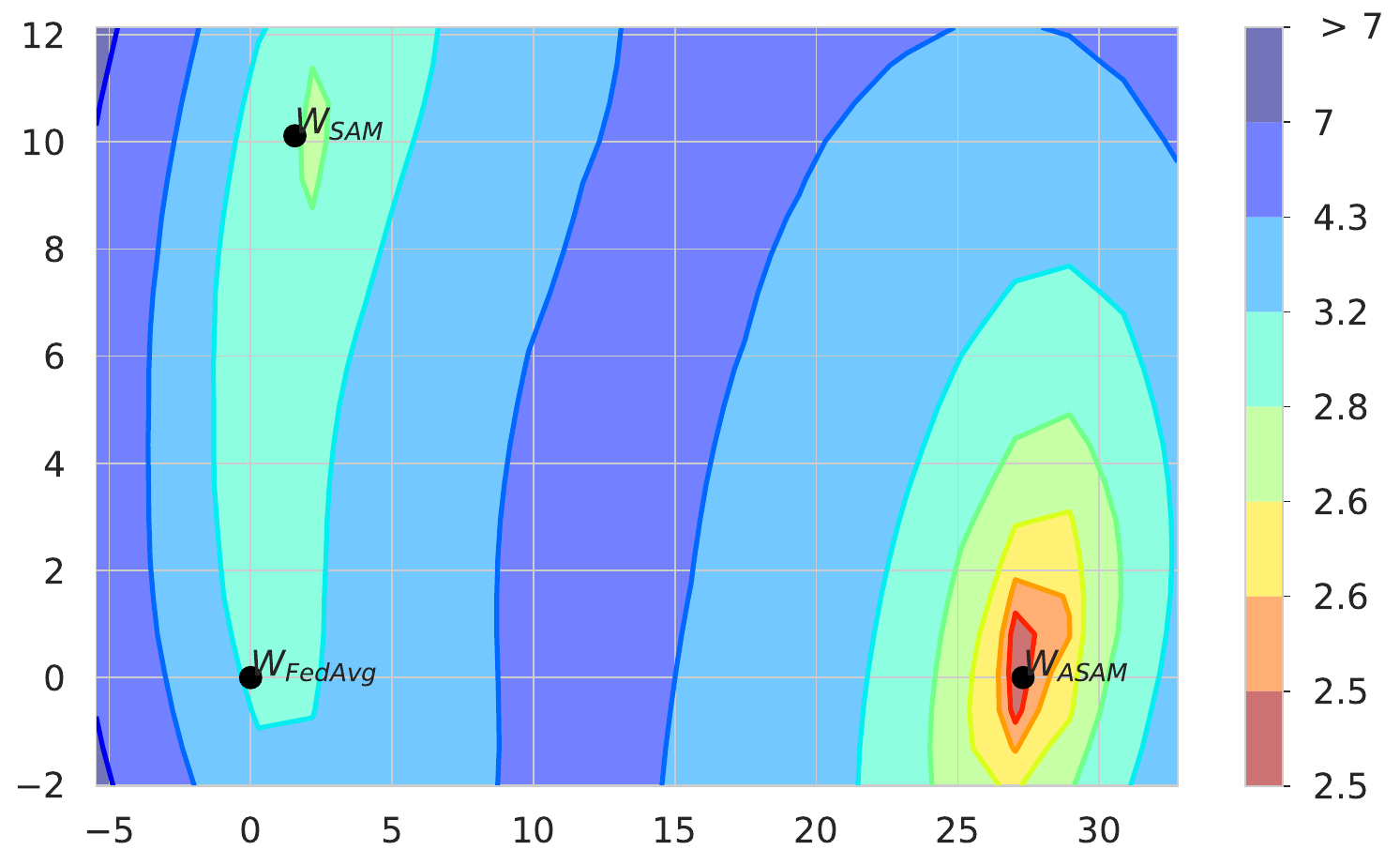}}
    \subfloat[][]{\includegraphics[width=.25\linewidth]{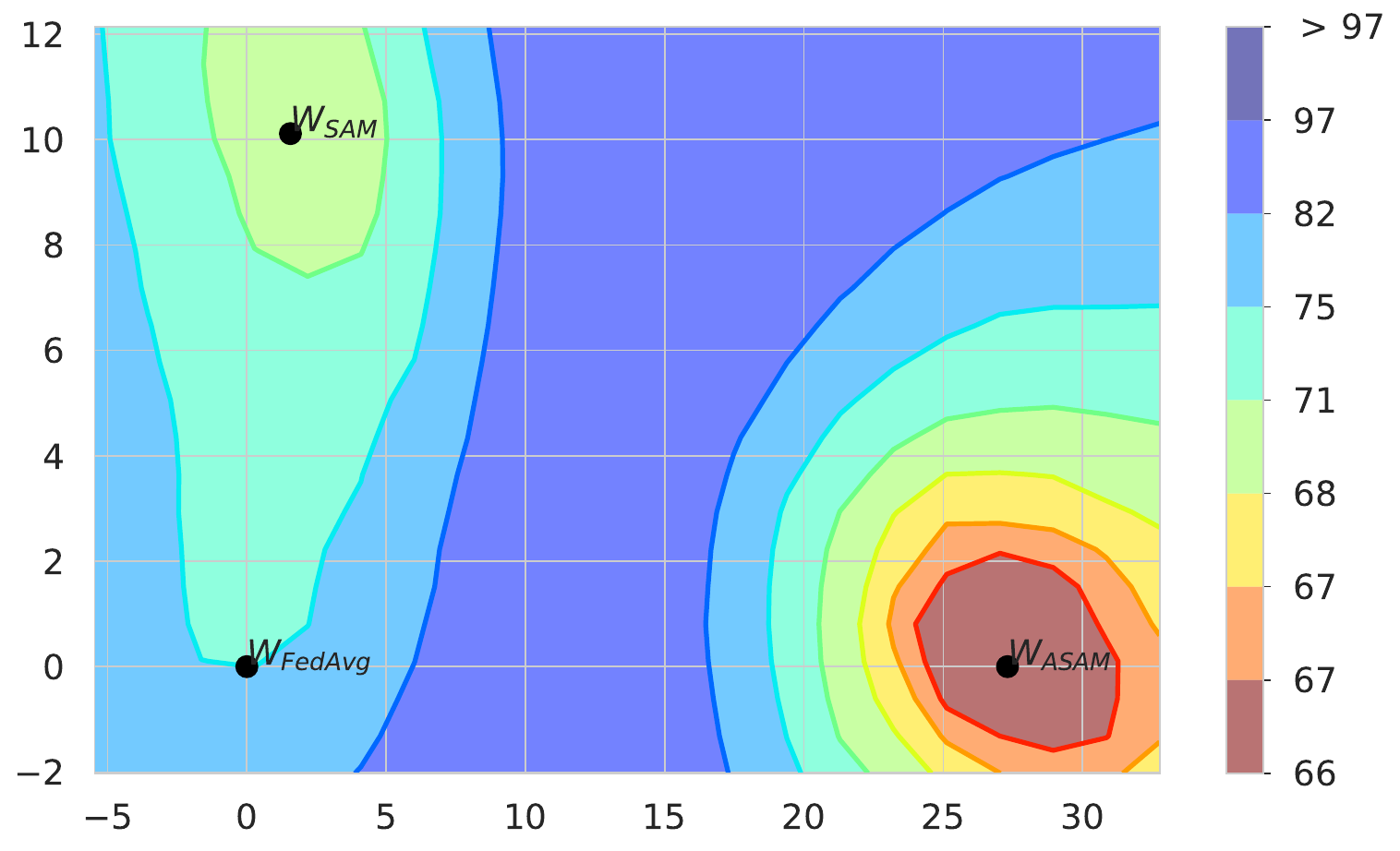}}
    \subfloat[][]{\includegraphics[width=.25\linewidth]{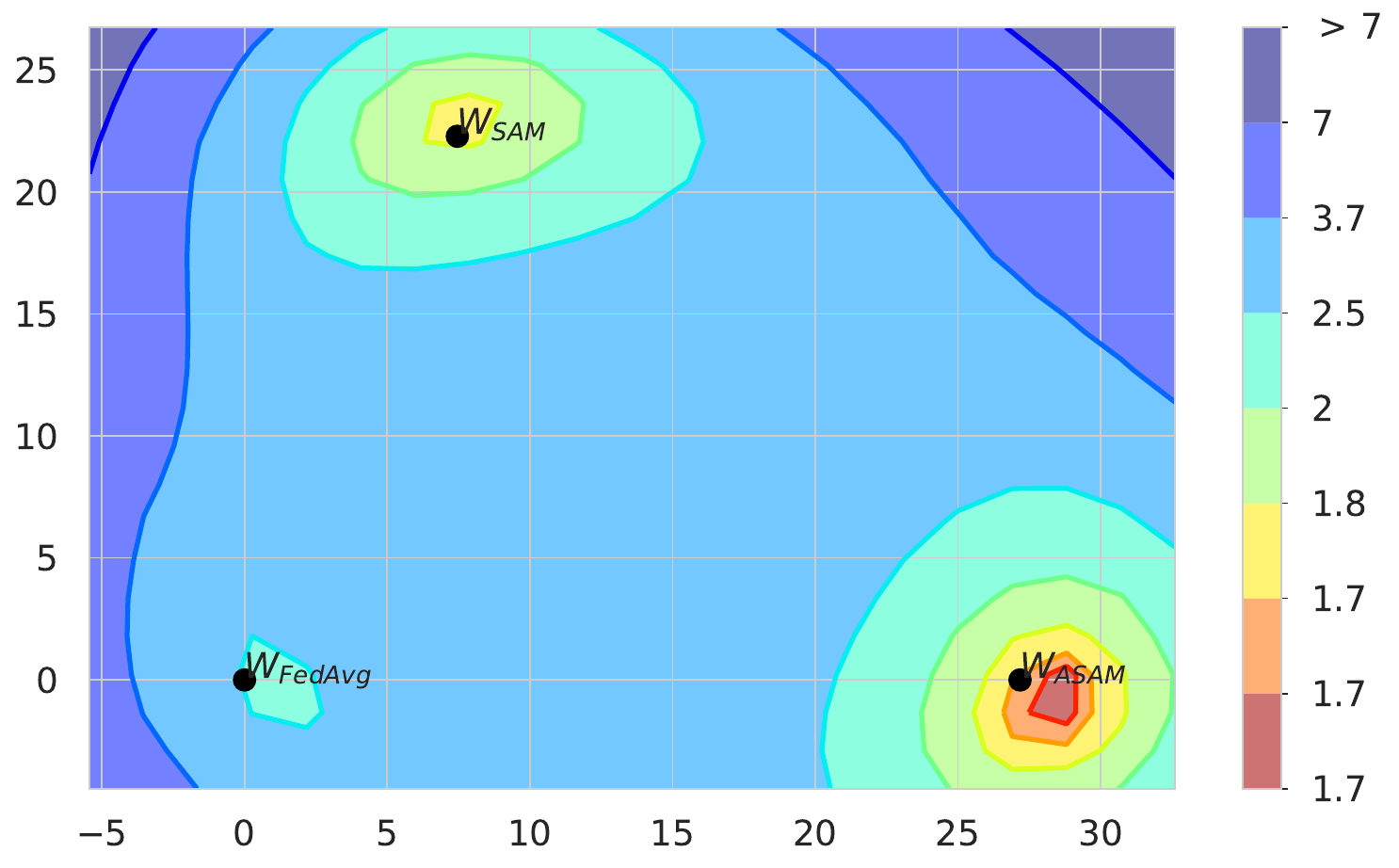}}
    \subfloat[][]{\includegraphics[width=.25\linewidth]{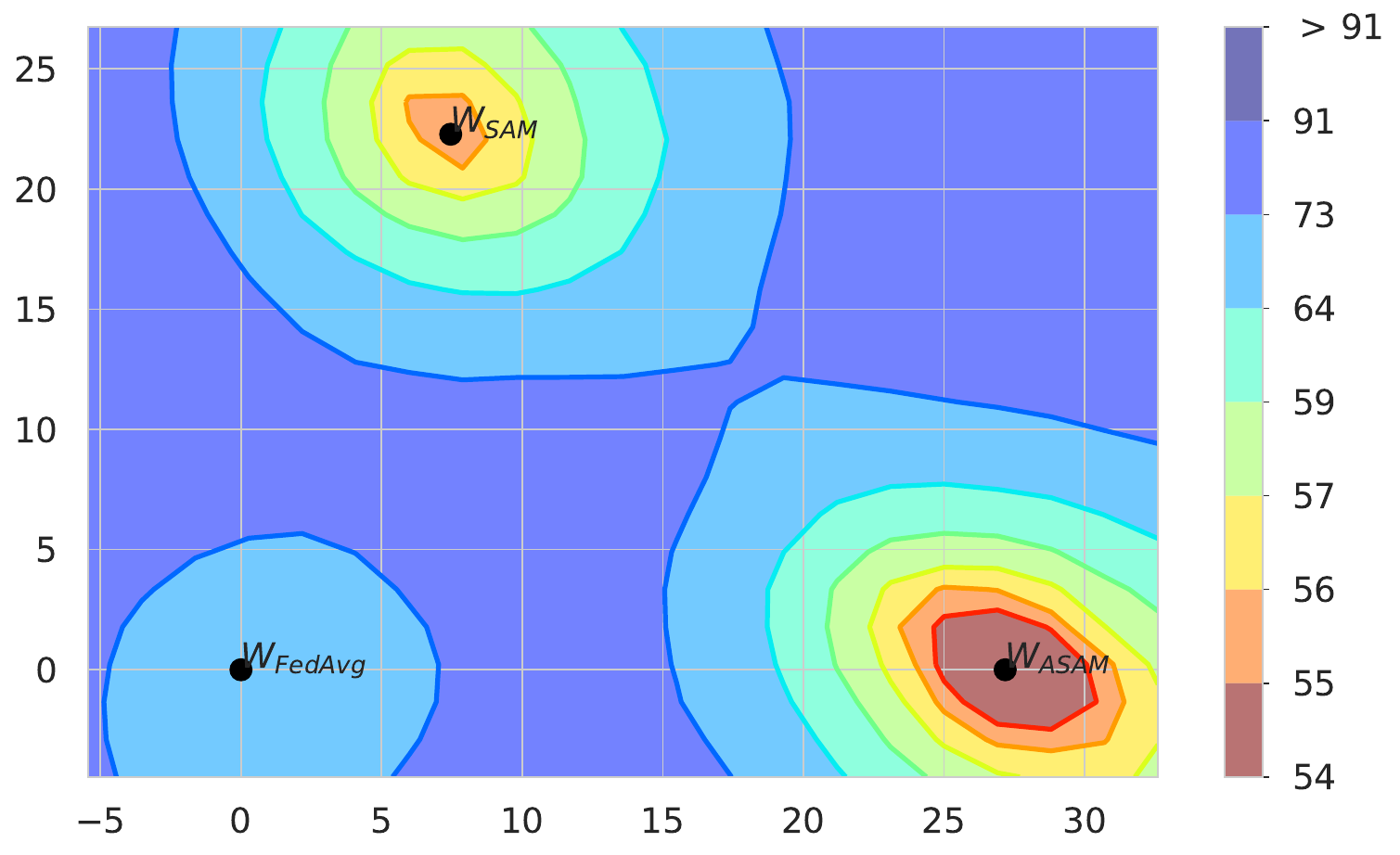}}
    \caption{Loss surfaces comparing the convergence points of \fedavg, \fedsam and \fedasam after $20k$ training rounds on \textsc{Cifar100}. The minima reached by \sam and \asam are found within low-loss neighborhoods. \textbf{(a)} Train loss surface $\alpha=0$. \textbf{(b)} Test error surface $\alpha=0$. \textbf{(c)} Train loss surface $\alpha=0.5$. \textbf{(d)} Test error surface $\alpha=0.5$. }
    \label{fig:convg_algs}
\end{figure}

\begin{figure}[!t]
    \centering
    \includegraphics[width=.4\linewidth]{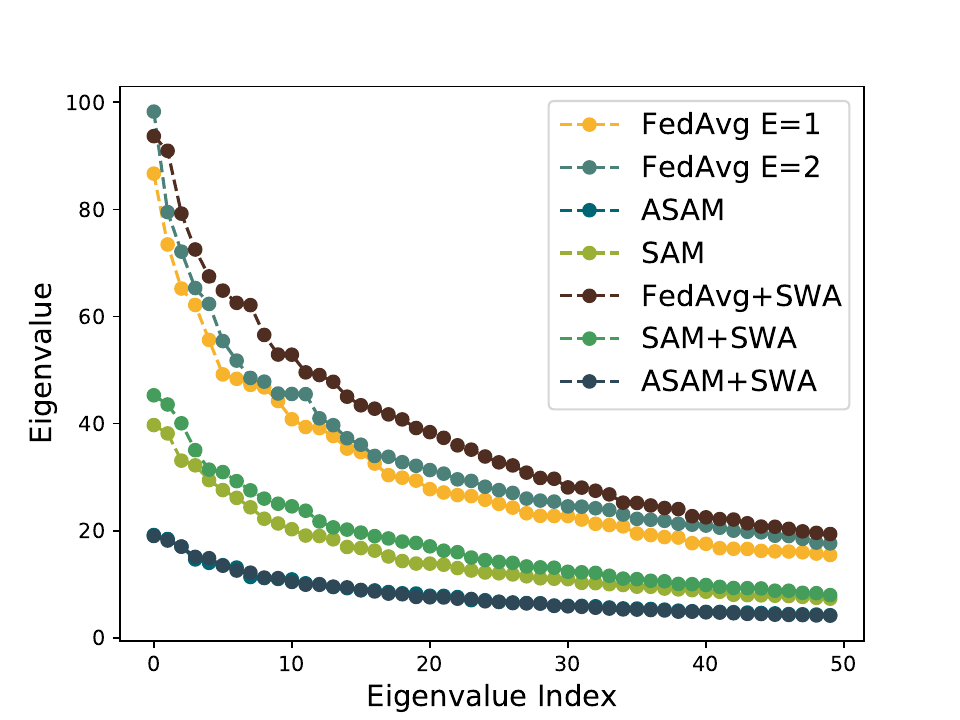}
    \caption{Top 50 eigenvalues of the global model with $\alpha=0.5$ on \textsc{Cifar100}.}
    \label{fig:a05_eigs}
\end{figure}

\begin{figure*}[!t]
\begin{minipage}{\columnwidth}
\centering
   \subfloat[][\fedavg $\alpha=0$]{\includegraphics[width=.25\linewidth]{images/eigs_a0_fedavg.png}}
   \subfloat[][\fedsam $\alpha=0$]{\includegraphics[width=.25\linewidth]{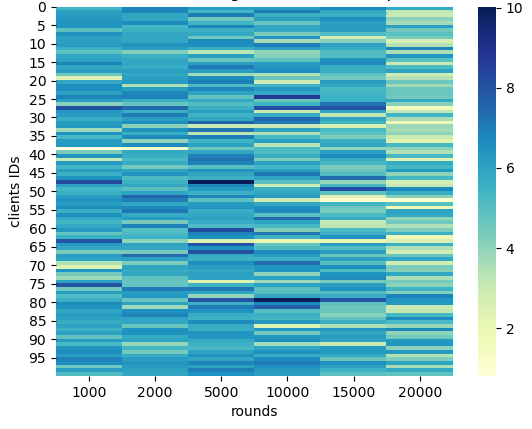}}
   \subfloat[][\fedasam $\alpha=0$]{\includegraphics[width=.25\linewidth]{images/eigs_a0_asam.png}}\\
    \subfloat[][\fedavg $\alpha=0.5$]{\includegraphics[width=.25\linewidth]{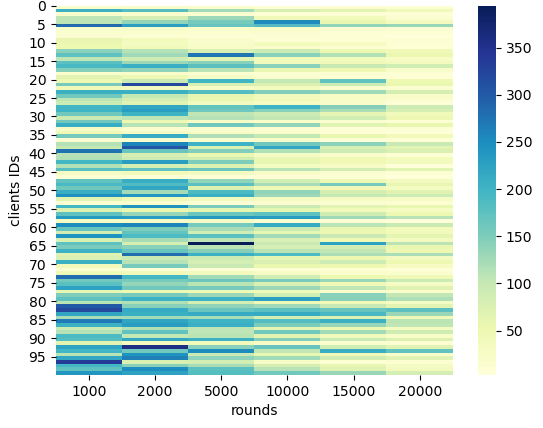}}
    \subfloat[][\fedsam $\alpha=0.5$]{\includegraphics[width=.25\linewidth]{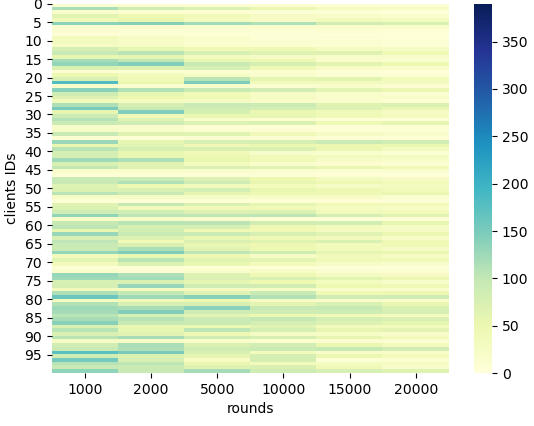}}
    \subfloat[][\fedasam $\alpha=0.5$]{\includegraphics[width=.25\linewidth]{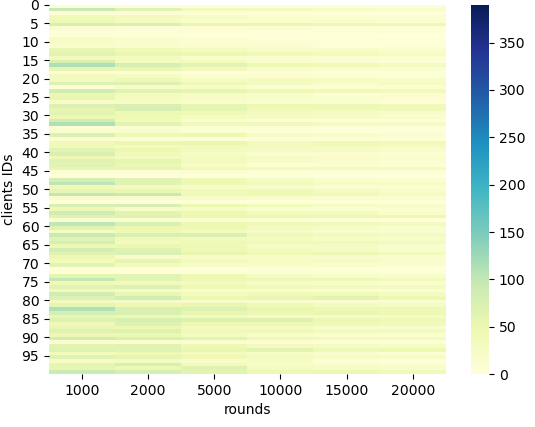}}\\
    \subfloat[][\fedavg $\alpha=1k$]{\includegraphics[width=.25\linewidth]{images/eigs_a1k_fedavg.png}}
    \subfloat[][\fedsam $\alpha=1k$]{\includegraphics[width=.25\linewidth]{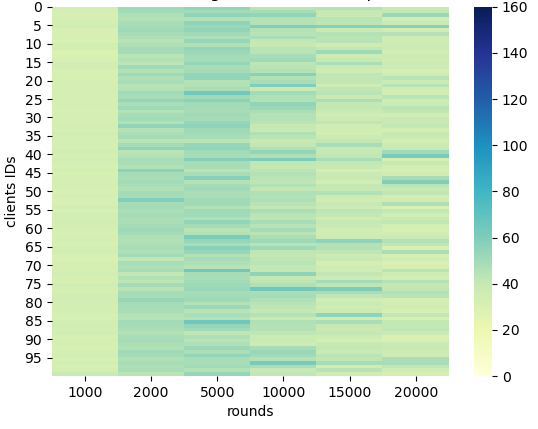}}
    \subfloat[][\fedasam $\alpha=1k$]{\includegraphics[width=.25\linewidth]{images/eigs_a1k_asam.png}}
    \caption{\footnotesize{Maximum Hessian eigenvalue computed for each client as rounds pass.}}
    \label{fig:clients_eigs_app}
\end{minipage}
\end{figure*}

All plots are best seen in colors.
\paragraph{Convergence plots} As shown in Sec. \ref{sec:cifar}, once combined with \fedavgm - \textit{i.e.} server-side momentum $\beta=0.9$ - \sam and \asam allow to reach convergence even in the most heterogeneous scenarios on both \textsc{Cifar10} and \textsc{Cifar100}. Fig. \ref{fig:convergence} shows the convergence plots of those runs. In addition, Fig. \ref{fig:conv_k5} compares the behavior of \fedavg, \fedsam, \fedasam and their combination with \swa on the most difficult setting, \textit{i.e.} $\alpha=0$ and 5 clients per round on both \textsc{Cifar} datasets, highlighting the stability and the positive gap in performance introduced by \swa. 
\paragraph{Loss Surfaces} Fig. \ref{fig:a05_client_conv} shows the convergence points of three local models trained with $\alpha=0.5$ on the corresponding test error surface, while Fig. \ref{fig:train_loss_plane} displays the train loss surfaces with $\alpha\in\{0,0.5,1000\}$. In addition, in Fig. \ref{fig:convg_algs} we compare the convergence points of \fedavg, \fedsam and \fedasam in the heterogeneous scenarios of \textsc{Cifar100}, \textit{i.e.} $\alpha\in\{0,0.5\}$, proving that \asam reaches the best local minimum. 
\paragraph{Hessian Eigenvalues} The top 50 eigenvalues of the global model trained with $\alpha=0.5$ are showed in Fig. \ref{fig:a05_eigs}. Fig. \ref{fig:clients_eigs_app} shows the complete comparison of the local Hessian eigenvalues partially shown in Sec. \ref{sec:het_fl}, introducing the values of $\lambda_{max}^k \: \forall k \in [K]$ resulting with \sam and $\alpha=0.5$. 

\captionsetup[subfloat]{font=scriptsize,labelformat=empty}

\captionsetup[subfloat]{font=scriptsize,labelformat=parens}

\captionsetup[subfloat]{font=scriptsize,labelformat=empty}

%% file: tables/server_opts.tex
\begin{table}[t]\centering
\caption{Final accuracy (\%) using different server-side optimizers with varying learning rate (LR) on \textsc{Cifar100} @ $20k$ rounds. 5\% clients participation. In bold the best results on both $\alpha=0$ and $\alpha=1k$.}\label{tab:server_optim}
\scriptsize
\setlength\tabcolsep{0.5cm}
    \begin{tabular}{lccc}
    \toprule
    Optimizer & LR & $\alpha=0$ & $\alpha=1k$\\
    \midrule
    \multirow{4}{*}{\texttt{SGD}} & 1 &\textbf{30.25}&\textbf{49.92}\\
    & 0.1 & 14.09 & 40.43\\
    & 0.01 & 2.67&11.35\\
    & 0.001 & 1.20 &1.12\\\midrule
    \multirow{4}{*}{\texttt{Adam}} & 1 & 1.00&{51.73}\\
    & 0.1 & 29.75&51.62\\
    & 0.01 & 13.72&40.12\\
    & 0.001 & 2.60&11.31\\\midrule
    \multirow{4}{*}{\texttt{AdaGrad}} & 1 & 1.00 &1.00\\
    & 0.1 & 1.77 &46.74\\
    & 0.01 & 26.25&51.44\\
    & 0.001 & 9.70&32.01\\
    \bottomrule
    \end{tabular}
\end{table}

%% file: tables/centralized2.tex
\definecolor{rosso}{HTML}{cc0000}
\definecolor{verde}{HTML}{279738}
\setlength{\tabcolsep}{4pt}
\begin{table}[h]
\begin{center}
\caption{\footnotesize{Comparison of improvements (\%) in centralized and federated scenarios ($\alpha\in\{0.5,1k\}$, 5 clients) on \textsc{Cifar100}, computed w.r.t. the reference at the bottom}}
\label{tab:centr2}
\tiny
\begin{tabular}{lccccccccc}
\toprule\noalign{\smallskip}
\multirow{2}{*}{Algorithm} & \multicolumn{3}{c}{Accuracy} & \multicolumn{3}{c}{Absolute Improvement} & \multicolumn{3}{c}{Relative Improvement}\\
\cmidrule(l){2-4} \cmidrule(l){5-7} \cmidrule(l){8-10} 
 & Centr. & $\alpha=0.5$ & $\alpha=1k$ & Centr. & $\alpha=0.5$ & $\alpha=1k$ & Centr. & $\alpha=0.5$ & $\alpha=1k$\\
\midrule
\sam & 55.22&44.73&54.01&+3.02&{\color{verde}{+4.30}}&{\color{verde}{+4.01}}&+5.79&{\color{verde}{+10.64}}&{\color{verde}{+8.03}}\\
\asam & 55.66&45.61&54.81&+3.46&{\color{verde}{+5.18}}&{\color{verde}{\textbf{+4.89}}}&+6.63&{\color{verde}{+12.81}}&{\color{verde}{\textbf{+9.80}}}\\
\swa&52.72&43.90&50.98&+0.52&{\color{verde}{+3.47}}&{\color{verde}{+1.06}}&+1.00&{\color{verde}{+8.58}}&{\color{verde}{+2.12}}\\
\samswa&55.75&47.96&53.90&+0.55&{\color{verde}{+7.53}}&{\color{verde}{+3.98}}&+1.06&{\color{verde}{+18.63}}&{\color{verde}{+7.97}}\\
\asamswa&55.96&49.17&53.86&+3.76&{\color{verde}{\textbf{+8.74}}}&{\color{verde}{+3.94}}&+7.20&{\color{verde}{\textbf{+21.62}}}&{\color{verde}{+7.89}}\\
\mixup&58.01&35.10&55.34&+5.81&{\color{rosso}{-5.33}}&{+5.42}&+11.13&{\color{rosso}{-13.18}}&{+10.86}\\
\cutout&55.30&37.72&53.48&+3.10&{\color{rosso}{-2.71}}&{\color{verde}{+3.56}}&+5.94&{\color{rosso}{-6.70}}&{\color{verde}{+7.13}}\\
\midrule
\multicolumn{7}{l}{\texttt{Centralized}: \textbf{52.20} - \fedavg $\alpha=0.5$: \textbf{40.43},  $\alpha=1k$: \textbf{49.92}}\\
\bottomrule
\end{tabular}
\end{center}
\end{table}
\setlength{\tabcolsep}{1.4pt}

%% file: tables/cifar10_data_augms.tex
\setlength{\tabcolsep}{4pt}
\begin{table}[]
\begin{center}
\caption{\fedavg, \sam, \asam and \swa w/ strong data augmentations (\mixup, \cutout) on \textsc{Cifar10}}
\label{tab:augms_cifar10}
\tiny
\begin{tabular}{llccccccccccc}
\toprule\noalign{\smallskip}
 & \multirow{2}{*}{Algorithm} & \multirow{2}{*}{SWA}&\multirow{2}{*}{Aug} &\multicolumn{3}{c}{$\alpha=0$} &\multicolumn{3}{c}{$\alpha=0.5/0.05$} & \multicolumn{3}{c}{$\alpha=1000/100$}\\
\cmidrule(l){5-7} \cmidrule(l){8-10} \cmidrule(l){11-13}
& & &&$5cl$& $10cl$ & $20cl$ & $5cl$& $10cl$ & $20cl$ & $5cl$& $10cl$ & $20cl$\\
\noalign{\smallskip}
\hline
\noalign{\smallskip}
\multirow{18}{*}{\rotatebox[origin=c]{90}{\textsc{Cifar10}}}&\fedavg&\ding{55}&\multirow{6}{*}{\rotatebox[origin=c]{90}{\texttt{None}}}&65.00 & 65.54 & 68.52 & 69.24 & 72.50 & 73.07  &84.46 & 84.50& 84.59\\
&\fedsam&\ding{55}&&70.16 & 71.09 & 72.90 & 73.52 & 74.81 & 76.04  &84.58 & 84.67 &\textbf{84.82}\\
&\fedasam&\ding{55}&&73.66 & 74.10 & 76.09 & 75.61 & 76.22 & 76.98  & 84.77 &84.72 &84.75\\
&\fedavg&\ding{51}&&69.71 & 69.54 & 70.19 & 73.48 &  72.80   &73.81 &84.35 & 84.32&84.47\\
&\fedsam&\ding{51}&&74.97 &73.73  & 73.06 & 76.61 & 75.84  & 76.22&84.23 & 84.37&84.63\\
&\fedasam&\ding{51}&&\textbf{76.44} & \textbf{75.51} & \textbf{76.36}  & \textbf{76.12} & \textbf{76.16}  & \textbf{76.86}& \textbf{84.88}&\textbf{84.80} &\textbf{84.79}\\
\cmidrule{2-13}
&\fedavg&\ding{55}&\multirow{6}{*}{\rotatebox[origin=c]{90}{\mixup}}&62.26&63.61&65.54&65.63&68.44&68.21&\textbf{82.38}&\textbf{84.46}&\textbf{83.58}\\
&\fedsam&\ding{55}&&67.35&69.32&69.78&70.34&72.98&72.54&81.88&82.24&82.25\\
&\fedasam&\ding{55}&&70.61&71.31&71.62&72.19&\textbf{72.84}&\textbf{72.72}&82.36&82.75&83.08\\
&\fedavg&\ding{51}&&66.31&66.89&66.26&69.79&69.12&68.80&82.27&82.88&82.67\\
&\fedsam&\ding{51}&&\textbf{72.42}&70.65&69.75&\textbf{73.36}&72.29&72.44&81.04&81.18&81.15\\
&\fedasam&\ding{51}&&72.37&\textbf{72.40}&\textbf{71.89}&72.54&72.36&72.32&81.86&81.70&81.92\\
\cmidrule{2-13}
&\fedavg&\ding{55}&\multirow{6}{*}{\rotatebox[origin=c]{90}{\cutout}}&61.12&64.47&64.20&66.45&69.09&68.99&\textbf{83.77}&83.91&\textbf{84.31}\\
&\fedsam&\ding{55}&&63.69&66.30&67.25&67.66&71.39&70.67&83.03&83.84&83.49\\
&\fedasam&\ding{55}&&68.50&69.26&69.75&69.23&\textbf{71.91}&\textbf{71.28}&83.73&\textbf{84.10}&84.00\\
&\fedavg&\ding{51}&&65.54&65.60&65.79&69.94&69.55&69.63&83.35&83.39&83.64\\
&\fedsam&\ding{51}&&69.40&68.45&67.36&71.36&71.56&70.99&82.61&82.75&82.52\\
&\fedasam&\ding{51}&&\textbf{71.30}&\textbf{71.12}&\textbf{70.91}&\textbf{72.79}&71.76&71.09&83.06&83.31&83.11\\
\bottomrule
\end{tabular}
\end{center}
\end{table}
\setlength{\tabcolsep}{1.4pt}